# Development of a Tendon Driven Variable Stiffness Continuum Robot with Layer Jamming

An Undergraduate Honors Thesis

Presented to The Department of Mechanical Engineering

In Partial Fulfillment of the Requirements for

Graduation with Distinction in Mechanical Engineering

By

Zhong Ouyang

November 5, 2019

The Ohio State University

Thesis Committee

Dr. Haijun Su, Advisor

Dr. Ruike Zhao, Committee Member

# Abstract

Robotics has transformed the live of human beings significantly in numerous ways. Among all categories of robotics, soft robot is probably the most intriguing and promising realm of robotics research. Compared with traditional robots featuring rigid-body components, soft robots have compliant and continuum structures. Soft robots tend to have a greater dexterity, and they are capable of navigating through tortuous paths or bypassing obstacles by adjusting the robotic shape to reach spots where conventional robotic arms cannot access. However, there exist several flaws of continuum robots with respect to lack of stiffness and load carrying capacity. To solve these inherent issues with continuum robots, previous research has already explored the application of layer jamming to improve the rigidity and load carrying capacity of one arm segment which can achieve motion within a 2D plane. The purpose of this research is to design, fabricate and test a tendon driven a continuum soft robot with three modular segments, each of which has a tunable stiffness enabled by layer jamming technology. Compared with previous studies, the robotic arm design of this project has a modular structure, which means the length of the robotic arm can be adjusted by addition of extra arm modules/segments to the existing robotic prototype. Furthermore, the new arm prototype supports motion within a 3-dimensional space. To achieve the goals, the design and fabrication for the variable stiffness robotic arm with compliant main structure and layer jamming mechanism has already been finished. Design and fabrication of the connector has also been finished to integrate several link modules into one robotic arm with multiple segments. The actuator located at the base of the arm has already been designed and tested. Finally, a stiffness test of one arm segment was conducted to verifying the load carrying capacity of the variable stiffness robotic arm, then the stiffness ratio of the layer jammed structure was calculated to analyze the stiffness improvement compared with unstiffened soft robot.

# Acknowledgments

First, I am very grateful to Dr. Haijun Su for allowing me to conduct undergraduate research and utilize facilities in DISL. Dr. Su has been consistently dedicated and committed to helping me finding solutions to issues confronted during every stage of my project, especially for details and verification of the conceptual design, kinematic analysis and experimental procedures for testing my variable stiffness continuum robot prototype.

In addition, I would like to express great gratitude to PhD student Xianpai Zeng in DISL for teaching me to utilize 3D printer, force gauge, displacement gauge and the vacuum pump in lab. I also would like to thank Xianpai for helping me contemplating detailed procedures for the load capacity test and actuation test.

I would like to thank Tylor Morrison and Yuan Gao for teaching me to utilize the laser cutter and data acquisition software during the fabrication and testing of my continuum robot prototype.

Finally, I have to thank my parents for their unwavering support in my pursuit of an engineering dream. Without financial and moral support from my parents, I could not be able to be concentrated on my coursework and research projects during the past four years.

# Table of Contents

# List of Figures

# Chapter 1 Introduction

## 1.1 Soft Robotics and Applications

Robotics has transformed the live of human beings significantly in numerous significant ways. Among all categories of robotics, continuum robot is probably the most intriguing and promising realm of robotics research. Compared with traditional robots featuring rigid bodied components, soft robots have compliant and continuum structures. Because of the compliant structure, the structure of soft continuum robots IS radically different. Since soft robots have quite flexible links, they have infinite number of joints, which gives them redundant degrees of freedom and high flexibility [2].

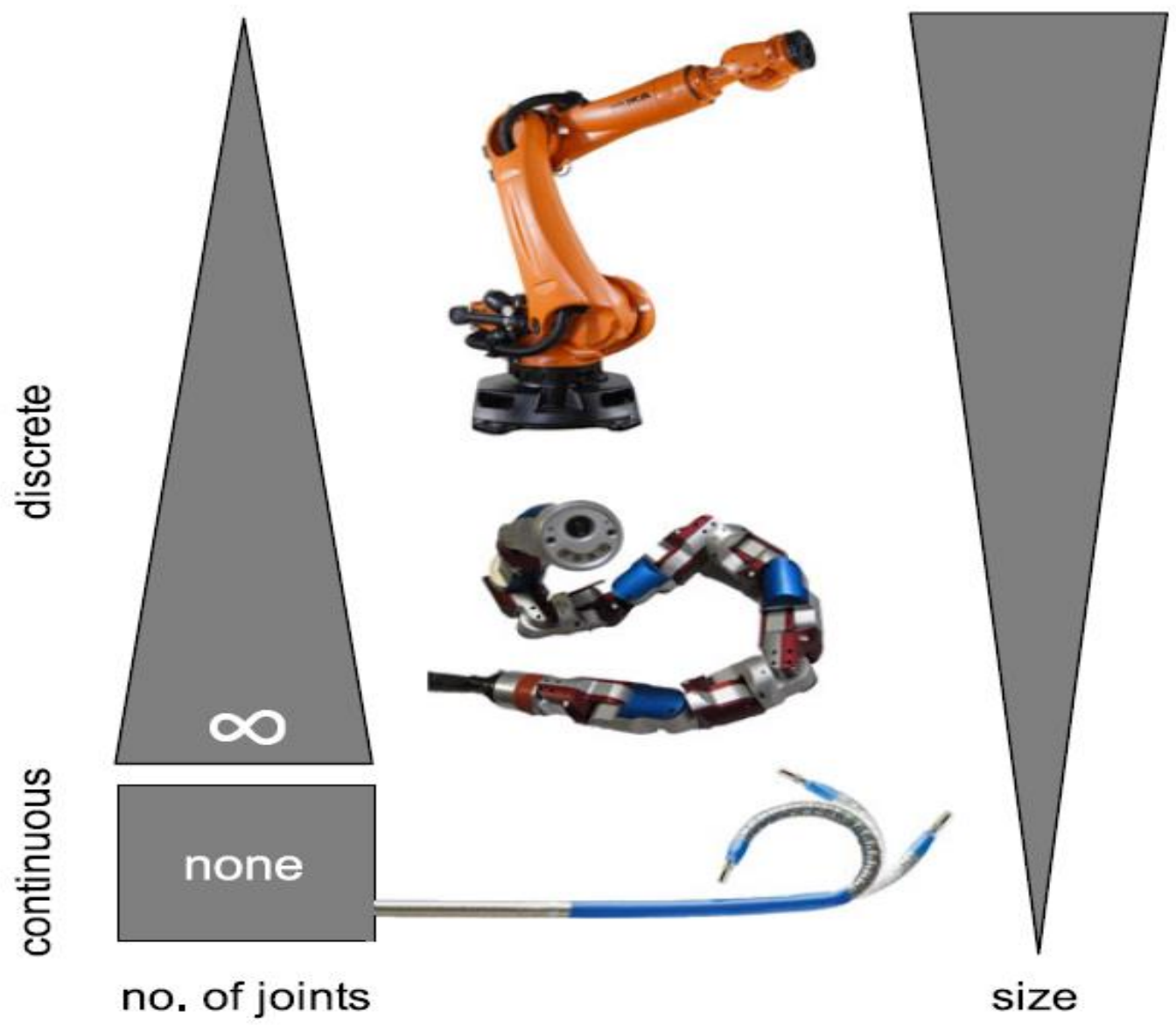

*Figure 1-1: Infinite Degrees of Freedom for Continuum Robots [2]*

This enables them to be fabricated into smaller scales and miniaturized easily. Flexible links also mean that soft robots tend to have a greater dexterity, which makes them able to be navigated through tortuous paths or bypass obstacles by adjusting the robotic arm shape to reach spots where conventional robotic arms cannot access [1]. In addition, due to the high flexibility, soft robotic structure is capable of absorbing energy, thus reducing the impact force and the damage of the machine itself and external objects when the robot arm is malfunctioning [1] [3]. The compliant characteristic, therefore, enables soft robots to achieve a safe interaction between the machine, user and external environment.

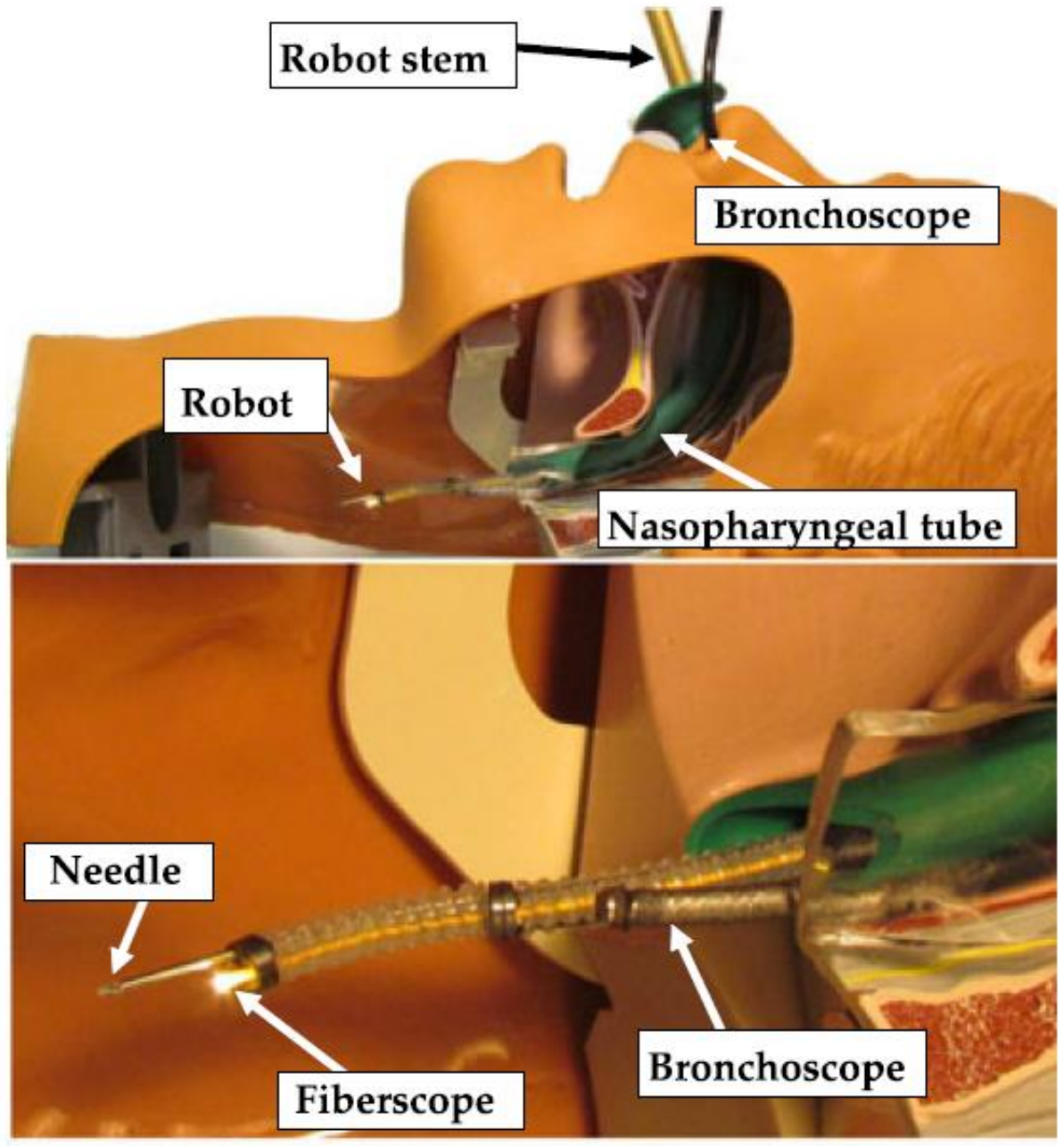

*Figure 1-2:Application of Continuum Robots in Surgery [2]*

One of the most significant application of continuum robots is for invasive surgery. For continuum surgical robots, since they have infinite degrees of freedom and high dexterity, this type of robot can reach remote surgical spots with minimum instrumentation without opening the patient's body. With a fully developed control algorithm, a continuum can be manipulated and reach the surgical spots with high precision and accuracy. Therefore, the continuum surgical robots can reduce patient discomfort and surgery duration significantly compared with conventional surgery methods. Figure 1-2 below shows the application of continuum robots in bronchoscope. As seen on the figure, the continuum robot can be manipulated with high dexterity and navigated through the torturous respiratory tract to enter the bronchus of a patient [2].

Remote inspection, search and rescue operations is another promising application for continuum robots. Again, because of the high dexterity and flexible structure, continuum robots can be navigated through torturous channels and bypass obstacles along the route. Small diameter and high dexterity allow continuum robots to work within in extremely confined spaces. This means that soft robot can be released after a natural disaster such as an earthquake or tsunami to enter the debris of a collapsed building. the continuum robot can serpentine under the debris where human being cannot reach to search for survivors. Figure 1-3 in the next page illustrates an extremely thin tendril continuum robot developed for search and inspection purposes. The robot has a high dexterity with a maximum diameter less than 1 cm, and can enter and inspect the sewage system through the drain cover [18].

However, there exists several flaws of continuum robots. Again, because of the high compliance, soft robots tend to deform when the robotic arm is under acceleration, causing difficulties in control of the position for each arm segment [4]. As the soft robot has infinite degrees of freedom, its shape is insufficiently constrained. Lack of stiffness also undermines the force and torque carrying capability of soft robots when compared with traditional rigid robotic arms.

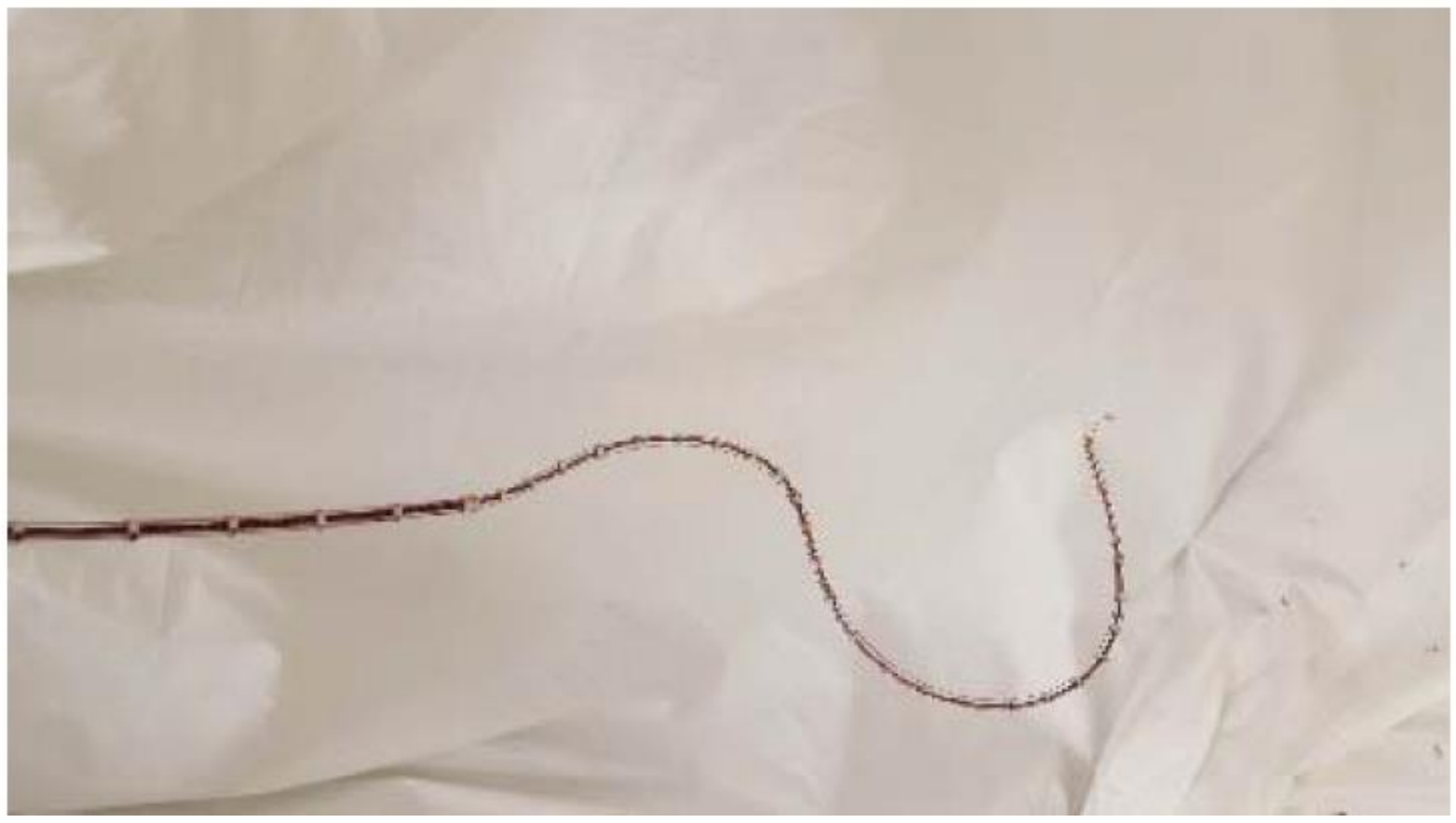

*Figure 1-3: Tendril Continuum Robot for Search and Inspection Purposes [18]*

## 1.2 Variable Stiffness Control

Since there are several inherent issues with soft robots, it is increasingly imperative to develop advanced methods and technologies to mitigate the performance constraints. As is mentioned in the previous section, soft robotics are lack of rigidity and prone to deformation under forces or acceleration, which generates issues of precision control and load carrying capacity. To resolve these issues, a variable stiffness control system should be introduced to the existing soft robot in order to combine the advantages of flexibility, dexterity, precision control, rigidity and load carrying capacity simultaneously.

One the one hand, the stiffness of the robot should be able to be adjusted to a high value, under a scenario where the main task of the robotic arm is to carry the load, achieve precise location control while maintaining the shape and resisting deformation [4]. It is crucial for a variable stiffness robot to maintain the high rigidity when its arm is under high speed motion to improve the work efficiency of an industrial robot, such as a welding robotic arm in an automotive factory. In an automotive plant, the stiffness of the robotic arm has to be sufficiently high to maintain relatively high motion speed, thus improving the speed of vehicle production. However, the stiffness should be lowered instead to reduce harm to humans when the robotic arm collides with human operators accidentally.

On the other hand, the stiffness of the robotic arm should be tuned to a low level to satisfy high flexibility and dexterity demands under other occasions where the capability of bypass obstacles plays a dominant role. For example, the manipulator of surgical robots is supposed to have a high dexterity and precision, in order to be navigated through tortuous channel inside a human body, thus reaching remote surgical spots with minimum instrumentation without opening the body of the patient [2]. However, the stiffness of the surgical robots should be enhanced after reaching the surgical spot, since the rigidity of the snaked like robotic arm should be sufficiently high to hold surgical instruments and withstand corresponding reaction forces. This advantage enables variable stiffness surgical robots to reduce patient discomfort and surgery duration significantly.

### 1.2.1 Technical Solutions for Variable stiffness

1.2.1.1 Change of the cross-section geometry

There exist several studies pertaining to tuning the stiffness of a robotic arm section continuously through direct cross sectional geometry variation. In Design Innovation and Simulation Lab (DISL) has already attempted several approaches to change the shape of the cross section by changing the pattern of an array of beams of a robotic arm. One method developed by DISL is to design a robot arm which consists of 4 rotatable beams. This 4-rotating-beams design can achieve a good stiffness ratio of 13.9, which means that the highest achievable stiffness is 12.9 time greater than the minimum stiffness [5]. This variable stiffness design can also achieve an angular stiffness ratio of 8.6 which subjected to torsional loads [5]. One advantage of the concept is that the structure can achieve a high actuation speed, since the configuration of the cross section can be change instantly by servo motor actuation. However, the 4-beam design might occupy a significant amount of volume, causing it difficult to be utilized for surgical operations where the operational space is quite narrow and torturous. Therefore, this 4-beam variable stiffness approach will not be utilized for further considerations.

#### 1.2.1.2 Variable stiffness through low melting point materials

Recently, a group of researchers have proposed to utilize low melting point materials (LMPMs) to achieve variable stiffness [7]. The discovered that wax can be utilized as the low melting point material due to its low expense and high accessibility [6]. As the temperature goes higher, the low melting point material becomes much softer, thus achieving the stiffness variation. Under a low temperature, the material can be quite hard since the LMPM has sufficient hardness to lock the lattice structure, thus preventing the relative displacement between the cells. Although the material has a tremendous amount of stiffness variation, it is difficult to achieve precise temperature control of the structure, and the addition of the heating system and temperature management devices might occupy a considerable amount of space. In addition, its response is quite slow compared with other variable stiffness methods, since it takes some time for the structure temperature to reach the specified value. Therefore, this technical approach was not applied in the author's variable stiffness robot development.

#### 1.2.1.3 Variable stiffness through magnetorheological and electrorheological materials

For electrorheological and magnetorheological materials, they can behave like fluids since their rheological properties can be adjusted under the applied electric or magnetic field [6]. This type of material is commonly utilized in the automotive industry for shock absorbers in the vehicle suspension system. With the addition of magnetorheological or electrorheological materials into compliant matrix, the overall structure can manifest variable stiffness. Under the application of the magnetic field, the magnetic particle can be aligned from the initial random state, then resist the longitudinal stretches of the structure, thus improving the longitudinal stiffness considerably [6]. However, the yield stress generated by this type of material is too small and the maximum stiffness is not sufficiently high too achieve sufficient load capacity. Therefore, this approach will not be considered for the purpose of this project.

#### 1.2.1.4 Variable stiffness through layer jamming

Layer jamming means the generation of friction between layers under the application of negative air pressure. To maintain the negative pressure, the compliant main structure has to be sealed inside a vacuum bag. To achieve variable stiffness control, the bag should be connected to a vacuum pump via a gas tube. As is shown in Figure 1-4, there are several friction layers placed inside a sealed bag with one end attached to the vacuum pump.

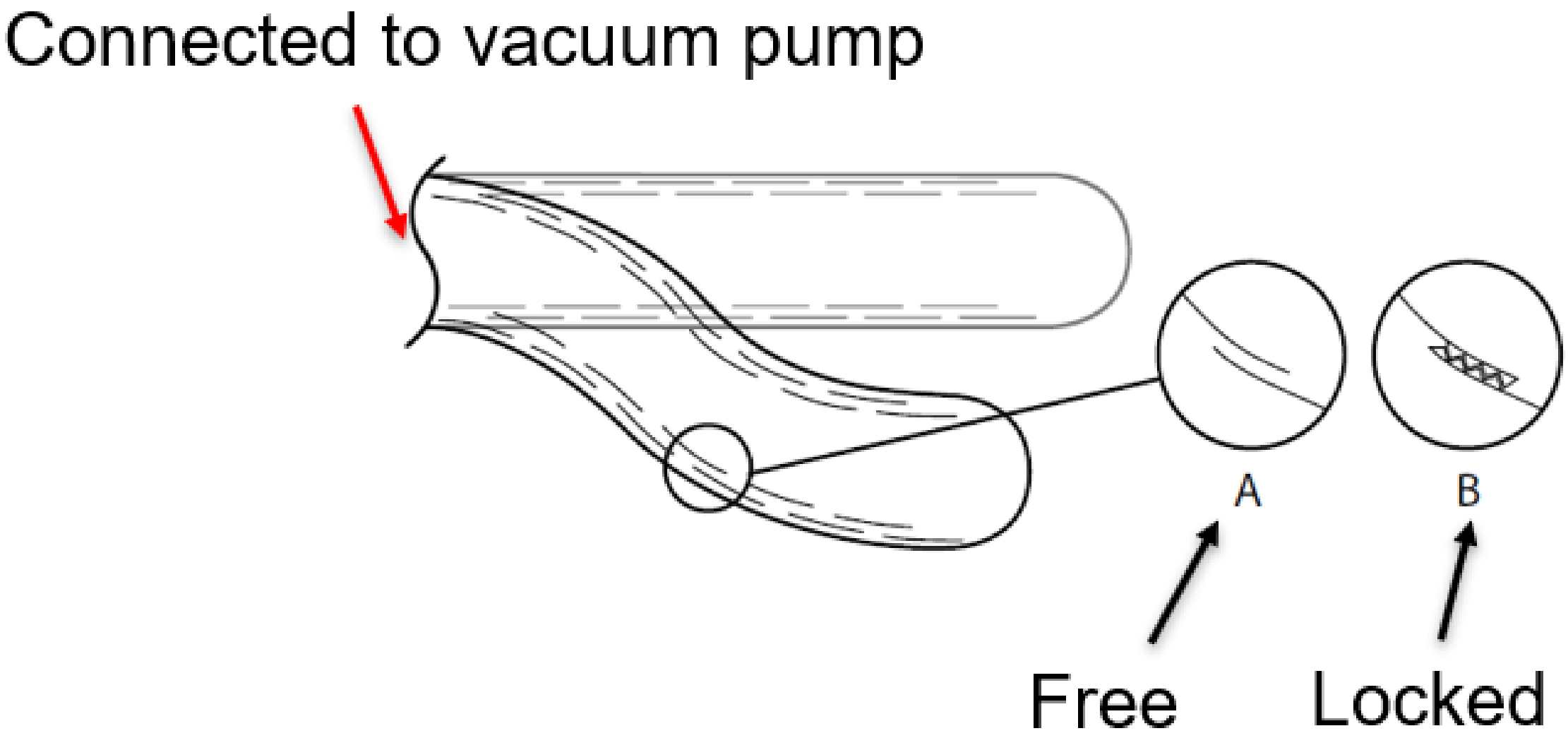

*Figure 1-4: Illustration of Layer Jamming [8]*

When the vacuum pump is not engaged, the pressure inside the bag is the same as the atmospheric pressure, which corresponds to almost zero pressure between the friction layers. In this case, the friction between the thin laminates can be very small, which allows relative displacement between the thin sheets. Hence the structure has great freedom to deform, thus manifesting a low stiffness. However, after starting the vacuum pump and the establishment of great negative pressure inside the vacuum bag, the relative pressure between the adjacent laminates can be extremely high and generate a great maximum friction force. This will constrain the relative displacement between the friction layers. In this case, the laminates and the whole structure is locked and manifest a high stiffness. By continuous adjustment of the vacuum pressure established inside the bag, the stiffness of the overall structure can be tuned continuously, thus achieving variable stiffness control.

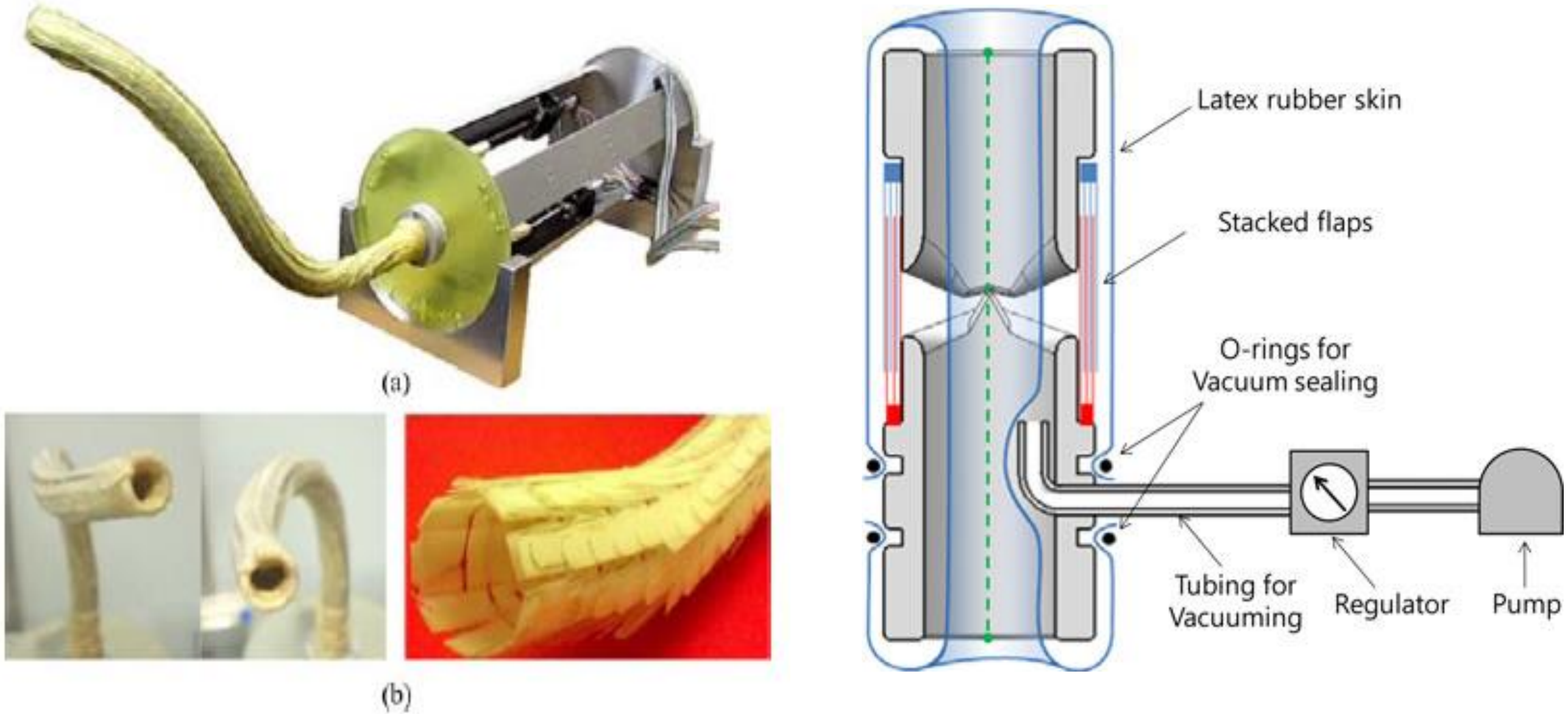

*Figure 1-5: Structure of a Tubular Continuum Robot [9]*

As is shown on Figure 1-5, the continuum structure of Kim et al [9] is enclosed by a vacuum bag connected to the pump. Stacked flaps are attached to the adjacent segments of the robot arm, and the friction layer can lock up when vacuum pressure is introduced inside the vacuum bag. The structure of the layer jamming system is quite compact and can be miniaturized easily for surgery operation purposes. In addition, its stiffness can be adjusted continuously by changing the vacuum pressure, which can also achieve a quick response in stiffness variation.

In addition, DISL has also explored approaches for variable stiffness robot design. Based on layer jamming, Yuan Gao of DISL has developed a novel variable stiffness compliant robotic gripper [19]. The gripper has 3 fingers, and each finger has a compliant back bone which can bend within a plane, as is presented in Figure 1-6. Each finger is actuated by a pair of antagonistic tendons of which the ends are fixed to the finger tip, and the other ends are driven by a servo motor. For the layer jamming system, friction layers are attached to both ends of the backbone, then the compliant backbone and the friction layers are sealed in a vacuum bag.

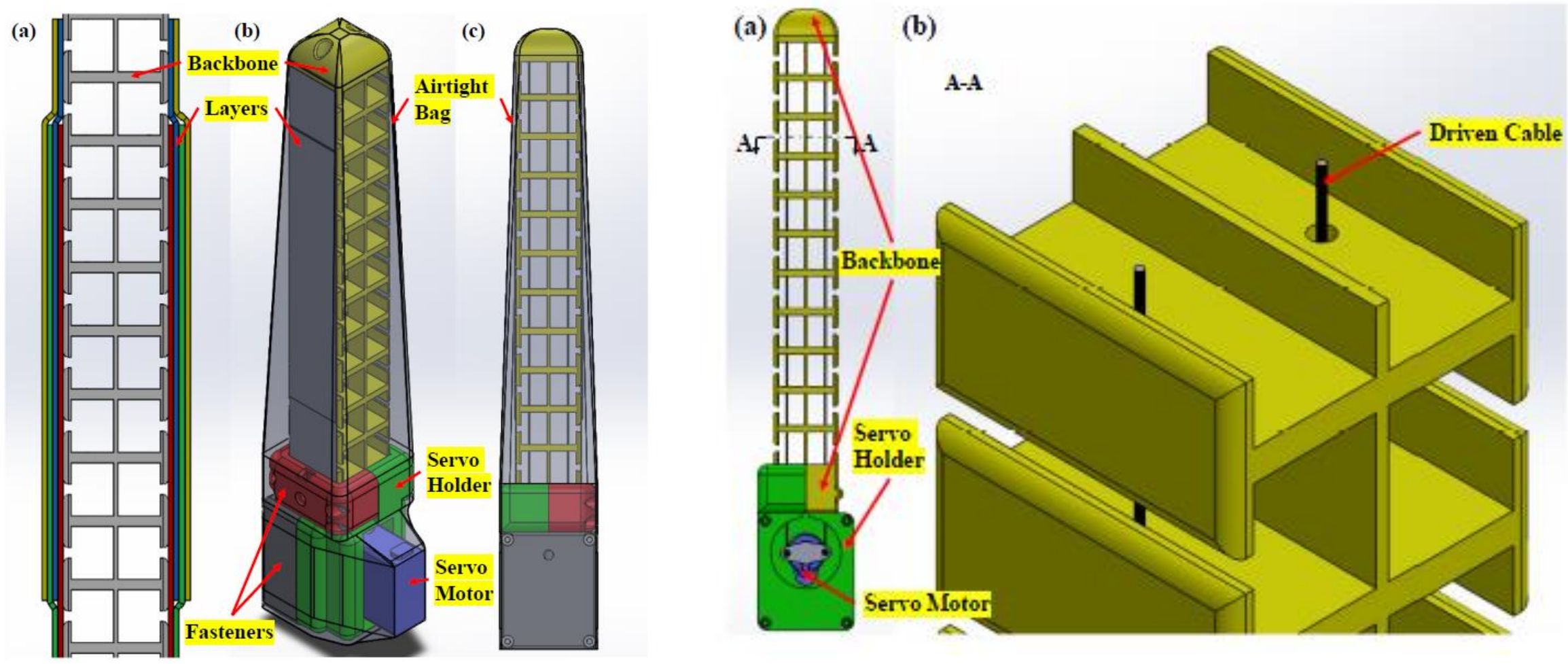

*Figure 1-6: Design of The Variable Stiffness Gripper Finger Based on Layer Jamming [19]*

The results of variable stiffness via layer jamming is quite successful in this application, since Gao's testing results indicate that the stiffness of the gripper under the negative pressure of 12.5 psi is 33 times higher than the original state [19]. Although this gripper design can achieve a high stiffness via layer jamming, it has issue with the dexterity of the finger. The structure is capable of planar motion only, since the backbone of the compliant structure is a piece of PLA sheet which allows planar bending while constrains special bending in other directions.

Since layer jamming has already be verified in DISL as a premising method to achieve high stiffness variation, this technique is chosen for further development in this variable stiffness continuum robot project.

## 1.3 Actuation for Continuum Robot

### 1.3.1 Shape Memory Alloy Actuation

Shape Memory Alloy (SMA) is generally believed as a revolutionary material which could transform robotic actuation research, because they have a high power density, great capability to recover the shape to the original state and high compatibility with tissues of human beings [11]. For shape memory alloy, its original shape can be well defined and maintained, while its shape can be transformed after experiencing a temperature change. The NiTi alloy is probably the most commonly used SMA material for robotic actuation [8]. SMA is actually a very promising material for robotic actuation because of its high power density and great compactness. However, it has long actuation time and the response is quite slow. It was observed that the robot finger developed by DISL previously required up to 30 seconds to complete response and reach the desired position [10]. Since the response of SMA is not sufficiently high, this actuation method cannot be considered to develop the continuum robot which demands a fast response.

### 1.3.2 Pneumatic Actuation

Pneumatic system can also be utilized to actuate soft robotic arms. The pneumatic system for soft robot actuation tends to consist of multiple chambers arranged in a specified pattern. The pneumatic chambers can also be called artificial muscles which are connected to a set of pneumatic valves respectively in order to change the internal pressure. The pressure change inside the chamber can cause the length variation, which corresponds to the contraction or extension of the air muscle. The coordinated extension or contraction of a set of air muscles contributes to the bending direction and the bending angle of one segment of the compliant structure, as is shown in Figure 1-7 [12].

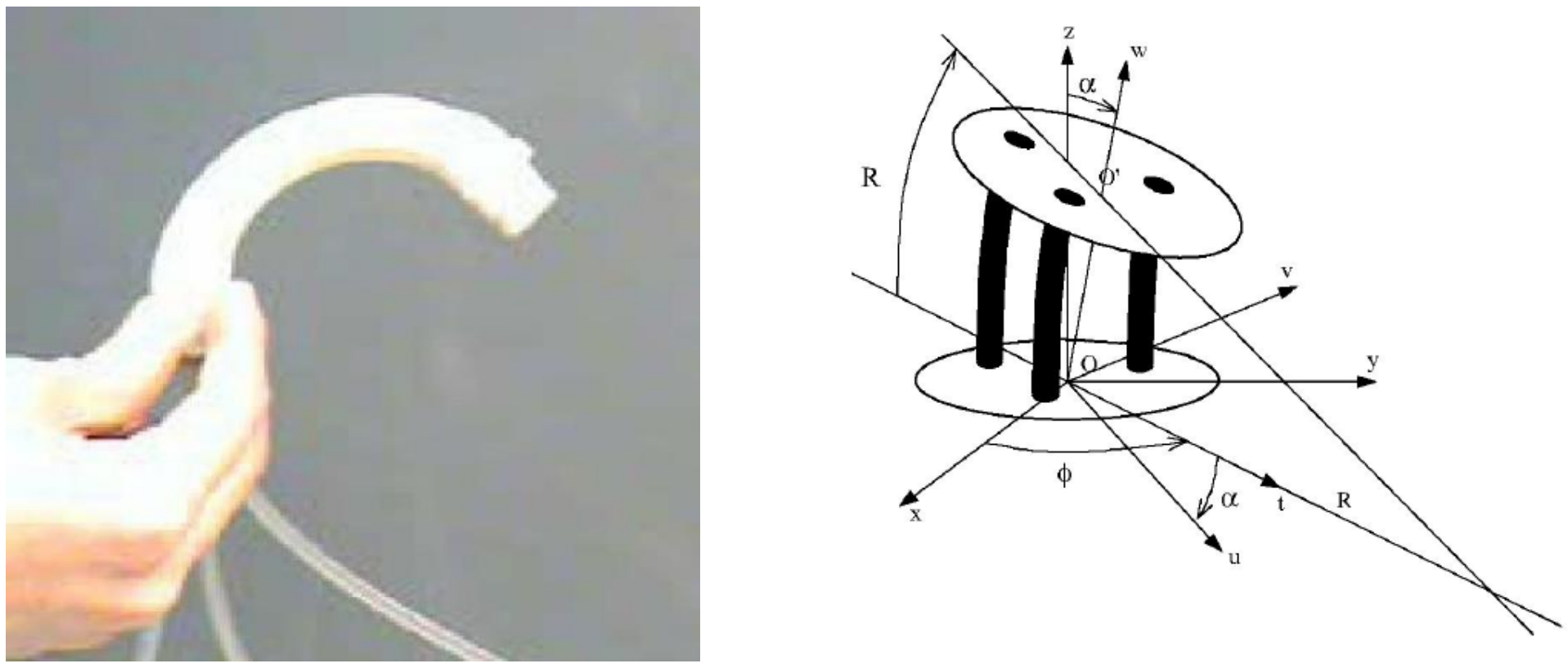

*Figure 1-7: A Pneumatic Actuated Robot and its Kinematics [12]*

Although the pneumatic chambers can be miniaturized in a continuum robot, it still has several limitations which conflicts with the variable stiffness requirements for this project. Pneumatic actuation means that the main body has to be made of silicone rubber, which has a quite small stiffness as a matter of fact, and the maximum stiffness after the addition of layer jamming system might be not sufficient to achieve load capacity demands. Hence, pneumatic actuation is not considered for the continuum robot to be designed.

### 1.3.3 Cable Actuation

Cable/tendon actuation is probably the most common method of driving continuum robots. As is shown in Figure 1-8, a cable driven continuum robot tends to have multiple spacer disks which are connected in series by a primary backbone located at the center axis [13] [14]. Distributed around the center backbone are several cables/tendons passing through the series of spacer disks, and the end of the cables are fixed to the end disk. As the cable is pulled by actuators, the cable length inside the continuum structure is decreased, thus forcing the structure to bend toward the side of the pulled cable. Through the coordinated displacement of each driving cable, the continuum structure can bend towards any specified direction with a user defined bending angle.

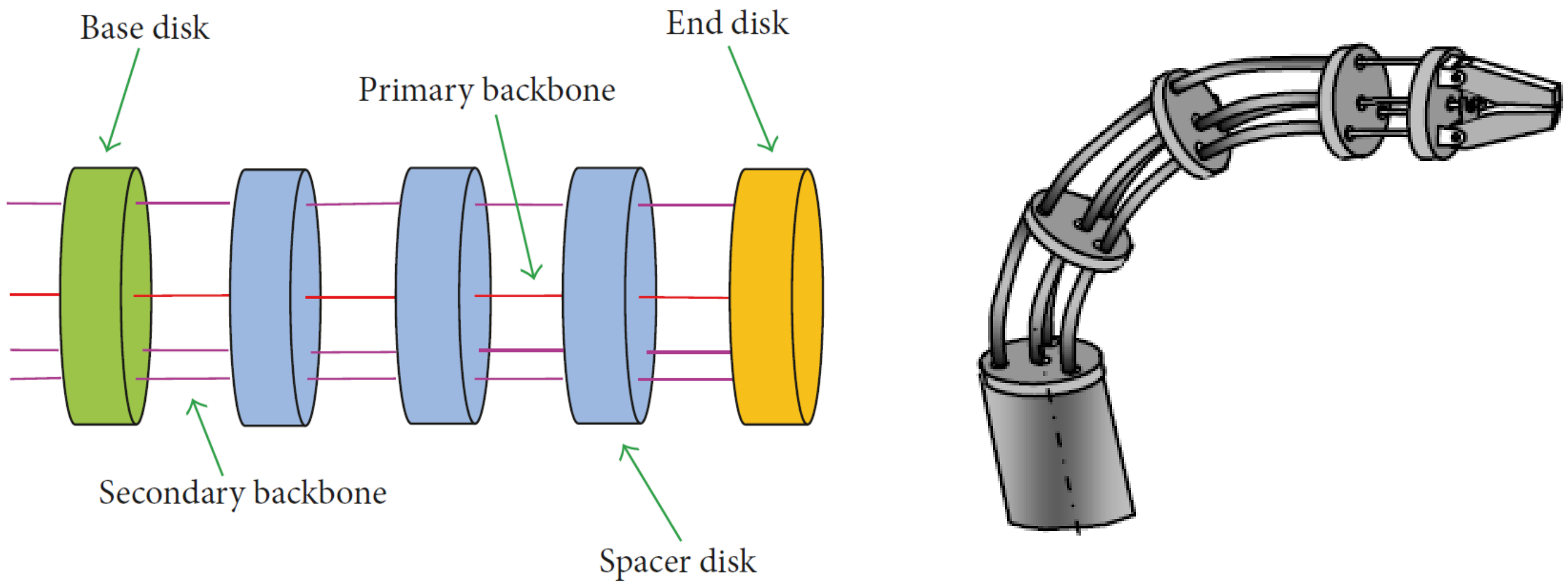

*Figure 1-8: Example of A Cable Driven Continuum Robot [13][14]*

The curvature and bending plane of each section can also be controlled through coordinated translation of the group of cables. The right side of Figure 1-8 shows a bended continuum robot after displacements of the 3 actuation cables. In this case, the position of the end of the continuum robot can be manipulated directly by changing the lengths of cables remaining inside the compliant structure, since the translation of the cables can be driven directly through electric motors. Once figuring out the kinematic relation between the cable length residing inside the compliant structure and the tip position, the shape and position of the robot arm can be precisely manipulated by controllers. Since the motion is directly driven by cables, the continuum robot can also achieve a pretty fast response once the motor shaft reaches the desired angular position. Based on these reasons, tendon actuation was decided as the approach of driving the continuum robot.

In addition, previous researches have already validated that cable actuation can be seamlessly integrated with layer jamming system. Amanov from Leibniz University developed a 2-segmented variable stiffness continuum robot which is actuated by 4 antagonistic tendon pairs [20]. The schematic of Amanov's robot system is available in Figure 1-9 as follows [20].

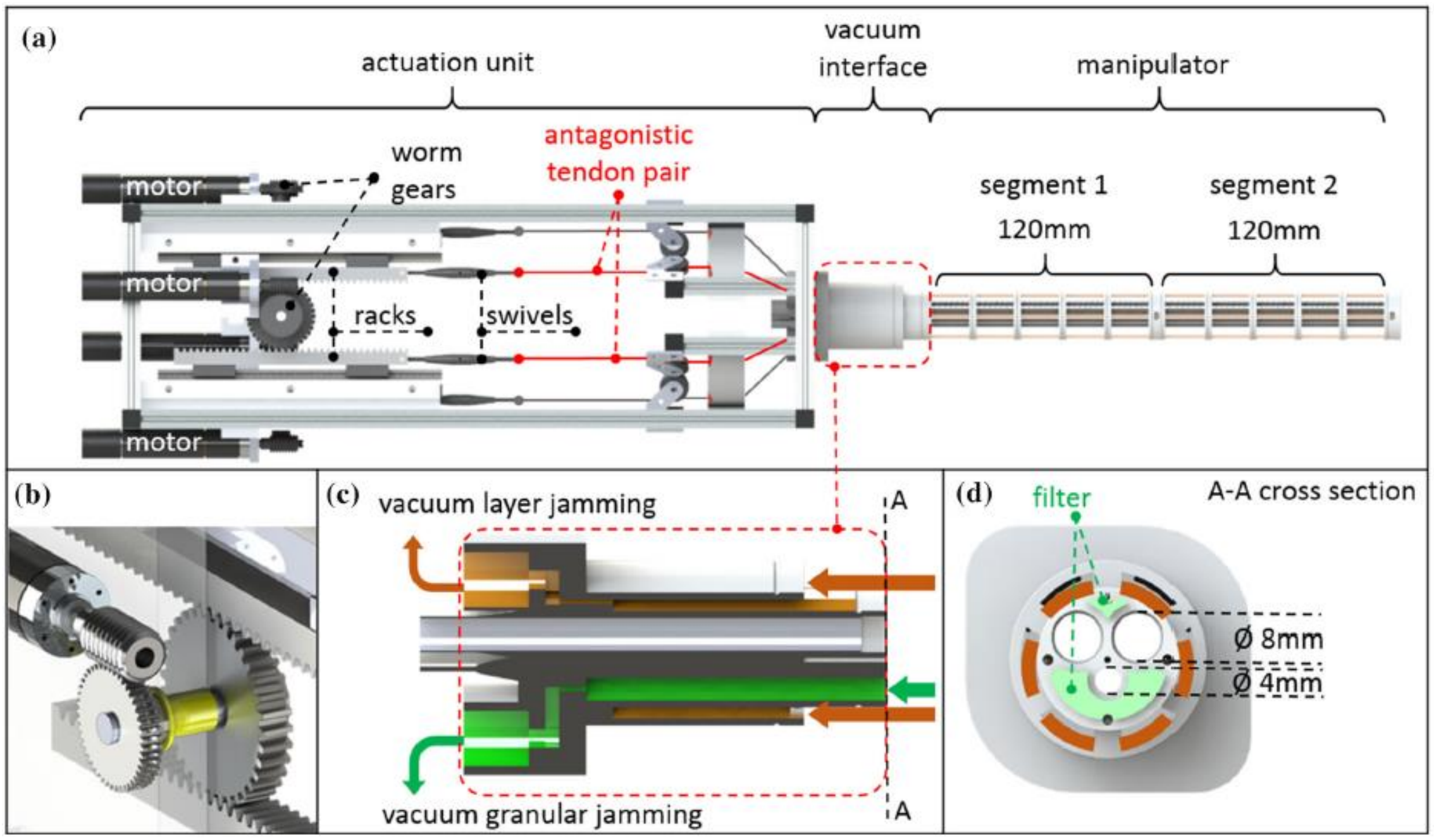

*Figure 1-9: Variable Stiffness Continuum Robot Developed by Amanov [20]*

Figure 1-9 shows that the continuum robot has 2 segments, with the end of each segment attached to 2 antagonistic tendon pair separately; each segment consists of 4 intermediate spacer disks. The stiffness of the robot arm can be varied through the combination of layer jamming and granular jamming. The robotic arm is manipulated by 4 antagonistic tendon pairs, with each pair actuated by the system of motor, worm bears and rack. Although the robot has a great load capacity of 4.42 N, there exist several flaws of the design. The first issue is related with the actuation system: since the tendon pair is pulled by the linear actuator, the rack is supposed to have sufficiently length to actuate each segment from the original position to the maximum bending angle. The result is the designed actuation unit has a great length which is even longer than the robotic arm. The second issue is concerned with the modularity of the robot arm design, since the robot arm length of this existing prototype cannot be adjusted. Because the 2 segments of the continuum robot arm are enclosed by a single vacuum bag, new segments cannot be added to the existing arm structure based on provisional user demands, and the distal segment cannot be removed from the robotic arm to reduce the arm length neither.

## 1.4 Objectives

The purpose of this research is to design, fabricate and test a tendon/cable driven variable stiffness robotic arm with layer jamming with modular structure. To solve the inherent issues with continuum robots, previous research has already explored the application of layer jamming to improve the rigidity and load carrying capacity of one arm segment which can achieve motion within a 2-dimensioal plane. Compared with previous studies, the robotic arm design of this research has a modular structure, which means the length of the robotic arm can be adjusted by addition of extra arm modules/segments to the existing robotic prototype, so that the variable stiffness robot can satisfy requirements of different operational environments. Another target of this research is that the actuation system design should be smaller and lighter, thus allowing the dock of the continuum robotics system to be mounted within a confined space, and the layout of the group of actuators shall be more compact likewise. In addition, the stiffness of the robotic arm is required to have the ability to be adjusted continuously under a wide range. The actuation time of the new continuum manipulator prototype is also supposed to be sufficiently low. Furthermore, the new arm prototype supports motion within a 3-dimensional space.

To achieve the goals, the author has already designed and fabricated a prototype of the variable stiffness robotic arm with compliant main structure and layer jamming mechanism. Design and fabrication of the connector has also been finished to integrate several link modules into one robotic arm with multiple segments. The actuator located at the base of the arm was designed and tested, in order to actuate and control the motion of each arm segment simultaneously. Finally, the load capacity test and actuation test of the prototype was be conducted to verify the load carrying capacity of the variable stiffness robotic arm.

## 1.5 Overview of Thesis

This thesis consists of five chapters. Chapter 1 covers the significance and application of variable stiffness robots and continuum robots, technical approaches to continuum robot actuation and variable stiffness control, and objectives of this research. Chapter 2 comprises of the detailed design and fabrication of components of the variable stiffness continuum robot developed in this project. Chapter 3 covers the analysis of the kinematic model of the 2-segmented continuum robot which reveals the mathematical relation between the tendon displacements and the motion of robotic arm. Chapter 4 covers the methodology and the results of a series of tests to verify the functionality of the variable stiffness continuum robot and to evaluate the performance of the prototype. Chapter 5 covers the summary and discussion of the future works to improve the design of the current variable stiffness robot prototype.

# Chapter 2 Detailed Design and Fabrication

In this chapter, first, the detailed design of a single segment of the robotic arm will be discussed. Second, the design of the connector will be covered since it is significant for satisfying the modular structure requirements of the variable stiffness robot arm. Then the design of the spool will be analyzed, and the actuator assembly will be presented. Motor selection will also be covered in this chapter. Next, the technical requirements and detailed design of the dock will be analyzed. Finally, electronics of the stepper motor control circuit will be discussed, and the final assembly of the multiple segmented variable stiffness robot arm will be presented.

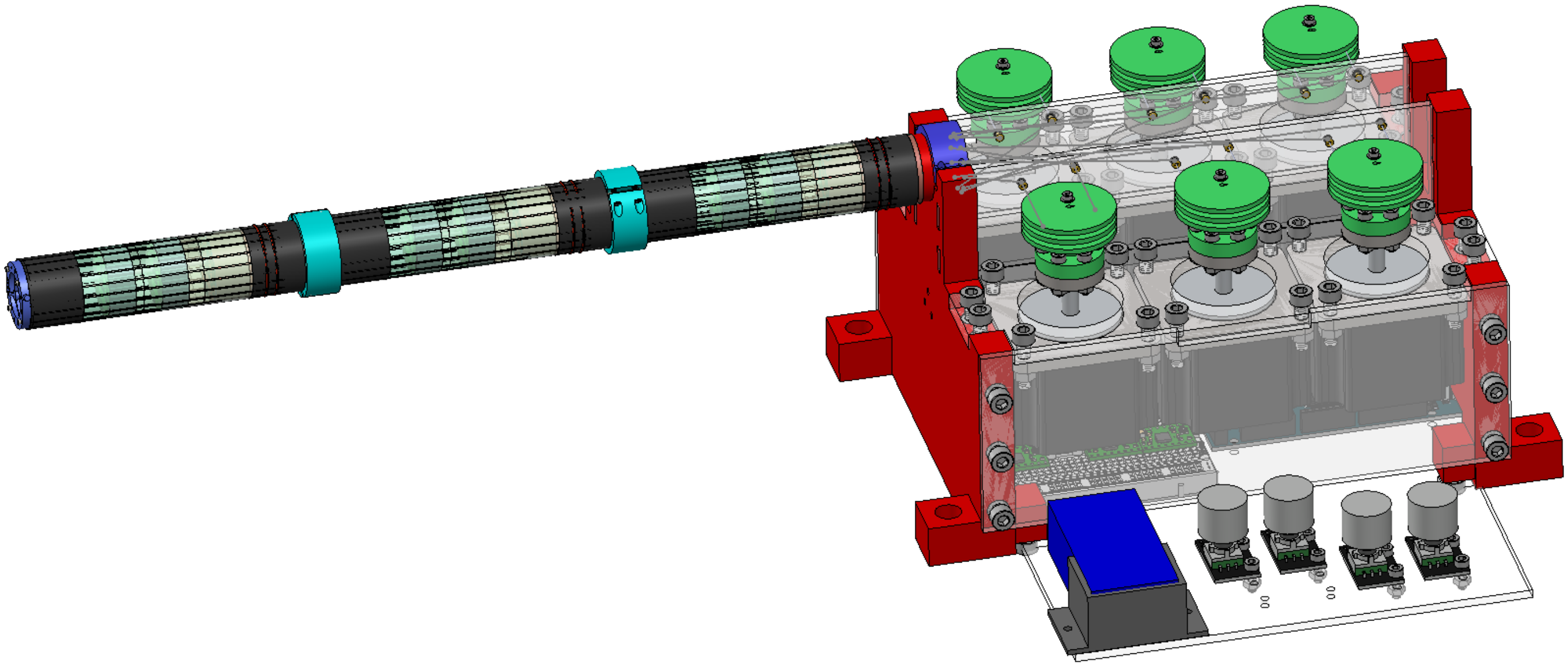

*Figure 2-1: Final Design of The Variable Stiffness Continuum Robot with Three Segments*

After the completion of the design for the robotic arm segments, connector, dock, spools and actuation control system, the CAD model of assembly for the variable stiffness continuum robot with 3 segments was created. The final design of robot prototype is shown in Figure 2-1 in the previous page.

As is shown in Figure 2-1, the continuum robot designed has 3 arm segments actuated by 6 stepper motors. The tendon pairs are attached to the spools which are secured on the motor shafts. The board with electronic components is placed at the bottom of the dock. The actuation system can be mounted to the table via 4 lugs at the bottom of the dock. For the tendon guidance mechanism, the 6 antagonistic tendon pairs are routed to the spools through tiny holes of 2 parallel guidance plates. Detailed design of the variable stiffness continuum robotic arm is described in the following sections.

## 2.1 Initial Design for a Single Segment of the Robotic Arm

For the initial design of the robot arm segment, it features tendon actuation and the variable stiffness is achieved through layer jamming. The first design of the robot arm is presented in Figure 2-2 of the following page.

As is presented on Figure 2-2, the main structure of variable stiffness arm segment consists of several intermediate spacer places and 2 end plates which are connected by the center backbone. The compliant structure is actuated by 2 antagonistic tendon pairs, which means that there are 4 cables in total attached to one arm segment. The tendons pass through each intermediate plate and end plate, and the ends of the tendons are attached to the distal side of the end plate. The layer jamming system is available on this design, as is shown on Figure 3-1, the compliant structure is covered by 2 concentric cylindrical membranes, of which both ends are glued and sealed to form a vacuum bag. The annular cavity inside the vacuum bag is utilized to accommodate the friction layers. One side of the bag is designed to be connected to the vacuum pump to introduce negative pressure and generate a sufficient amount of friction between the cylindrical friction layers, thus causing layer jamming effect and enhancing the stiffness of the robot

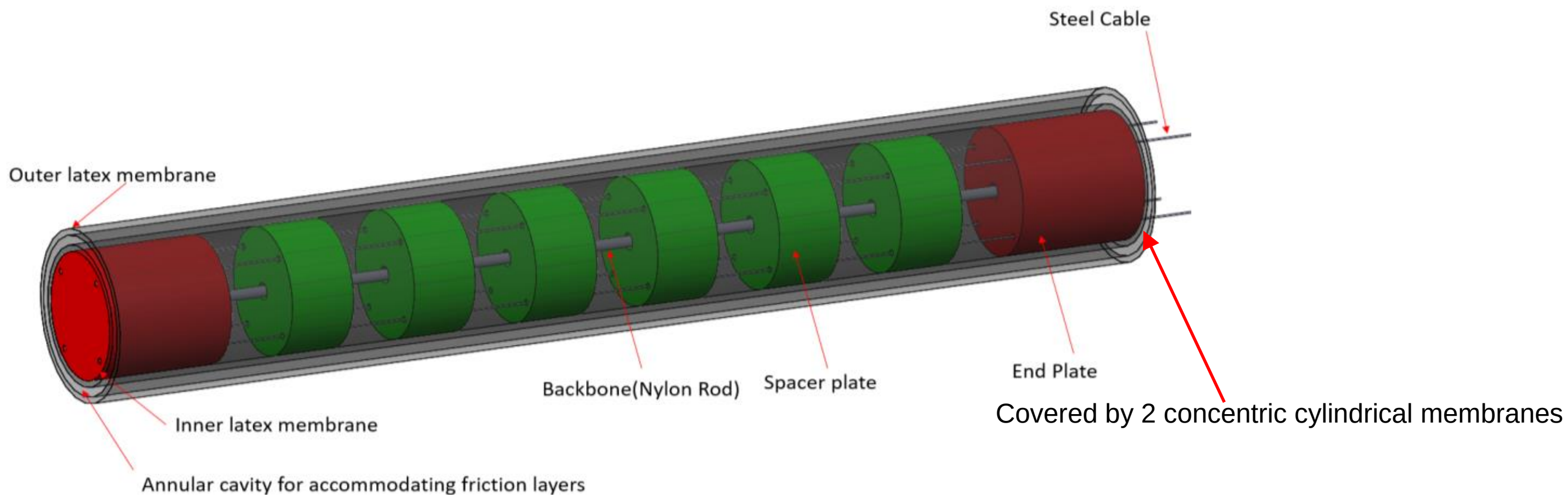

*Figure 2-2: Initial design of the variable stiffness arm segment*

As is presented on Figure 2-1, the main structure of variable stiffness arm segment consists of several intermediate spacer places and 2 end plates which are connected by the center backbone. The compliant structure is actuated by 2 antagonistic tendon pairs, which means that there are 4 cables in total attached to one arm segment. The tendons pass through each intermediate plate and end plate, and the ends of the tendons are attached to the distal side of the end plate. The layer jamming system is available on this design, as is shown on Figure 2-2, the compliant structure is covered by 2 concentric cylindrical membranes, of which both ends are glued and sealed to form a vacuum bag. The annular cavity inside the vacuum bag is utilized to accommodate the friction layers. One side of the bag is designed to be connected to the vacuum pump to introduce negative pressure and generate a sufficient amount of friction between the cylindrical friction layers, thus causing layer jamming effect and enhancing the stiffness of the robot arm segment.

Although the concept of the design is valid. It has some issues with the layout of vacuum tubing; and the vacuum bag is not securely fixed on the compliant main body neither, which might abate the stiffness improvement after the introduction of negative pressure inside the bag. As the vacuum bag is loosely attached to the main structure, the layer jamming system is not well integrated into the robotic arm segment. This means that the compliant main structure cannot be locked even when the friction layers are locked after starting the vacuum pump and establishing a negative pressure. Faced with these flows with the original design, a new design will be proposed and presented to resolve these issues in the next section.

## 2.2 Final Design of a Single Segment of The Robotic Arm

To securely attach the vacuum bag to the compliant structure of the robotic arm segment and ensure that the vacuum pressure inside the cylindrical bag can be established, the final robot arm segment design is contemplated and presented in Figure 2-3 below.

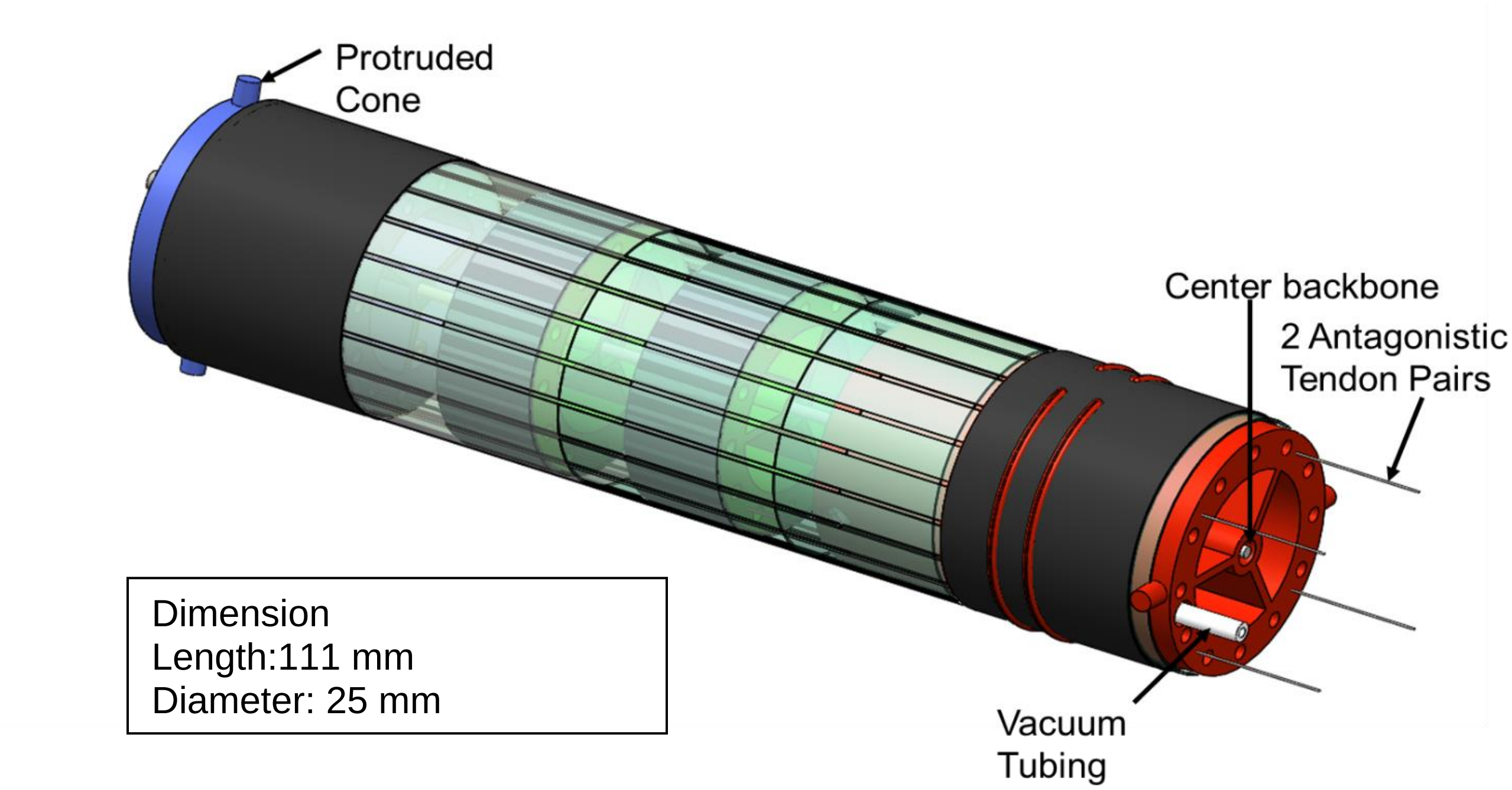

*Figure 2-3: Final design of one single segment of the robotic arm*

Compared with the initial design, the final design of the robot arm segment utilizes the same strategy for actuation and variable stiffness control. As is shown on Figure 2-3, the arm segment is actuated by 2 antagonistic tendon pairs which the orientation of 90 degrees away from each other. The vacuum tubing is also added to the structure so that the bag wrapping the compliant body can be connected to the vacuum pump. To satisfy the modular design requirements for each segment of the robotic arm, features of protruded cones are added to both ends of the arm segment, in order to make sure that both side of the arm segment can mate with the connector precisely and the connection between adjacent arm segments is sufficiently secure. In addition, the number of intermediate spacer plates is reduced from 6 to 2, compared with the initial design, to save time for fabrication of each arm segment.

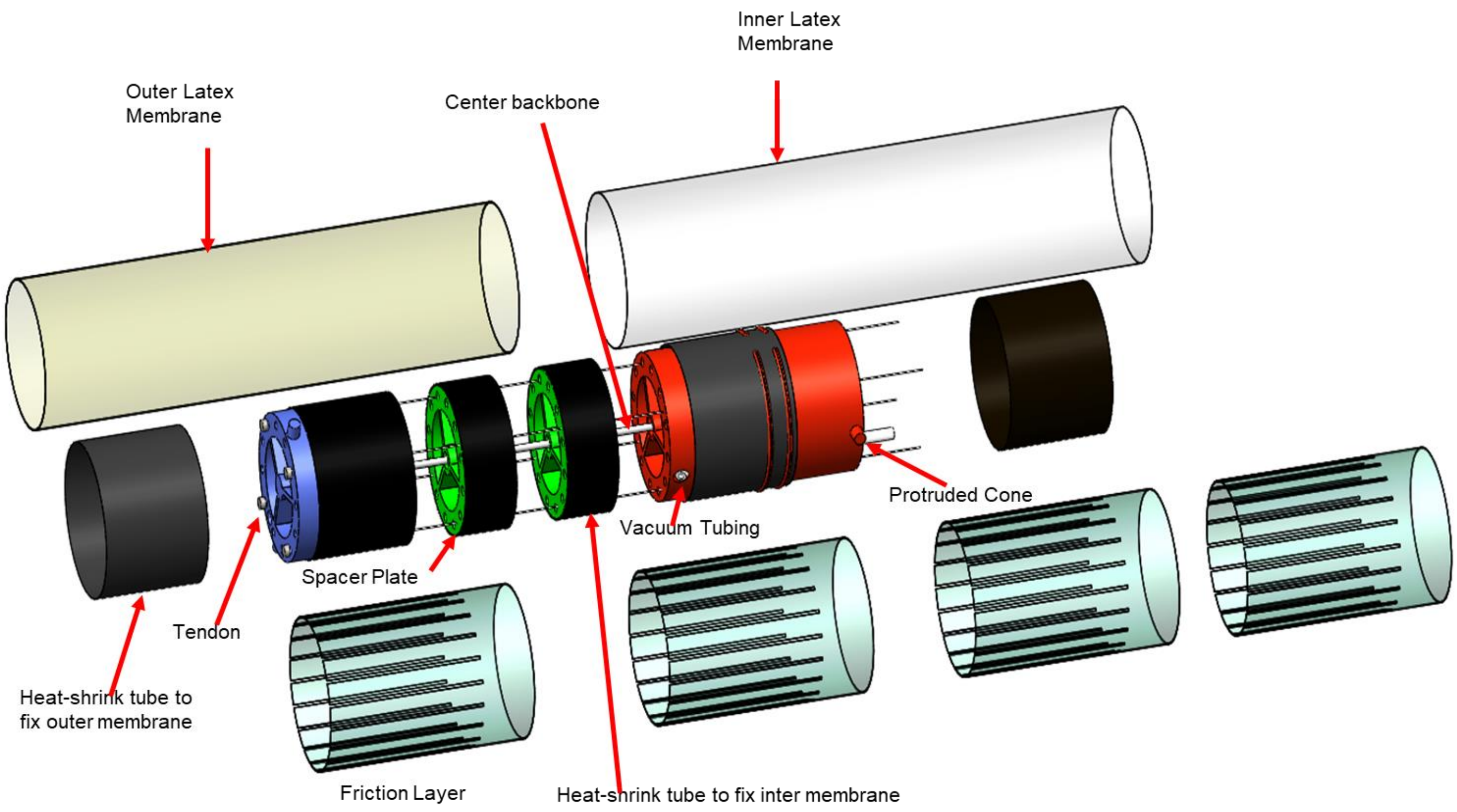

*Figure 2-4: Exploded View of One Segment of the Variable Stiffness Robotic Arm*

Figure 2-4 above shows the exploded view of one segment of the variable stiffness robotic arm, and Figure 2-5 in the following page shows the compliant structure of the arm segment. As is shown on Figure 2-4, the spacer plates and the end placed are 3D printed with the material of polylactic acid (PLA), and the inner membrane and the outer membrane are made of latex rubber.

The fabrication process of the variable stiffness robot segment is explained as follows. To fabricate the vacuum bag, latex sheets were cut into rectangular shape with desired dimensions, then the opposite sides of the rectangular sheet were glued together to form the cylindrical membrane. To firmly affix the vacuum bag on the compliant structure, the heat-shrink tube is introduced. After the assembly of the compliant structure, the tendons were threaded through the spacer plates and the ends of the tendons actuating the motion of the segment were glued to the distal spacer plate. Then the compliant structure was wrapped by the inner membrane of the bag, and the outer surface of the inner membranes would be enclosed by the heat shrink tube. The heat-shrink tube was then heated up and shrink under a heat gun, thus fixing the inner membrane securely on the compliant structure. Next, the cylindrical friction layers were glued to the outer cylindrical surface of the heat-shrink tubes which adhere firmly onto the compliant structure. Then the unfinished assembly was wrapped by the outer membrane, and both ends of the inner membrane and the outer membrane were glued together to form a sealed bag. Finally, 2 segments of the heat-shrink tube were used to wrap up both ends of the cylindrical bag. The heat-shrink tubes were then heated to fix the outer membrane of the cylindrical bag on both ends of the arm segment.

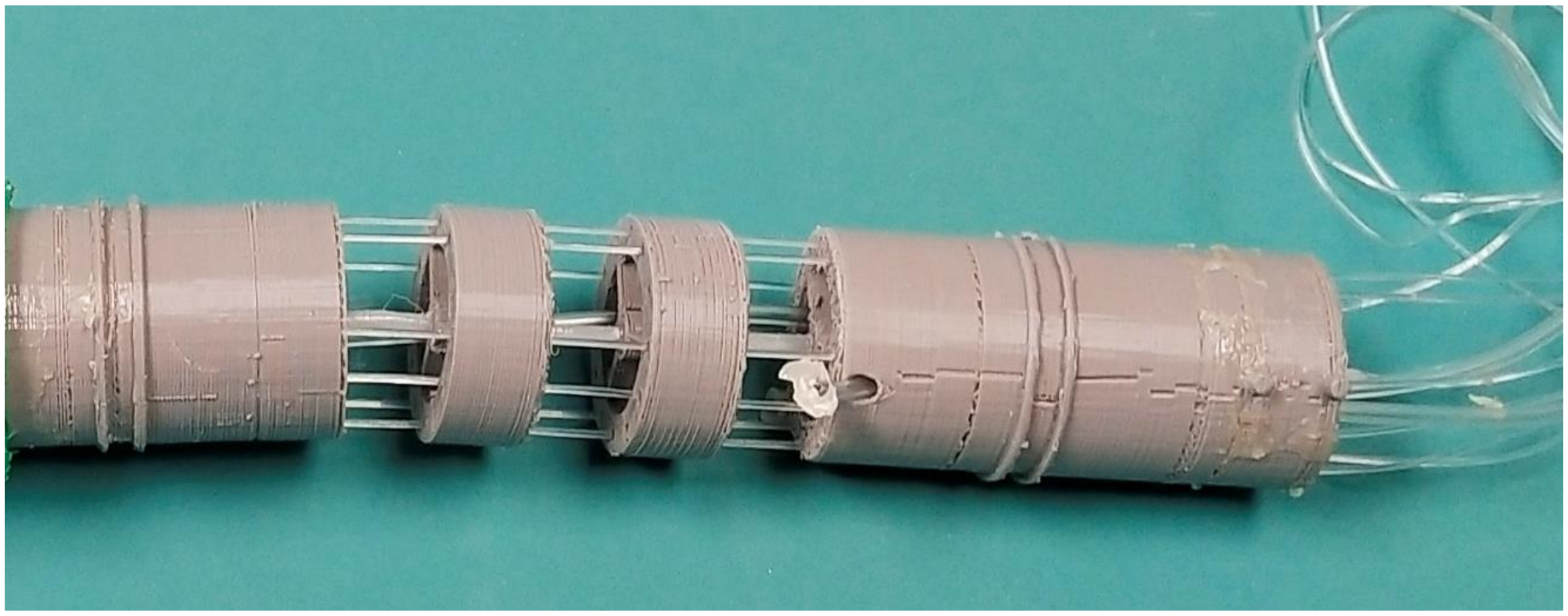

*Figure 2-5: Compliant Structure of One Segment of the Robot Arm*

The layout of the friction layers for each arm segment is illustrated in Figure 2-6 as follows. As is shown on Figure 2-6, the layer jamming mechanism has 2 friction layers in total, and there is only one surface of friction between the adjacent overlapped flaps. The friction layers are well arranged so that the number of friction layers has a constant value of 2 along the longitudinal direction of the robot arm segment. For detailed dimension of the layer jamming system, the distance between the adjacent spacers disks is 10 mm, and the thickness of the 2 intermediate spacer disks is also 10 mm; the diameter of the disks of the compliant structure is 25 mm for this final design. In the future, the number of friction layers should be increased and the spacing between the disk should be decreased appropriately to improve the stiffness ratio of the robotic arm, which is the ratio between the stiffness of the robot arm after the introduction of negative pressure and before the engagement of the vacuum pump.

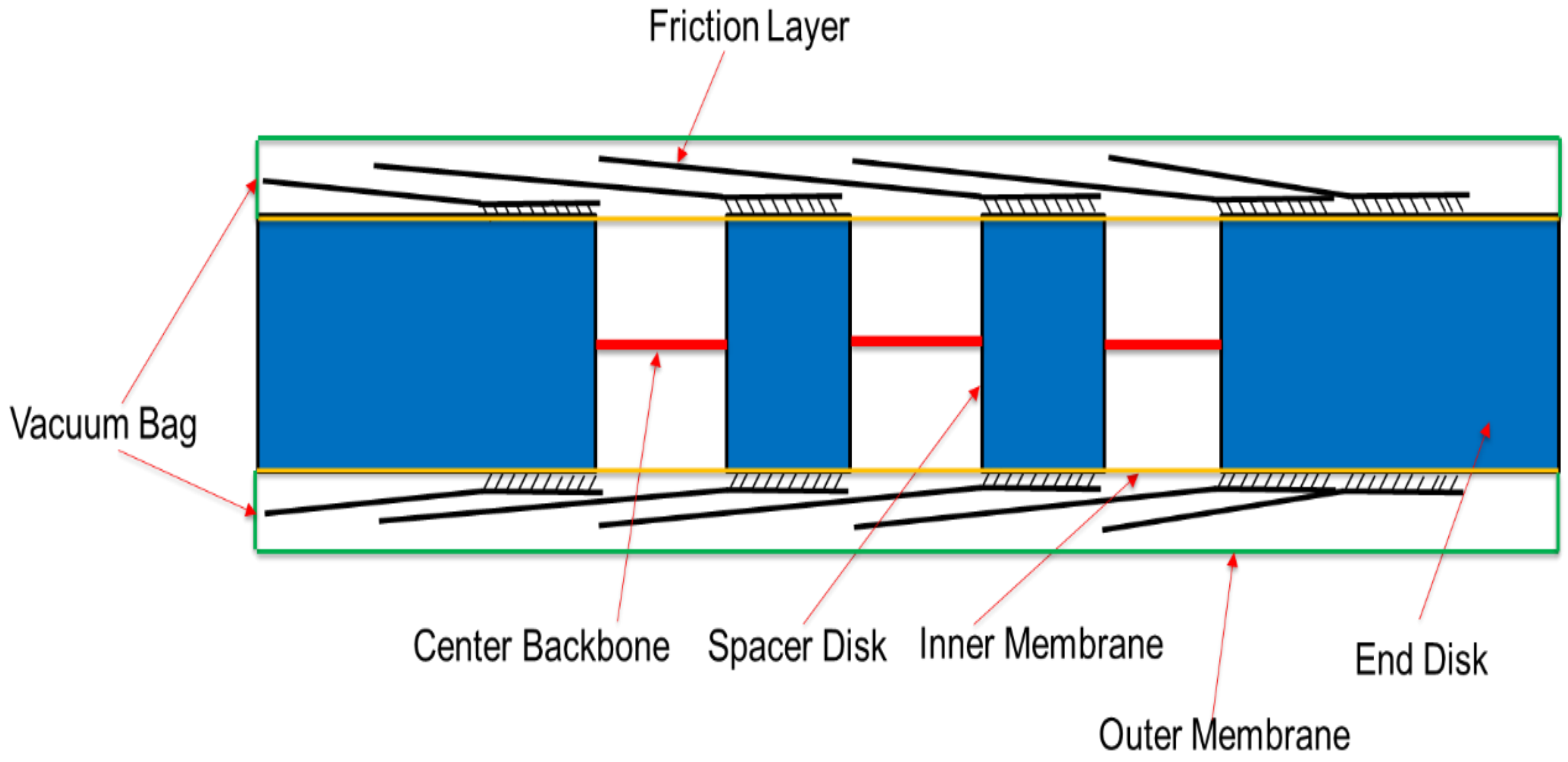

*Figure 2-6: Layout of The Friction Layers of One Arm Segment*

## 2.3 Connector Design

One of the design requirements is that the variable stiffness robot should have a modular structure, to ensure that the length of the robotic arm can be adjusted by addition of extra arm modules/segments to the existing robot arm prototype, thus satisfying the needs for a different arm length in different operation environment. To satisfy this design requirement, the connector design is presented in Figure 2-7 in the following page. The function of the connector is to mate with 2 adjacent robot arm segments precisely and form a secure joint between them. As is shown on Figure 2-7, the connector has 2 halves, the upper half and the lower half. Both of the 2 halves of the connector feature 2 tapered holes, which are used to mate with the protruded cones on the end disks of the adjacent segments of a robotic arm. The lower half of the connector is embedded with heat-set inserts to ensure that the upper half and the lower half of the connector can be screwed together and form a solid connection between the contiguous arm segments.

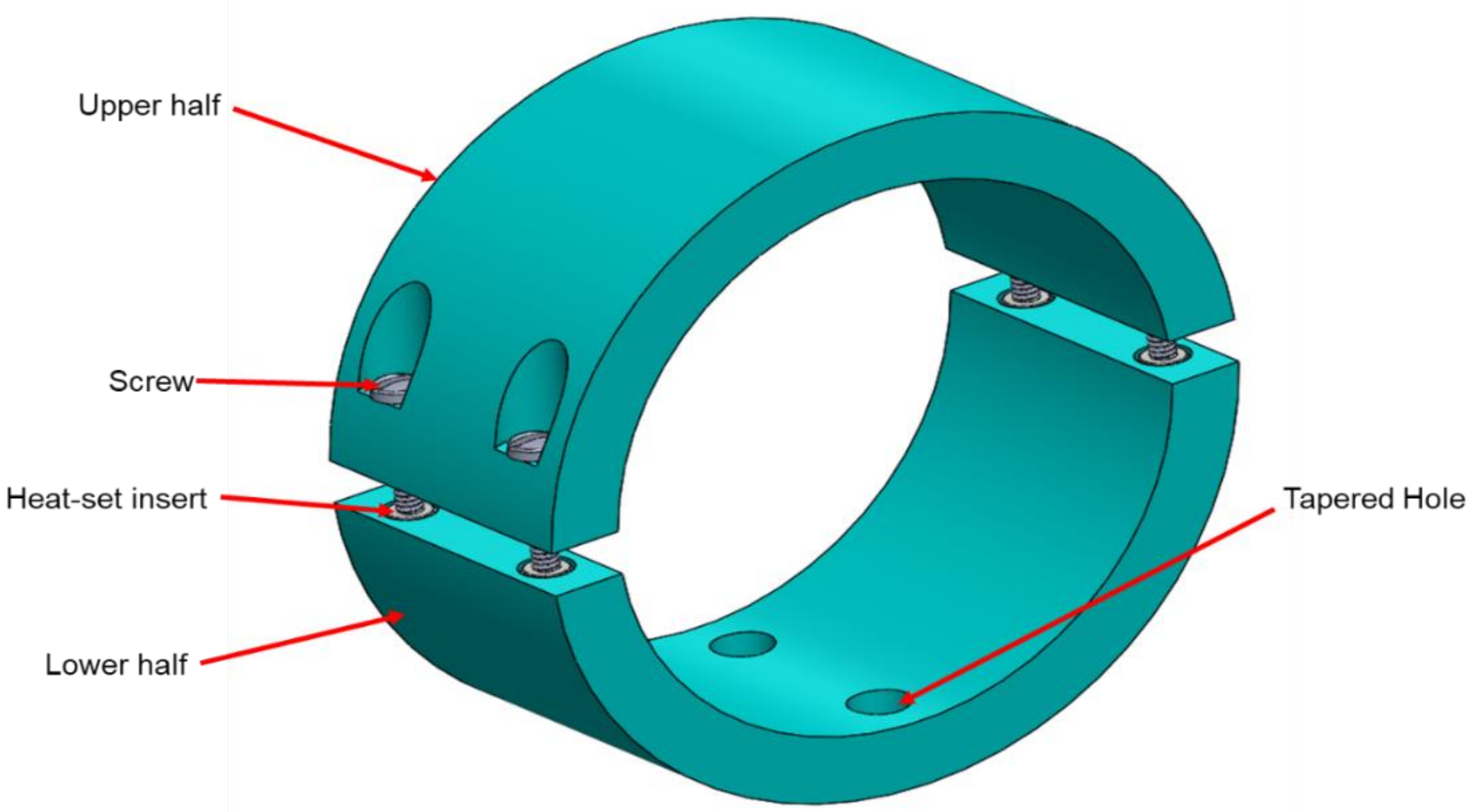

*Figure 2-7: Design of The Connector of Multiple Segmented Robotic Arm*

Since the connector and arm segment have already been designed, the CAD model of 2-segmented arm assembly is established to examine the modular design concept of the variable stiffness robotic arm. The design of the 2-segmented variable stiffness arm is shown in Figure 327 of the next page.

As is presented in Figure 2-8, the variable stiffness robotic arm has 2 segments connected in series, including segment 1 and segment 2, by the connector. The 2 segmented arm is actuated by 4 tendon pairs, with each segment actuated by 2 orthogonal tendon pairs. Since the robot arm has two segments, 2 vacuums tubes are required to connect the vacuum pump to the 2 segments separately. The proximal end of the arm will be fixed to the dock and connected with the actuation system which will be discussed later in this chapter.

Figure 2-9 in the next page shows the fabricated variable stiffness robotic arm with 2 segments. After the fabrication was finished, the vacuum tubing of the robotic arm was connected to the vacuum pump to take a qualitative test for gas tightness of the layer jamming system. The test shows that the vacuum bags have no gas leakage and the rigidity of the structure had a considerable improvement under the introduction of negative air pressure.

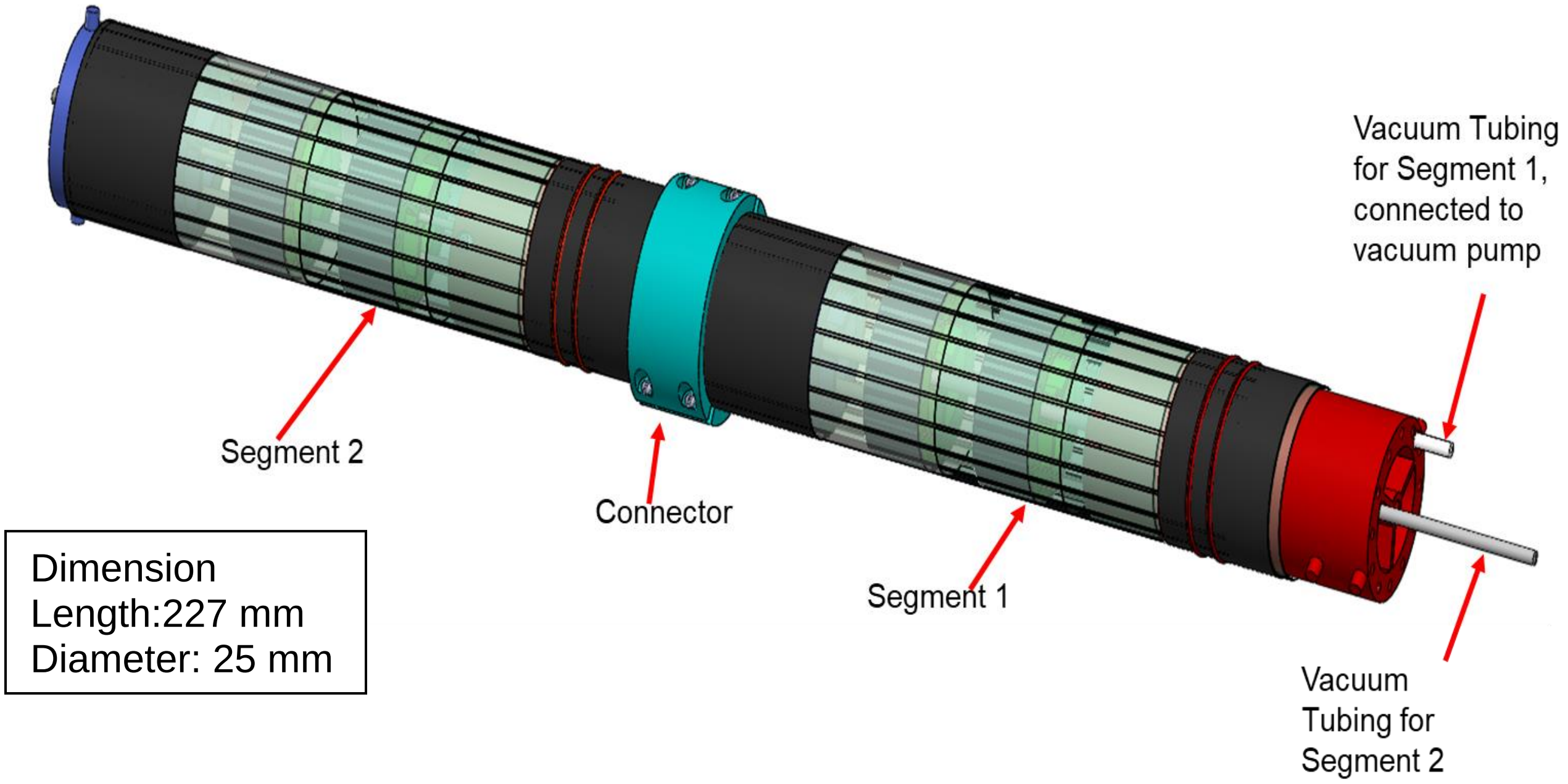

*Figure 2-8: Assembly of The Robotic Arm with Two Segments*

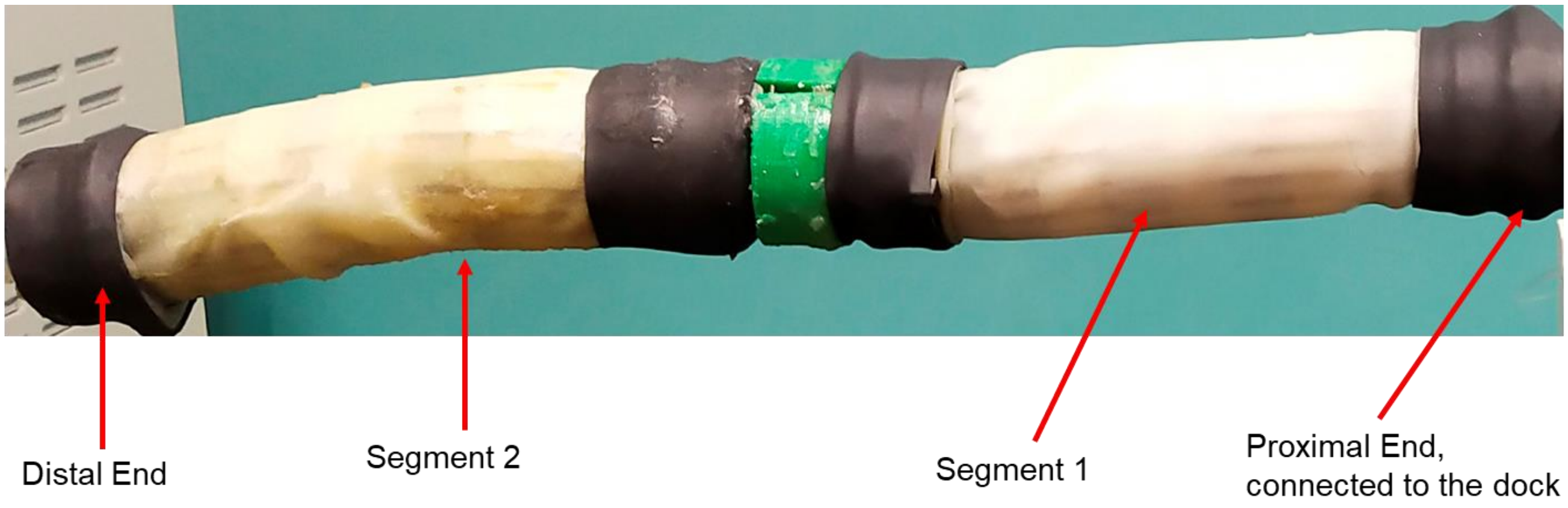

*Figure 2-9: The Variable Stiffness Robotic Arm with Two Segments Fabricated in Lab*

## 2.4 Spool Design

The function of the spool is to convert the motor shaft rotation to linear translation of the tendons driving the continuum robotic arm. One spool is supposed to actuate one pair of antagonistic tendons. Since each arm segment is actuated by 2 antagonistic tendon pairs, 2 spools are required to drive each segment of the robotic arm. Since one spool is required to drive one tendon pair, the spool should have features which can fix the ends of 2 antagonistic tendons. To satisfy these requirements, the spool design was contemplated and illustrated in Figure 2-10 as follows.

As is shown on Figure 2-10, the spool is made of PLA by 3D printing. To drive 2 antagonistic tendons of the tendon pair, 2 annular grooves were added this design. To fix the tendon ends to the spool, channels inside the spool were designed to guide the tendons to the top of the spool, where the tendon ends can be fixed. One tendon is winded around the cylindrical surface of the lower groove, with the end navigating through the channel inside the groove to the top of the spool. The other tendon is winded around the cylindrical surface of the upper spool with the end going all the way up to the spool top. Then both ends of the tendon pair can be fixed on the spool by the screw and washer. The spool can be bolted to the aluminum mounting hub, which can then be securely connected with the motor shaft. The spool has a radius of 10 mm for this design of the actuation system. Figure 2-11 on the next page shows the spool assembly which is fabricated in DISL.

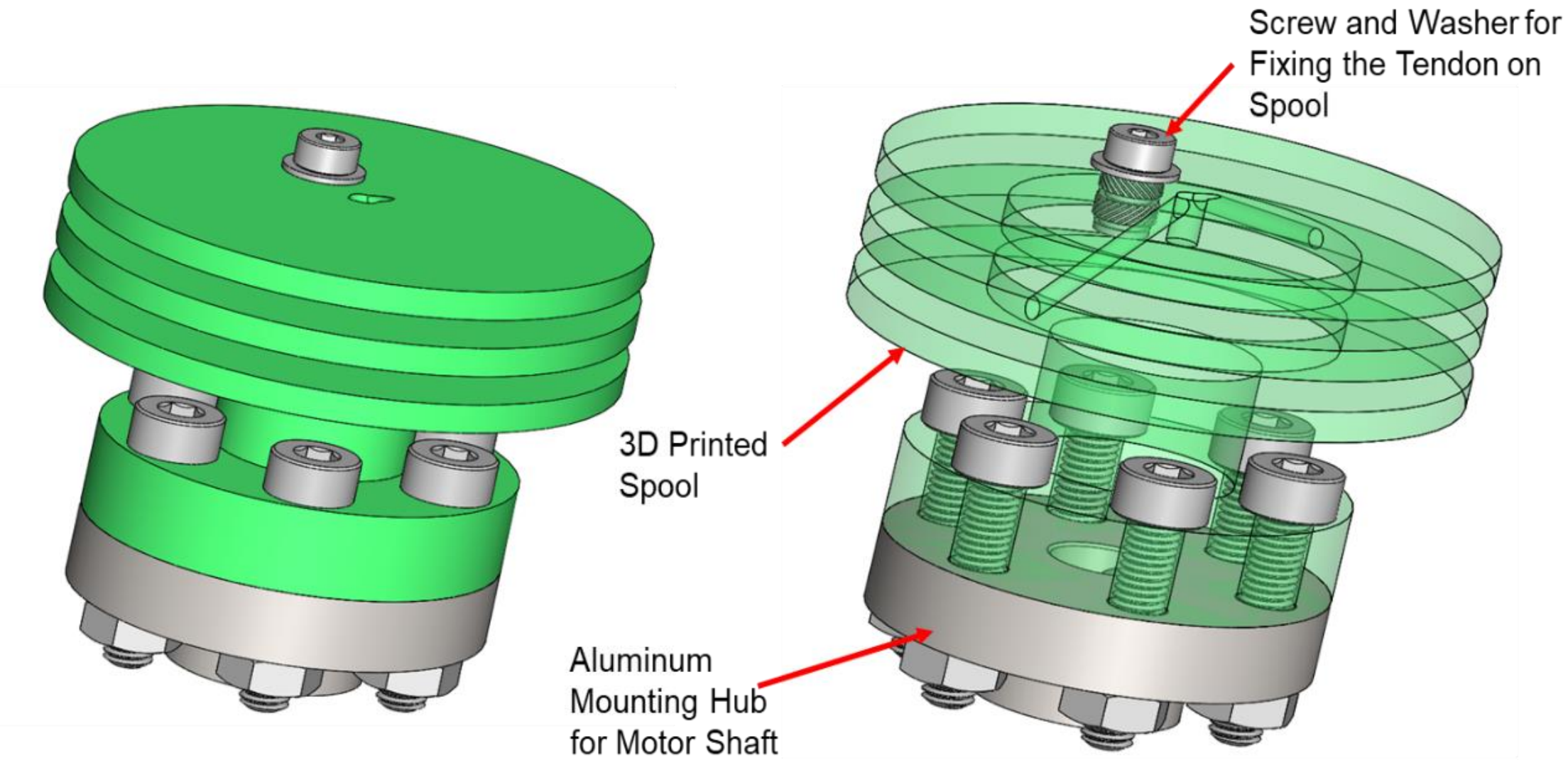

*Figure 2-10: Design of The Spool for The Actuation System*

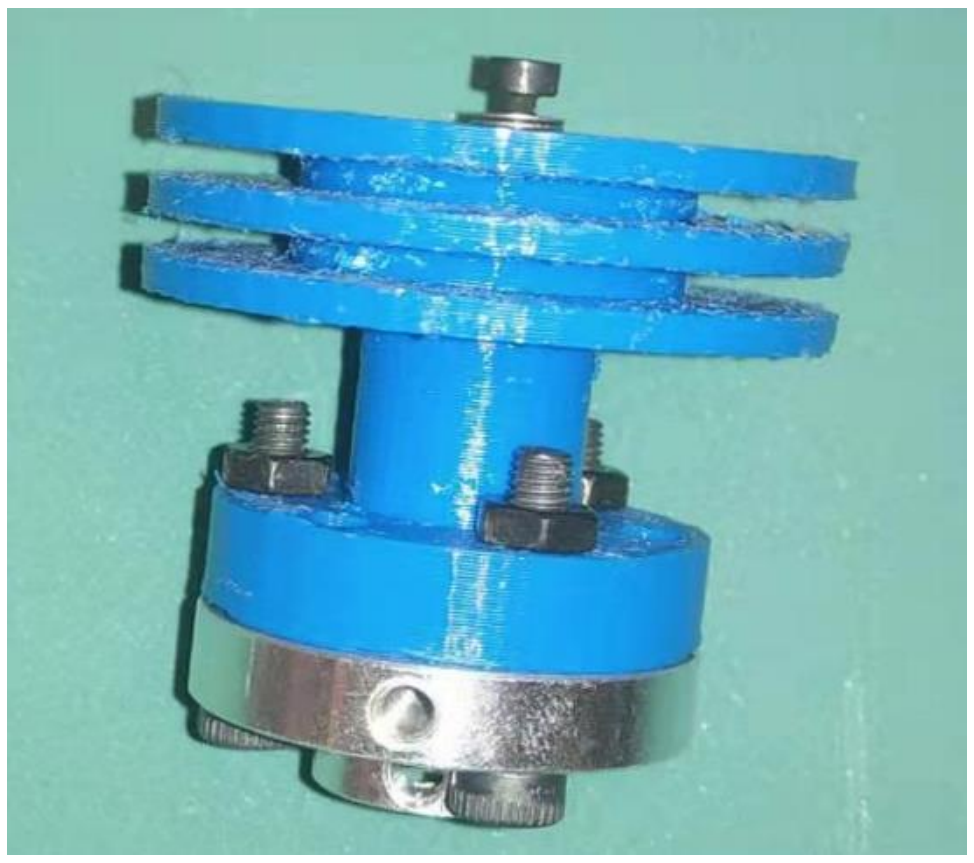

*Figure 2-11: Spool Assembly Fabricated in DISL*

## 2.5 Selection of the Actuation Motor

It is necessary to find a motor which has sufficient power and torque to pull the antagonistic tendon pairs and drive segments of the robotic arm separately.

Assume that the maximum tension on tendon is 3 kg, we have

Maximum load carried by the tendon:

$$F_{max} = 30\ N$$

We also have the radius of the spool:

$$r=10 \text{ mm}$$

Torque required to drive the tendon at the maximum load capacity:

$$T_{design} = F_{max}\, r$$

$$T_{design} = 300\ N - mm$$

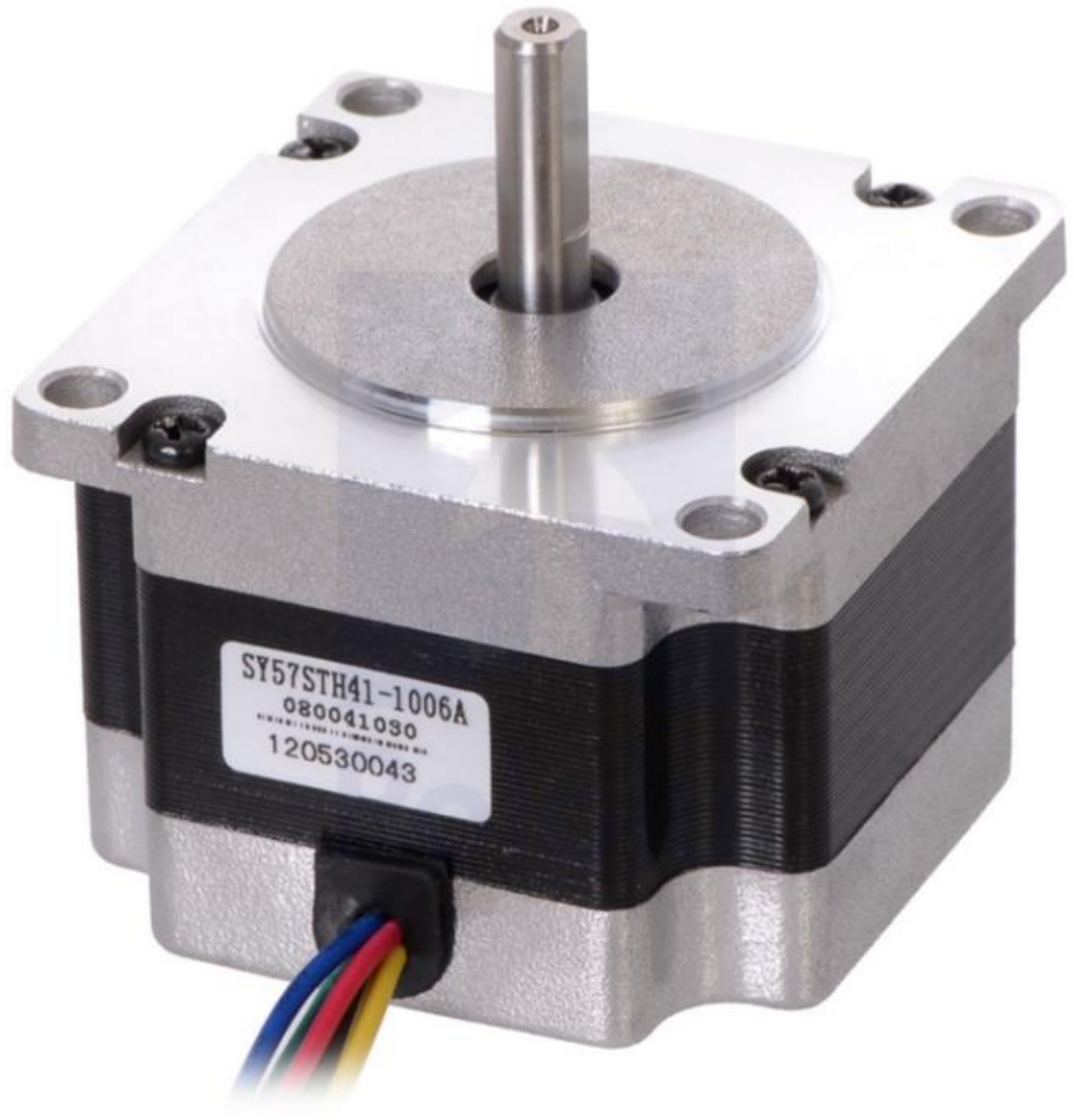

*Figure 2-12: NEMA Size 23 Stepper Motor for Actuating the Variable Stiffness Robotic Arm [15]*

For a NEMA Size 23 stepper motor, it can output a holding torque of 392 N-mm. Since the holding torque of the motor is greater than then maximum load on spool, it has enough torque to actuate the tendons. Hence, NEMA size 23 stepper motor is chosen to actuate each segments of the variable stiffness robotic arm. The motor has the step angle of 1.8 degree, which corresponds to 200 steps per revolution. The maximum allowable current per phase is 1 A for the motor. Figure 2-11 [15] below shows the NEMA size 23 motor purchased for building actuation system for robot arm.

## 2.6 Actuator Design

The actuator was designed after determining the model of the stepper motor to drive the antagonistic tendon pairs of the robotic arm segment and figuring out the geometry. The actuator assembly is presented in Figure 2-13 as follows. As is shown on Figure 2-13, the spool is fixed on the motor indirectly through the aluminum mounting hub. The mounting hub is compatible with the 6.35 mm diameter of the stepper motor shaft. The spool is fixed on the mounting hub through several M3 bolts and nuts. Figure 2-14 is a picture showing the purchased aluminum mounting hubs and corresponding tools to install the hub on the motor shaft [16]. The mounting hub is secured to the motor shaft through 2 set screws.

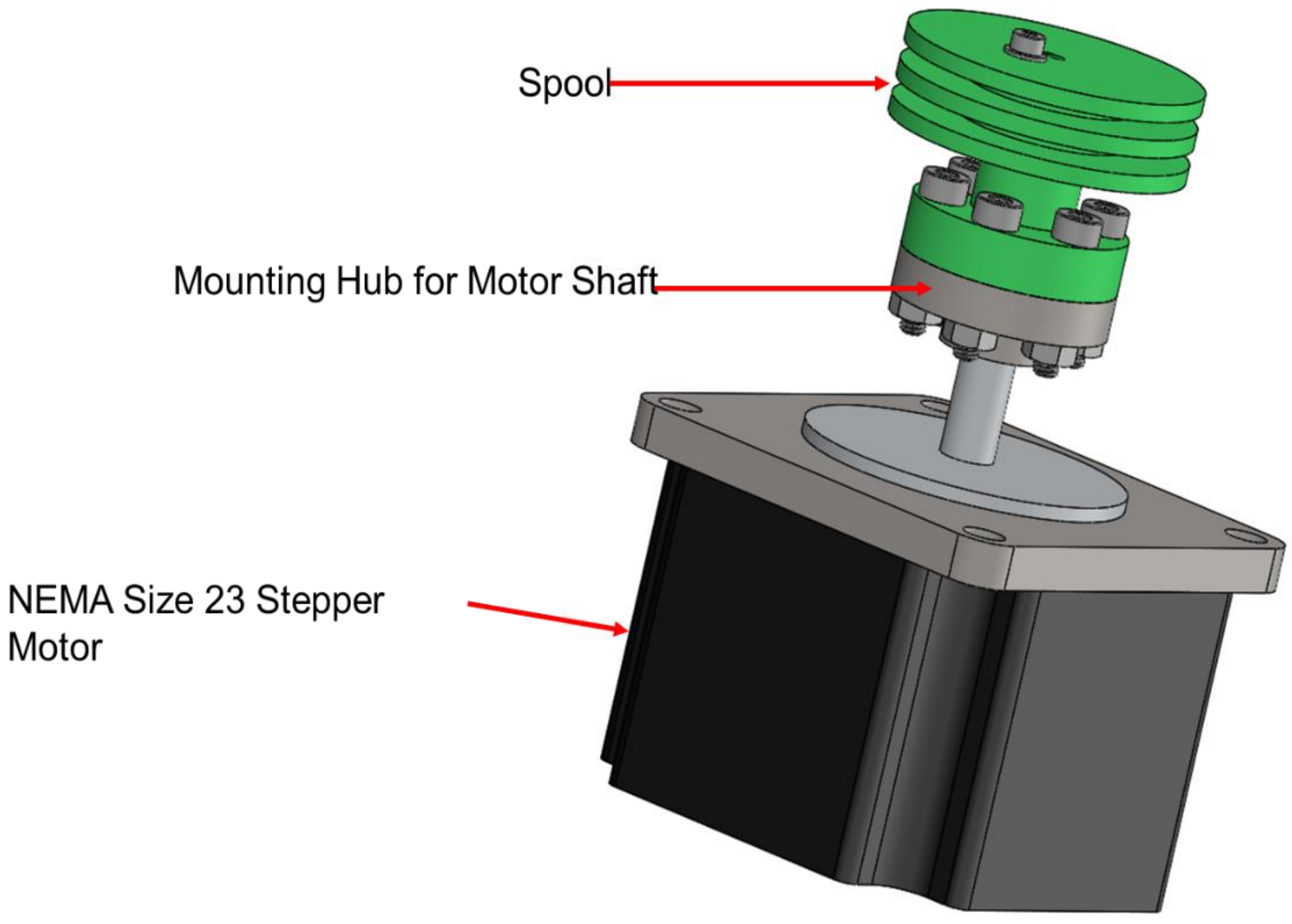

*Figure 2-13: One Actuator of the Variable Stiffness Robotic Arm*

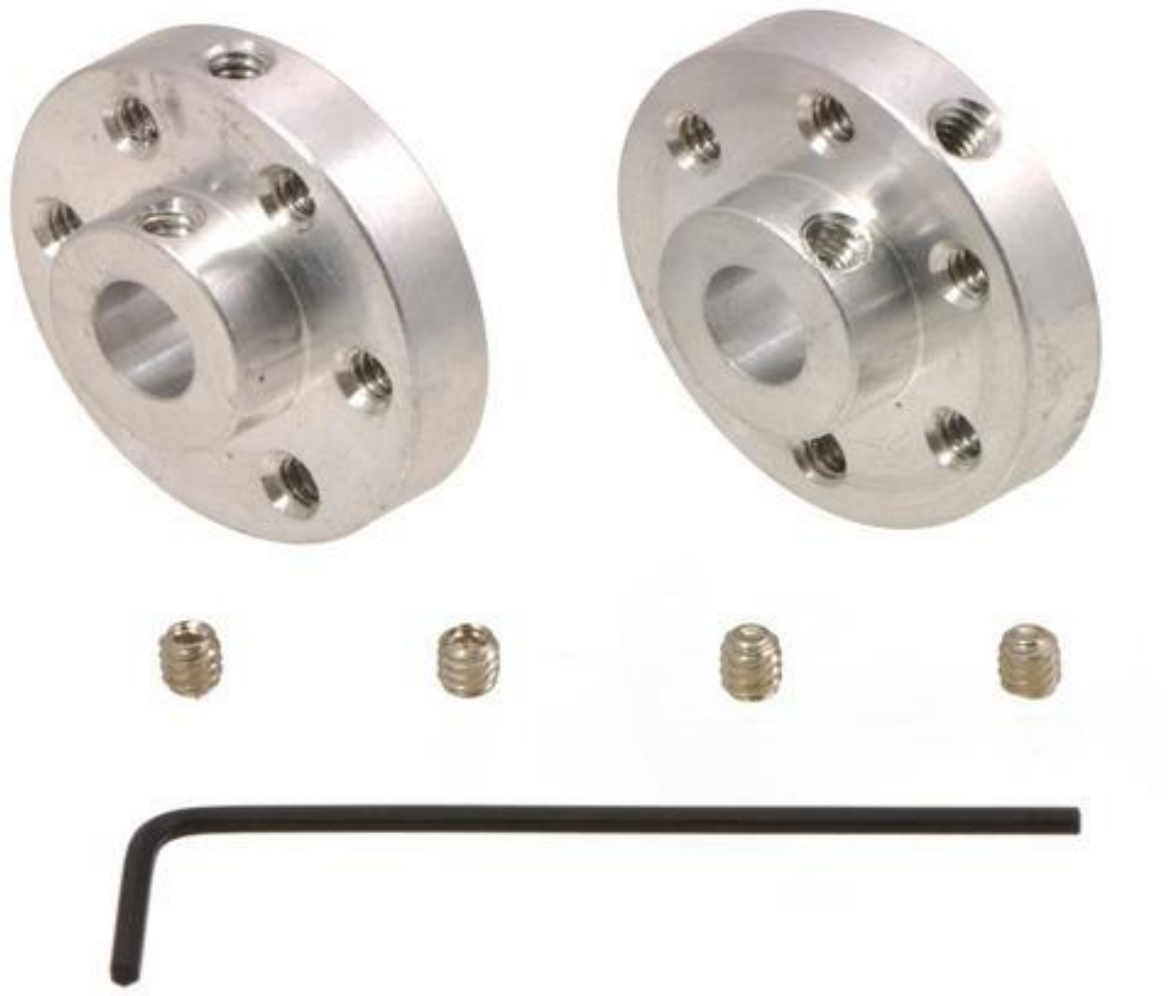

*Figure 2-14: Aluminum Mounting Hub and Corresponding Tools Required to Fix the Hub on The Motor Shaft [16]*

## 2.7 Dock Design

The function of the dock is to provide a housing to accommodate multiple actuators driving several segments of the variable stiffness continuum robotic arm. The dock is also the main frame of the robotic system, and proximal end of the robotic arm should be fixed to the front of the dock. There are several functional requirements for the dock design. First, the dock is supposed to have a good compactness, which means that the volume of the dock should be miniaturized to save space, while still capable of accommodating as much actuators as possible. Second, the dock should also support easy assembly and maintenance. This means that the dock should be well designed so that users can assemble the dock quickly, and that the broken parts of the dock can also be removed and replaced quickly. The third requirement is that the dock should allow users to install actuators quickly without difficulties. The dock is also supposed to have a light weight while still having sufficient rigidity to withstand reactional torsional loads from the actuators.

The design of the dock for the variable stiffness robot was completed by following the technical requirements mentioned above. The final design of the dock is presented in Figure 2-15 as follows.

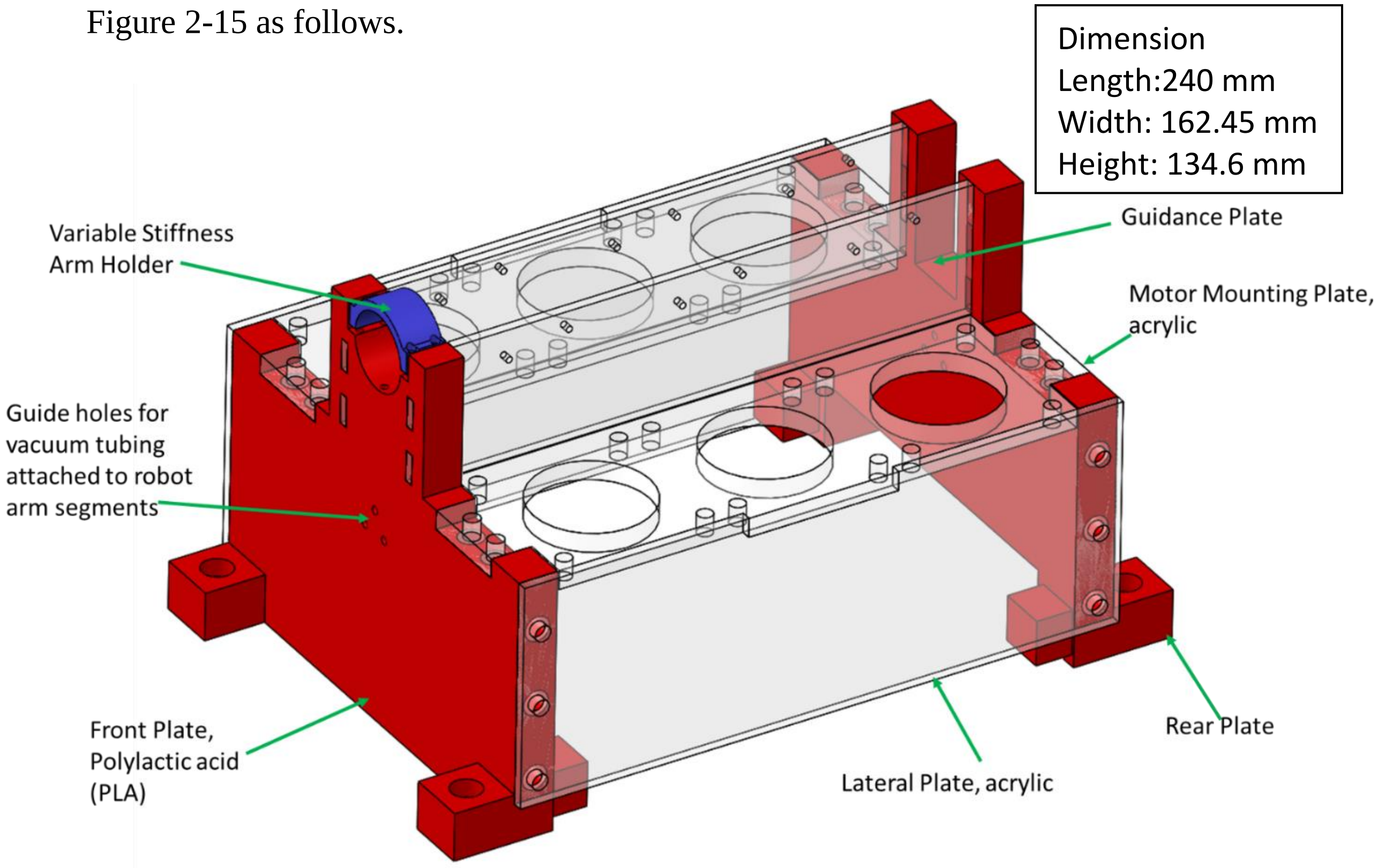

*Figure 2-15: Design of The Dock for the Variable Stiffness Robot System*

The dock has several types of components including front plate, lateral plate, rear plate, lateral plate, motor mounting plate, guidance plate and the robot arm holder. Among all components of the dock, the front plate, rear plate and the arm holder are 3D printed with the material of PLA; the guidance plates and the lateral plates are laser cut parts with the raw material of 1/8 in thickness acrylic sheets; the motor mounting plates are also laser cut parts with the raw material of acrylic sheets with the thickness of 1/4 in. Figure 2-16 illustrates most parts of the dock fabricated in DISL lab, while Figure 2-17 presents actual assembly of the dock.

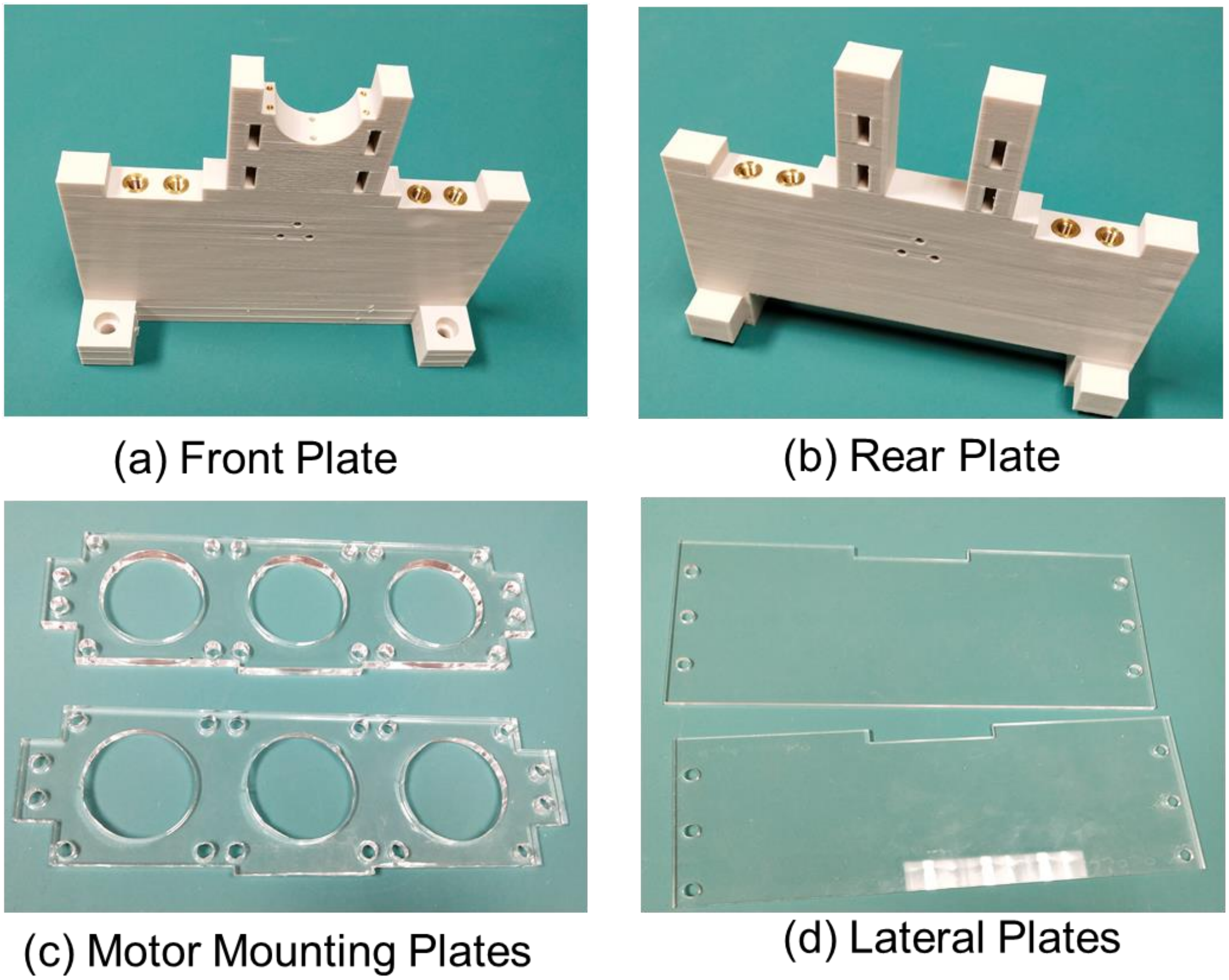

*Figure 2-16: Parts of The Dock Fabricated in DISL*

As is shown on Figure 2-15, to ensure sufficient compactness, the mounting location of actuators are well arranged so that the 2 actuators driving each segment of the robot arm can be accommodated opposite against each other on the dock. This improved the efficiency of packing the actuators significantly and can save a considerable amount of space. To satisfy easy assembly and maintenance requirements, all parts of the dock are joint by screws and bolts. Application of bolts and screws allows users to assembly the dock easily and quickly. Since the dock is not assembled using permanent joints, broken parts can be separated from the dock and replaced quickly and smoothly. This means that only broken parts of the dock are required to be replaced, hence it could save the expense of fabricating new docks to replace the old ones considerably, compared with docks featuring permanent joints. For requirements of easy installation of actuators, the dock design makes sure that the actuators can be installed on the motor mounting plates quickly by fastening the M5 bolts and nuts.

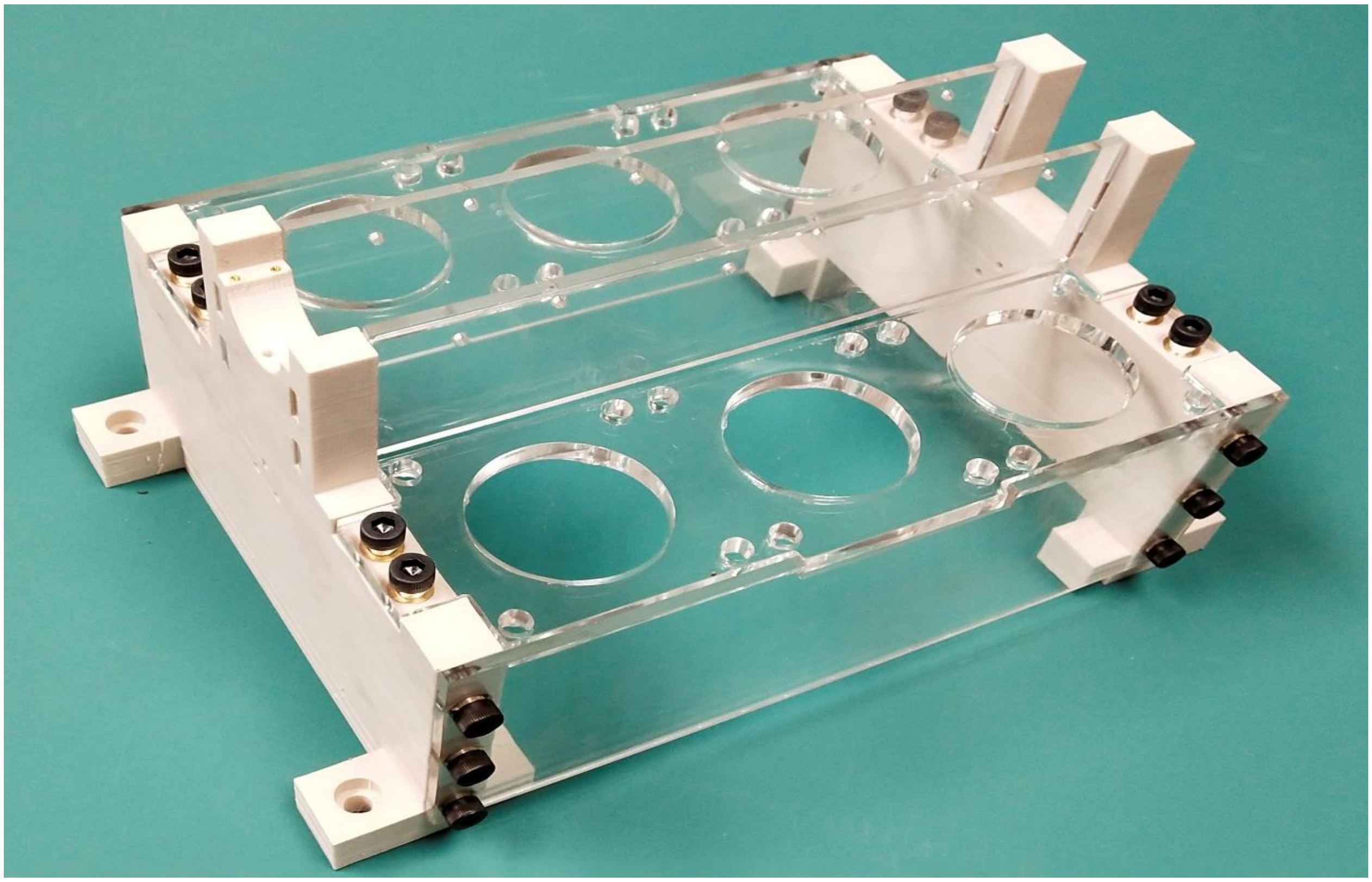

*Figure 2-17: Dock Fabricated in Lab*

As is shown on the top left side of Figure 2-15, the variable stiffness arm holder is located at the front of the dock. The holder is screwed to the front plate of the dock to mate with and securely fix the base of the variable stiffness robot arm. For the front plate and the back plate, each plate has 3 guide holes which are used to guide the vacuum tubing from multiple arm segments to the vacuum pump. After finishing the design of the actuator and the dock, the actuators can be mounted on the dock to form the actuation system of the variable stiffness continuum robot. The model of the actuation system is presented in Figure 2-18 below.

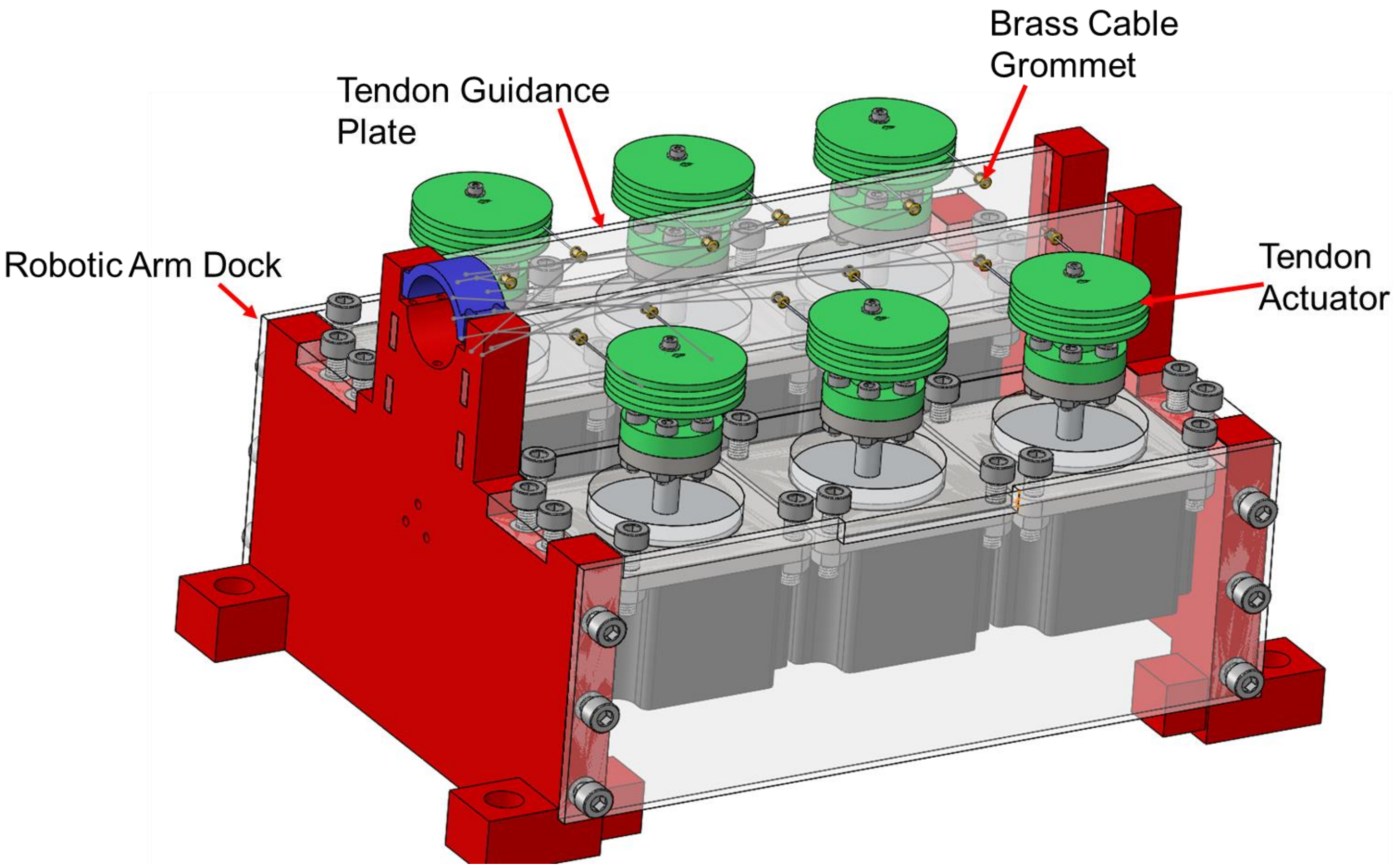

*Figure 2-18: CAD Model of The Actuation System for the Continuum Robot*

## 2.8 Tendon Guidance Mechanism

After figuring out the design of the actuator and multiple segmented variable stiffness robot arm, it is imperative to devise the tendon guidance mechanism. Since the spools cannot see the tendon pairs coming out of the robot arm directly and the tendon translation direction within the robot arm is not consistent with the motion direction along the rim of the spool, a tendon guidance mechanism is required to alter the tendon translation direction and guide the tendons to the spools. One solution of the guidance mechanism is to utilize a set of pulleys to guide the tendon pairs to each location of the spools. However, this would require 12 pulleys in total to route 12 tendons for a robot arm with 3 segments, and the corresponding pulley layout would be extremely complicated. And it would also be quite complicated to design corresponding features on the dock to mount the pulleys.

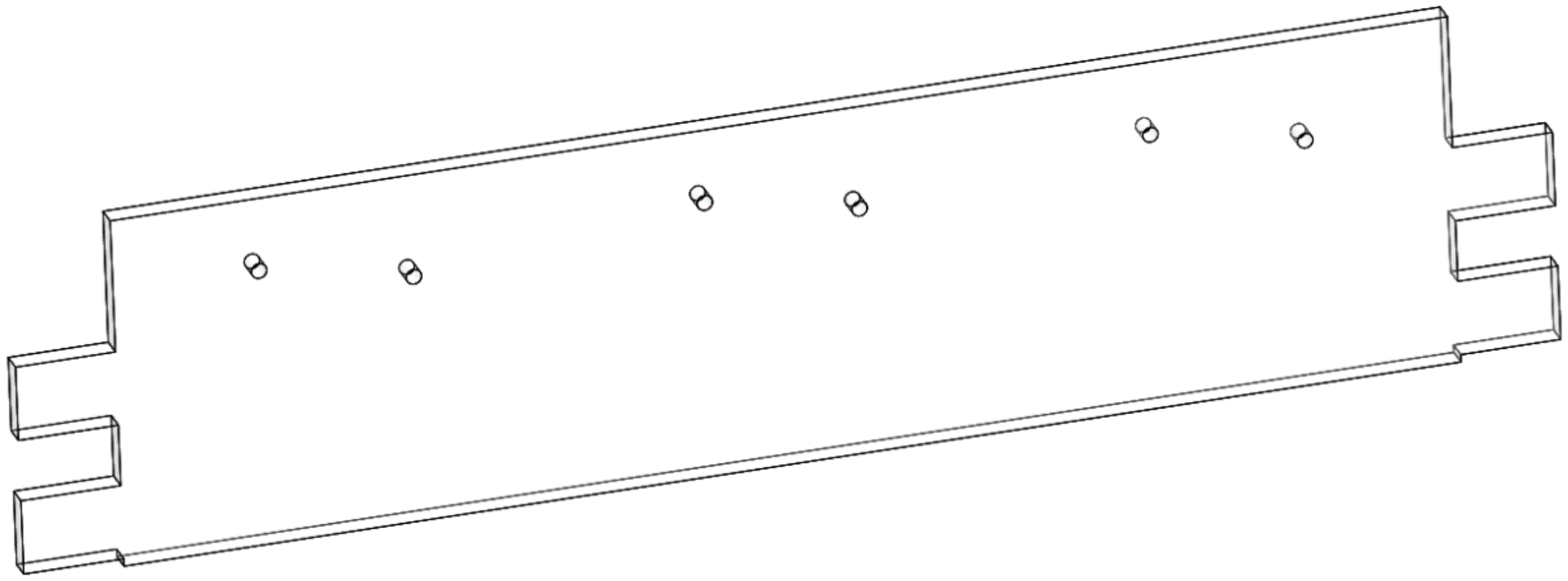

*Figure 2-19: Guidance Plate of the actuation system*

Compared with the first method, the second solution replaces pullies by small holes to divert the tendon pairs. In this case, one rigid plate with several small holes can divert multiple tendons simultaneously, while each pulley can only divert the motion of one tendon. The fabrication process for a guidance plate is simple as well since a pattern of feature of holes can be formed easily using the laser cutter. Therefore, the second solution was chosen and guidance plates were designed as a part of the tendon guidance mechanism. Figure 2-19 shows the design of the guidance plate which has 6 holes to divert 3 tendon pairs to 3 spools separately. To reduce the friction between the guidance plates and the tendons, the wall of the holes is covered by a brass cable grommet, thus improving the life span of the actuation tendons. The CAD model of the brass cable grommet is available in Figure 2-20 below, while Figure 2-21 shows a grommet installed on the guidance plate.

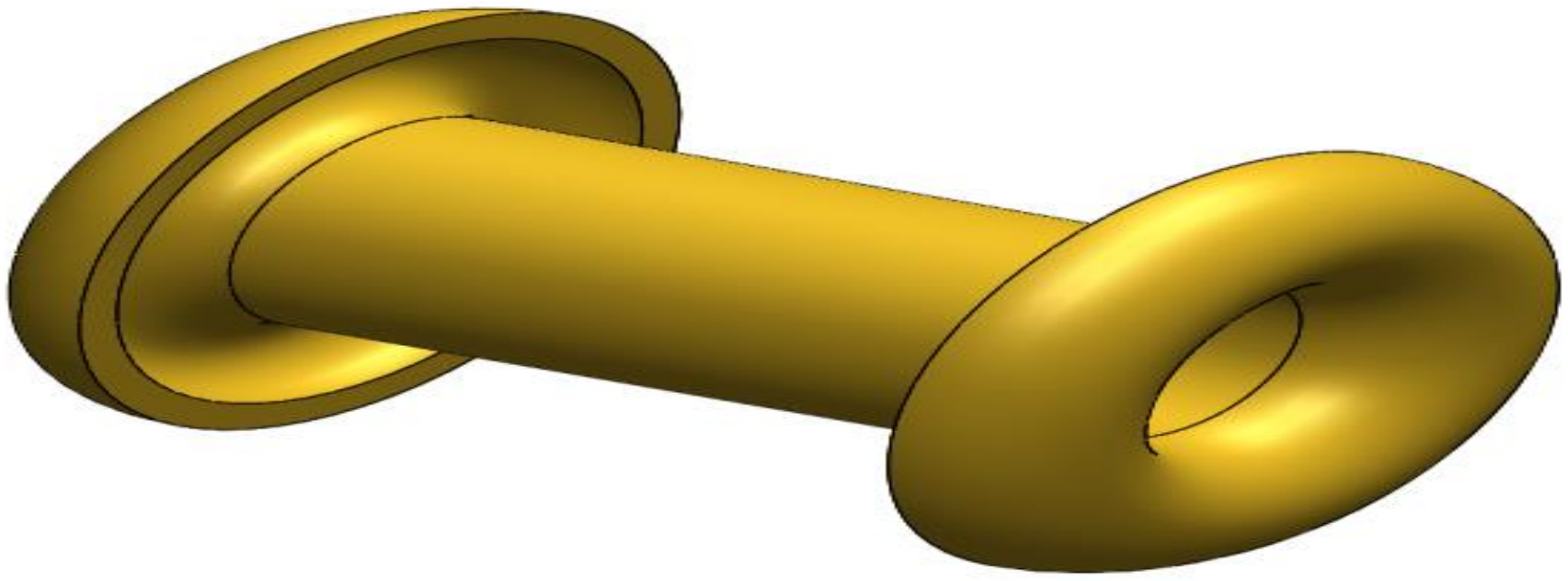

*Figure 2-20: CAD Model of the Brass Cable Grommet*

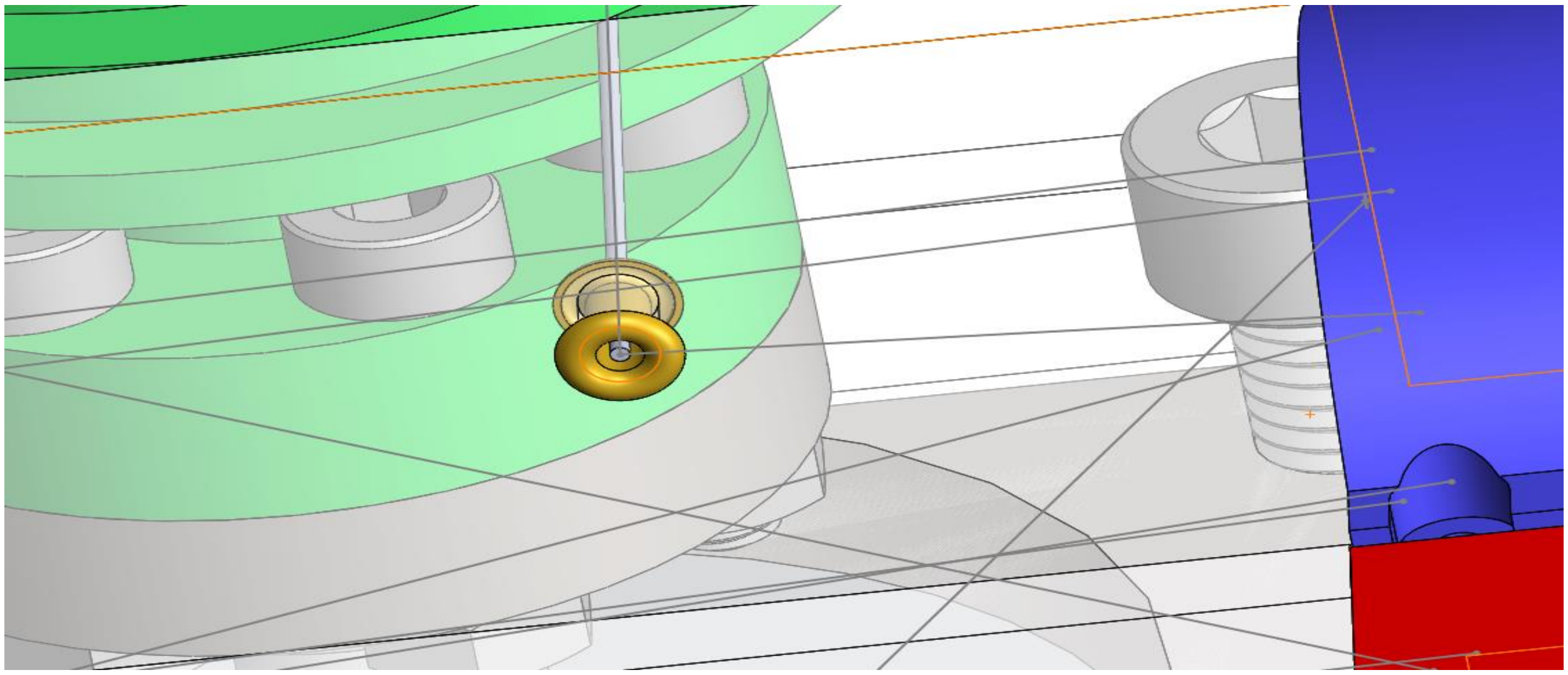

*Figure 2-21: A Brass Cable Grommet Installed on the Guidance Plate*

## 2.9 Electronics for Actuation System Control

The electric circuit utilized to control the actuation of the variable stiffness robotic arm with 2 segments consists of the 18 V stepper motor power supply, 4 100uF capacitors, 4 stepper motors, controller power supply, Arduino MEGA 2560 Microcontroller, and 4 rotary encoders. NEMA size 23 stepper motor is used to drive the tendons attached to the robotic. This type of motor has a maximum torque output of 392 N-mm, step angle of 1.8 degrees (200 steps/revolution), rated current of 1.0 A and weight of 0.45 kg. The stepper has sufficient torque to pull the tendons and drive the robotic arm segments. To output the maximum torque, the stepper motor should be connected to a 18 V power supply. The purpose of adding 4 capacitors to the circuit is to decouple the control circuits of the stepper motors from each other and reduce the noise of the signal transferred to the stepper motors. Since a microcontroller cannot output enough voltage and current to provide power for the stepper motor,  the stepper cannot be connected to the microcontroller directly. Hence a A4988 stepper motor driver is required to drive each motor. The picture of the A4988 stepper motor driver is available in the following Figure 3-22 [17].

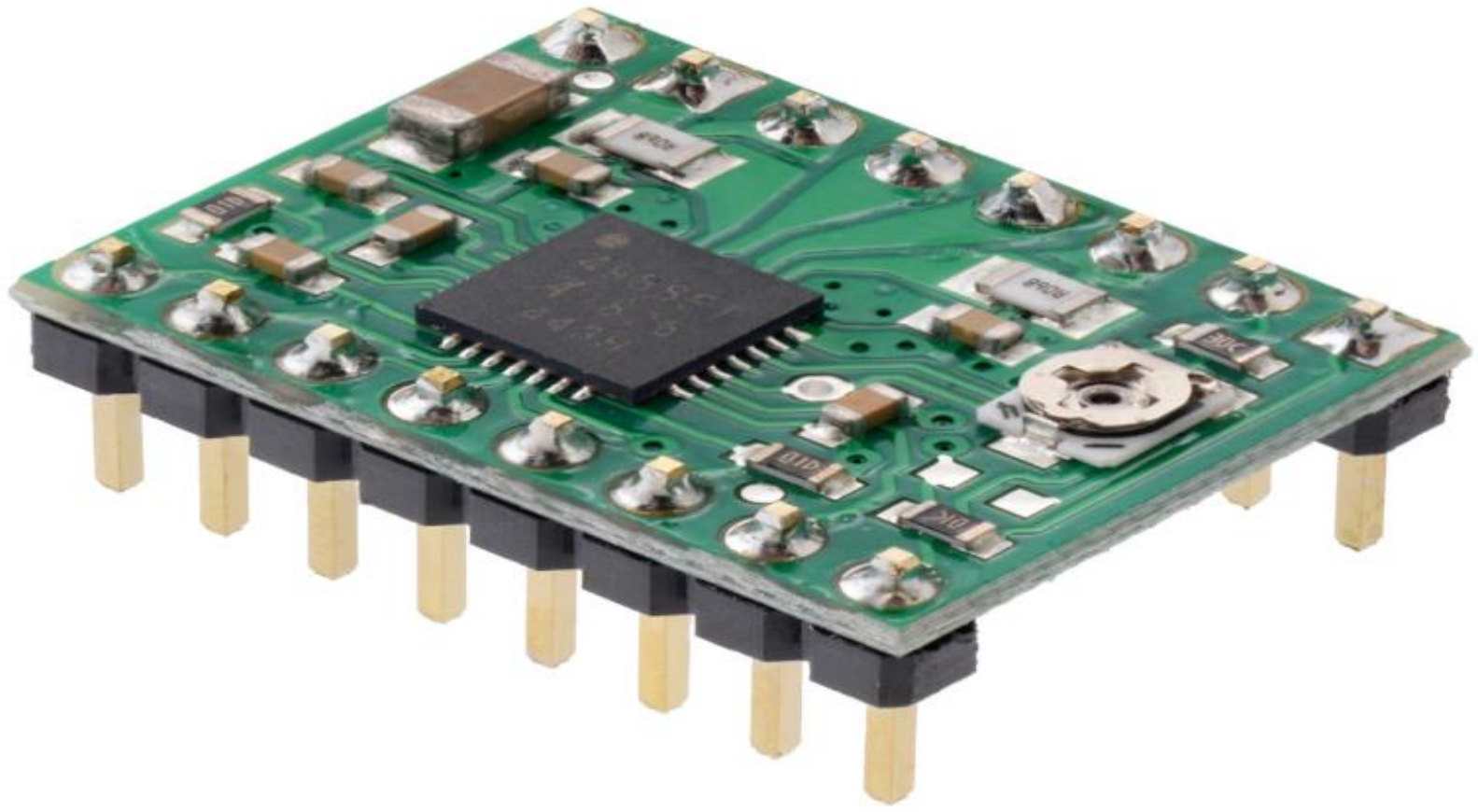

*Figure 2-22: A4988 Stepper Motor Driver [17]*

The A4988 motor driver is connected to the stepper motor through 4 pins. The 18 V motor power supply is not connected to the stepper motors directly, instead, it is hooked to power supply pins of the motor drivers. Through the STEP pin and the DIR pin on A4998, the microcontroller can communicate with the stepper motor driver and control the motor direction. To completely control the step size and motion of the motor, 5 pins are required to establish communication between each motor driver and the microcontroller. Since there are 4 stepper motors powered by 4 motor drivers, 20 pins are required to allow the microcontroller to dictate the motion of the variable stiffness continuum robot arm. Because Arduino Uno does have enough number of pins, the Arduino MEGA 2560 microcontroller is utilized instead to control the group of 4 actuators simultaneously. Figure 3-23 below shows the layout of the Arduino MEGA 2560 microcontroller used to controller the variable stiffness robot prototype.

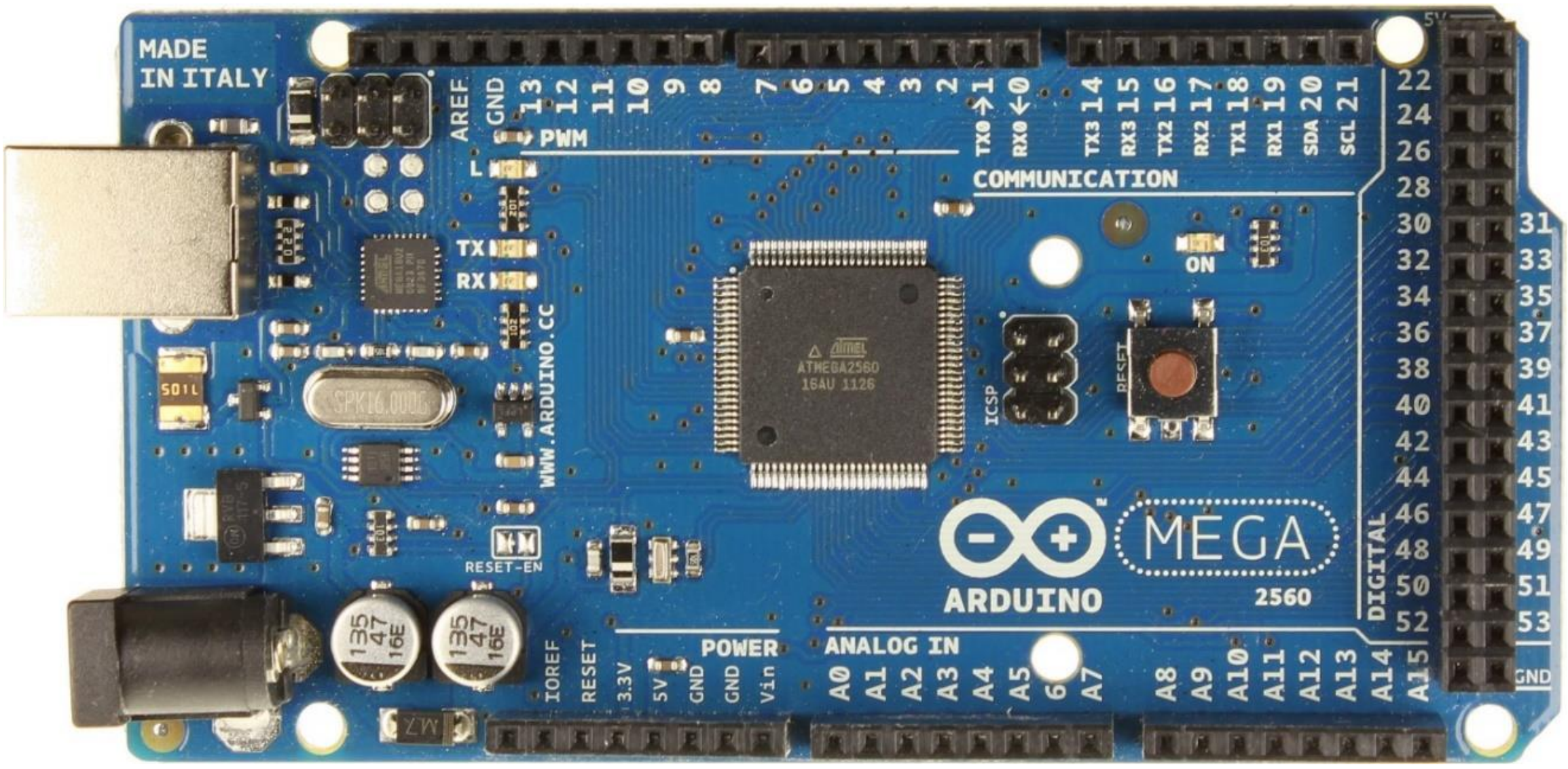

*Figure 2-23: Arduino MEGA 2560 Microcontroller [18]*

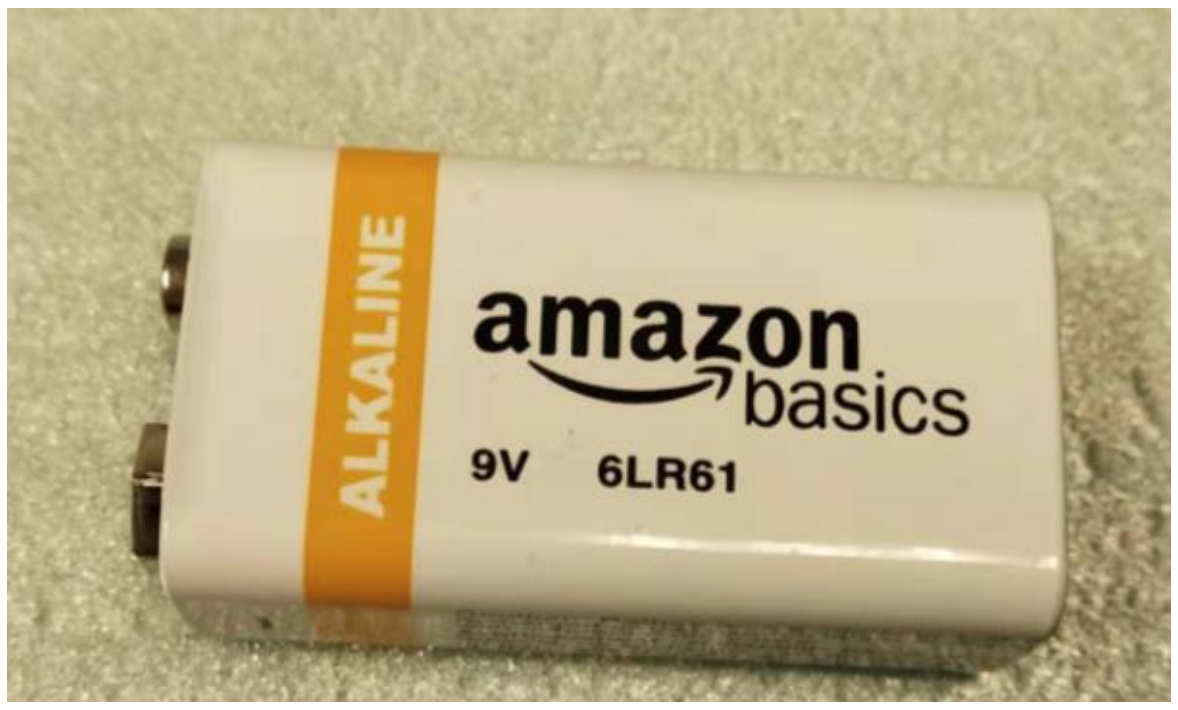

*Figure 2-24: 9V Battery Utilized to Power The Arduino 2560 MEGA Controller*

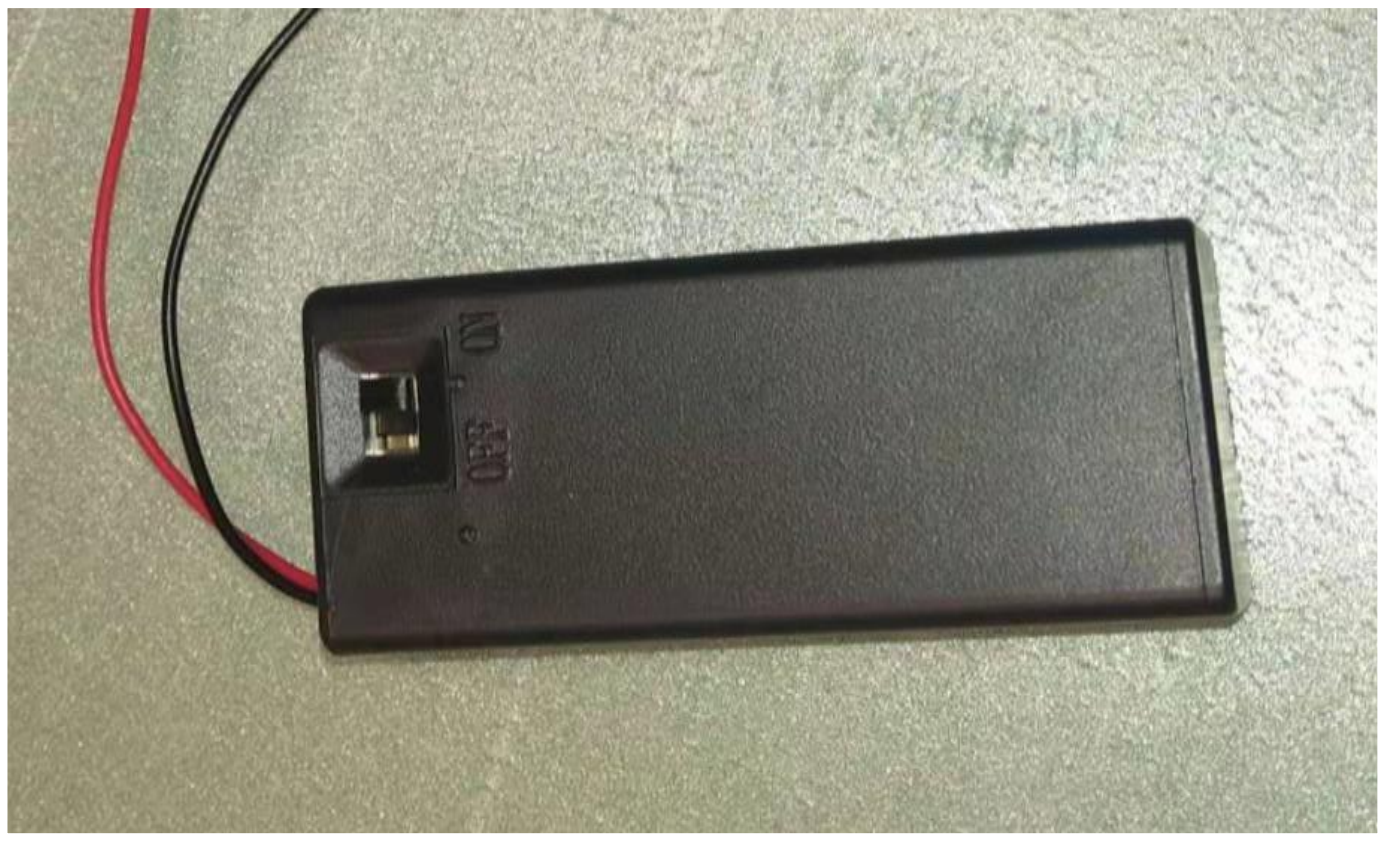

*Figure 2-25: Battery Holder with ON/OFF Switch*

Compared with Arduino Uno, Arduino MEGA 2560 has much more input/output pins including 54 digital input/output pins and 16 analog input pins. The controller is powered by a 9V battery as is shown in Figure 3-24 below.

To avoid the short circuit of the battery, the 9 V battery is placed inside the battery holder as is shown in Figure 3-25 above. Another advantage of the battery holder is that it possesses as ON/OFF switch, which precludes the requirement to purchase separate power switches.

Another significant advantage of choosing Arduino MEGA is that it possesses 6 external interrupt pins to which the rotary encoders are attached. In order to control the motion of the actuators with fast response, the microcontroller should be able to read data from the rotary encoders continuously to ensure that any desired motion input from the user can captured without losing input signal from the knobs. The interrupt pins on Arduino MEGA 2560 can capture pulses from the rotary encoder, and the number of pulses captured is related to the angular displacement of the stepper motor shaft. In other words, the knobs need to send clock signals to the microcontroller via the interrupt pins to synchronize the knob rotation with the displacement of tendons. Since one rotary encoder needs one interrupt pin on microcontroller to send the clock signal, a total of 4 actuators requires 4 interrupt pins in total, and Arduino MEGA 2560 can satisfy this requirement. Figure 3-26 shows the rotary encoder used to manipulate the robotic arms of the variable stiffness robot system.

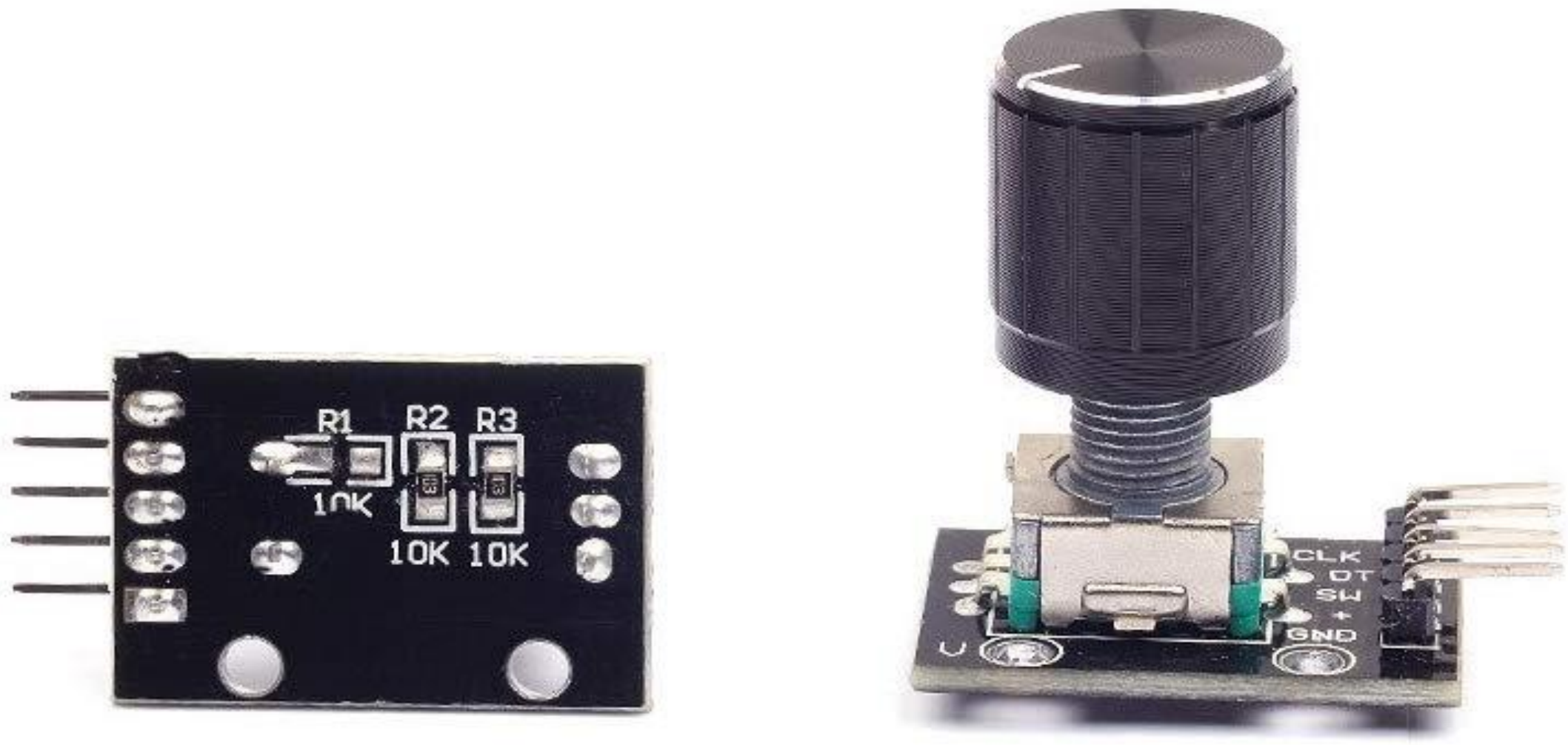

*Figure 2-26: Encoder for Manipulation of The Robotic Arm [18]*

The encoder (knob) utilized to control the motion of the multiple segments of the robotic arm has 5 pins including CLK, DT, SW, + and GND. The CLK pin is connected to the interrupt pin of the microcontroller to transfer pulses which corresponds to a desired bending angle and direction for each arm segment. The DT pin represents the digital output of the encoder. The SW corresponds to the output of the switch button on the knob. The + pin is connected to the logical power supply of 9 V. The GND pin should be connected to the ground. Among all of the 5 pins on the knob, pins CLK, DT and SW are connected to the microcontroller, therefore a total of 12 pins on the microcontroller need to be used to fully manipulate the 4 actuators with 4 knobs.

As is mentioned in the previous paragraph, the variable stiffness continuum robot has 2 segments is controlled by 4 knobs. The 4 knobs are divided into 2 pairs of knobs. The first pair of knobs 1 and 2 has the function of manipulating segment 1 of the robot arm, while the second pair of knobs 3 and 4 is in charge of manipulating the motion of segment 2. To be more specific, knob 1 receives the desired bending angle on ZX plane from the user, then converts and transfers the signal to the microcontroller. Similarly, knob 2 receives the desired bending angle on the ZY plane from the user, then converts and transfers the signal to the microcontroller.

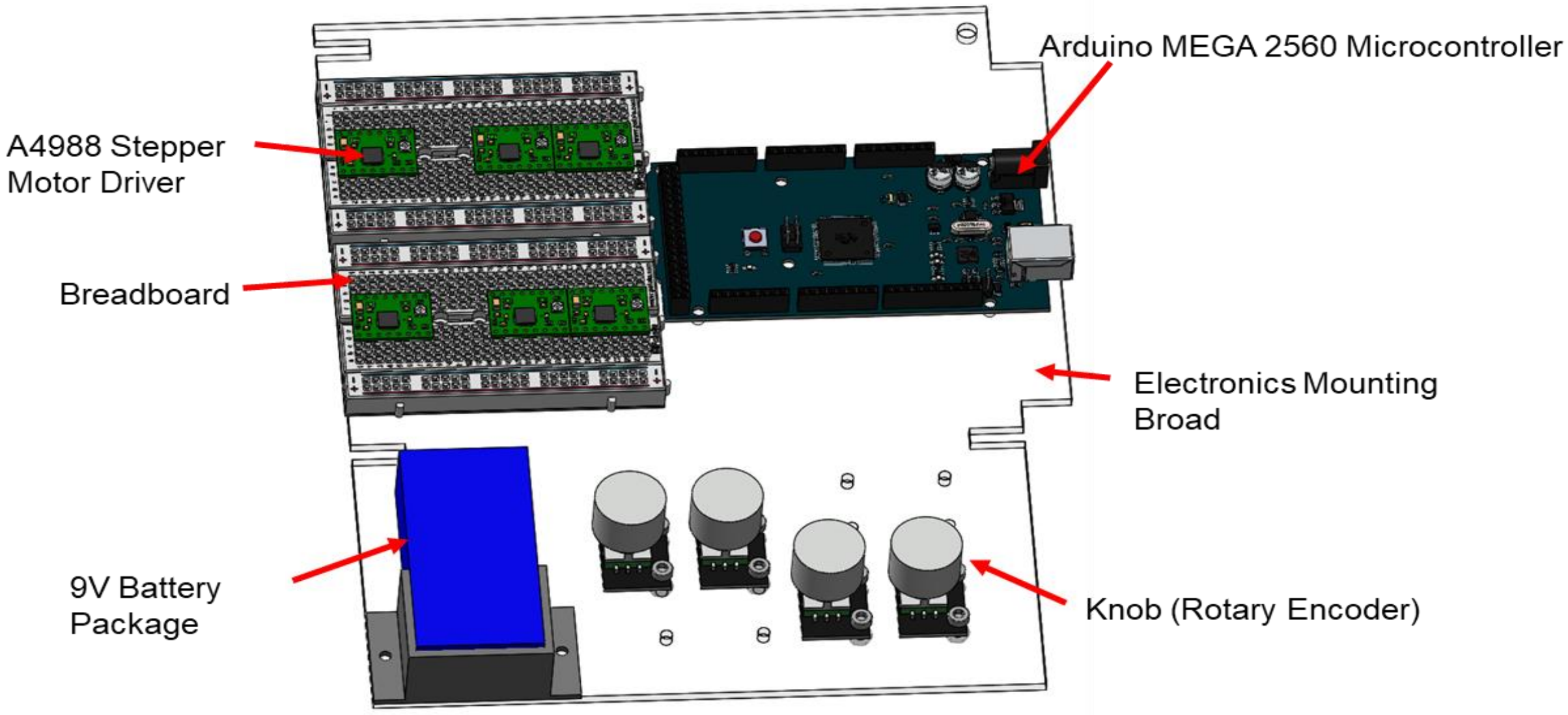

*Figure 2-27: CAD Model of the Actuation Control System*

After figuring out specifications of the electronic components of the motor control system, the layout of the circuit was designed as is presented in Figure 3-28 on the next page.  and the CAD model of the motor control system was created and verified. The CAD model of the control system of the actuators is available in Figure 3-27 in the previous page.

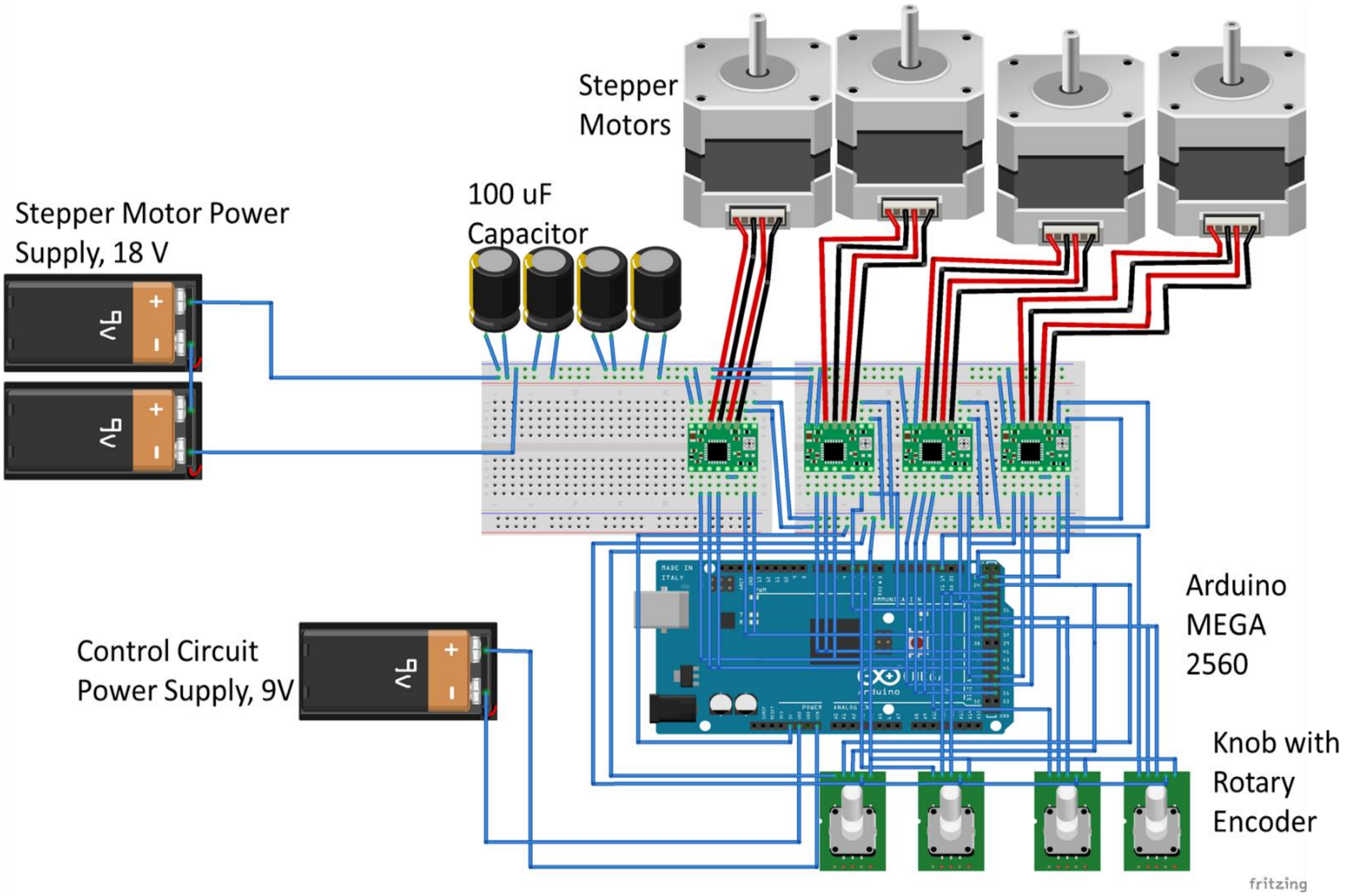

*Figure 2-28: Layout of The Stepper Motor Control Circuit*

Then all electronic components of the variable stiffness robot actuation system were purchased, fabricated and assembled. The prototype of the motor control board is presented in Figure 3-29 below.

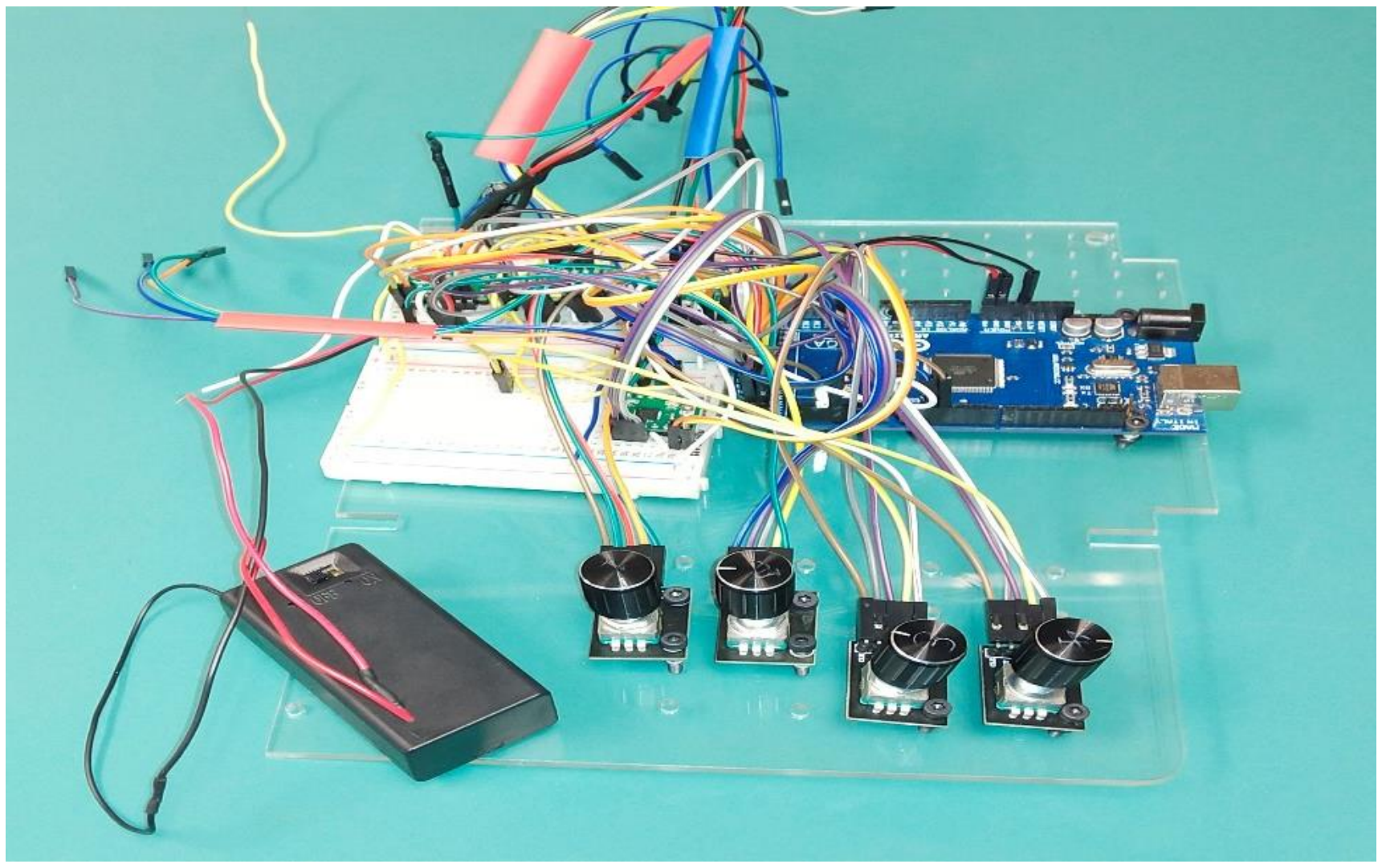

*Figure 2-29: Prototype of The Motor Control Board*

## 2.10 Overall Design of The Variable Stiffness Continuum Robot

After the completion of the design for the robotic arm segments, connector, dock, spools and actuation control system, the CAD model of assembly for the variable stiffness continuum robot with 3 segments was created. The final design of robot prototype is shown in Figure 3-30 in the following page.

As is shown in Figure 3-30 , the continuum robot designed has 3 arm segments actuated by 6 stepper motors. The tendon pairs are attached to the spools which are secured on the motor shafts. The board with electronic components is placed at the bottom of the dock. The actuation system can be mounted to the table via 4 lugs at the bottom of the dock. For the tendon guidance mechanism, the 6 antagonistic tendon pairs are routed to the spools through tiny holes of 2 parallel guidance plates. Because of lack of time to fabricate the third segment of the robotic arm in this semester, the prototype fabricated has 2 segments as is presented in Figure 3-31 in the following page.

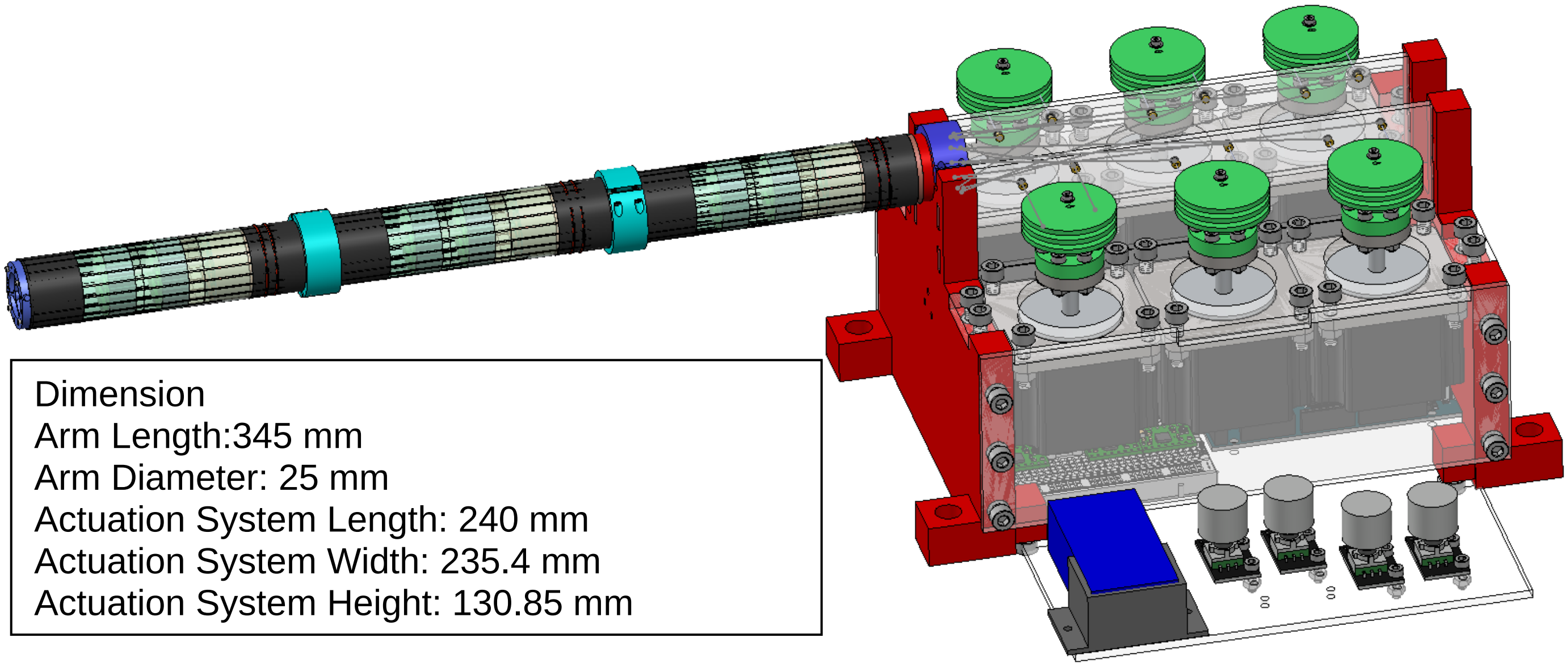

*Figure 2-30: Final Design of The Variable Stiffness Continuum Robot*

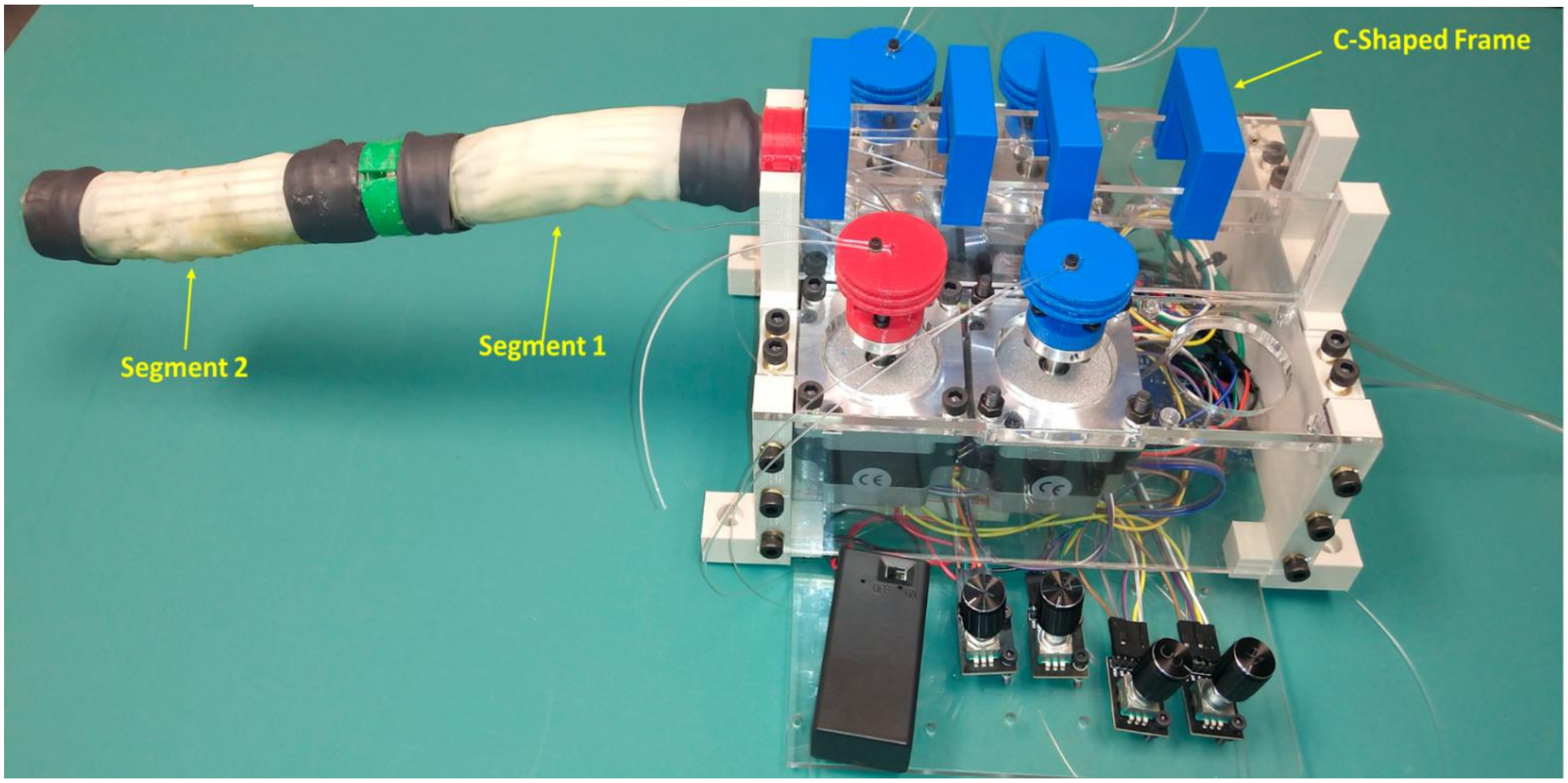

*Figure 2-31: 2-Segmented Variable Stiffness Continuum Robot Prototype*

Figure 2-31 presents the variable stiffness continuum robot prototype fabricated in DISL. The robot arm has 2 segments in total including segment 1 with one end mounted on the dock and segment 2 at the distal end. However, initial actuation test shows that the guidance plates experience a great deformation due to forces applied by the tendons on the wall surface of the tiny holes where the tendons travel through, this poses a serious issue to the actuation precision of the arm segments through tendon translation, since the travel paths of the tendons are indeterminable due to structural deformation of the guidance mechanism . To resolve this problem, a C-shaped frame was designed to improve the rigidity of the guide plates and the dock. As is shown in Figure 3-31, 4 C-shaped frames were mounted to the top of the 2 guide plates. This can improve the stiffness of the guide plate and maintain the actuation accuracy of antagonistic tendon pairs.

# Chapter 3 Kinematics and Actuation Methodology

To achieve precise actuation of the multiple segmented robot arm to ensure that the shape and motion of the continuum structure follow the desired kinematic parameters specified by the user, it is an indispensable work to analyze the kinematics of the tendon pairs attached to ends of different segments of the variable stiffness robotic arm. First, the kinematic model for a single arm segment will be established for the cases of planar bending and spatial bending respectively. Then the kinematic model of a 2-Segmented arm will be proposed and analyze special bending for a robotic arm with 2 segments connected in series moving simultaneously.

## 3.1 Planar Bending of One Single Arm Segment

Planar bending of one arm segment is probably simplest case of bending for a compliant beam structure. To help explain the geometric relations for a bended compliant structure, Figure 4-1 showing the geometry of a bended beam is presented below.

To simply the analysis of the planar bending, one basic assumption should be accepted that the curvature of the beam has the same value along the neutral axis of the robotic arm segment. Under this assumption, the bended arm segment can be approximated as a segment of a ring. As is shown in Figure 3-1, there are 2 tendons arranged opposite to each other with a distance of d. When the arm segment is bent with an angle of $\theta_x$, the length of the inner tendon is decreased by $\Delta x$. Assuming that the beam has the original length of $l$, the following geometric relations can be derived.

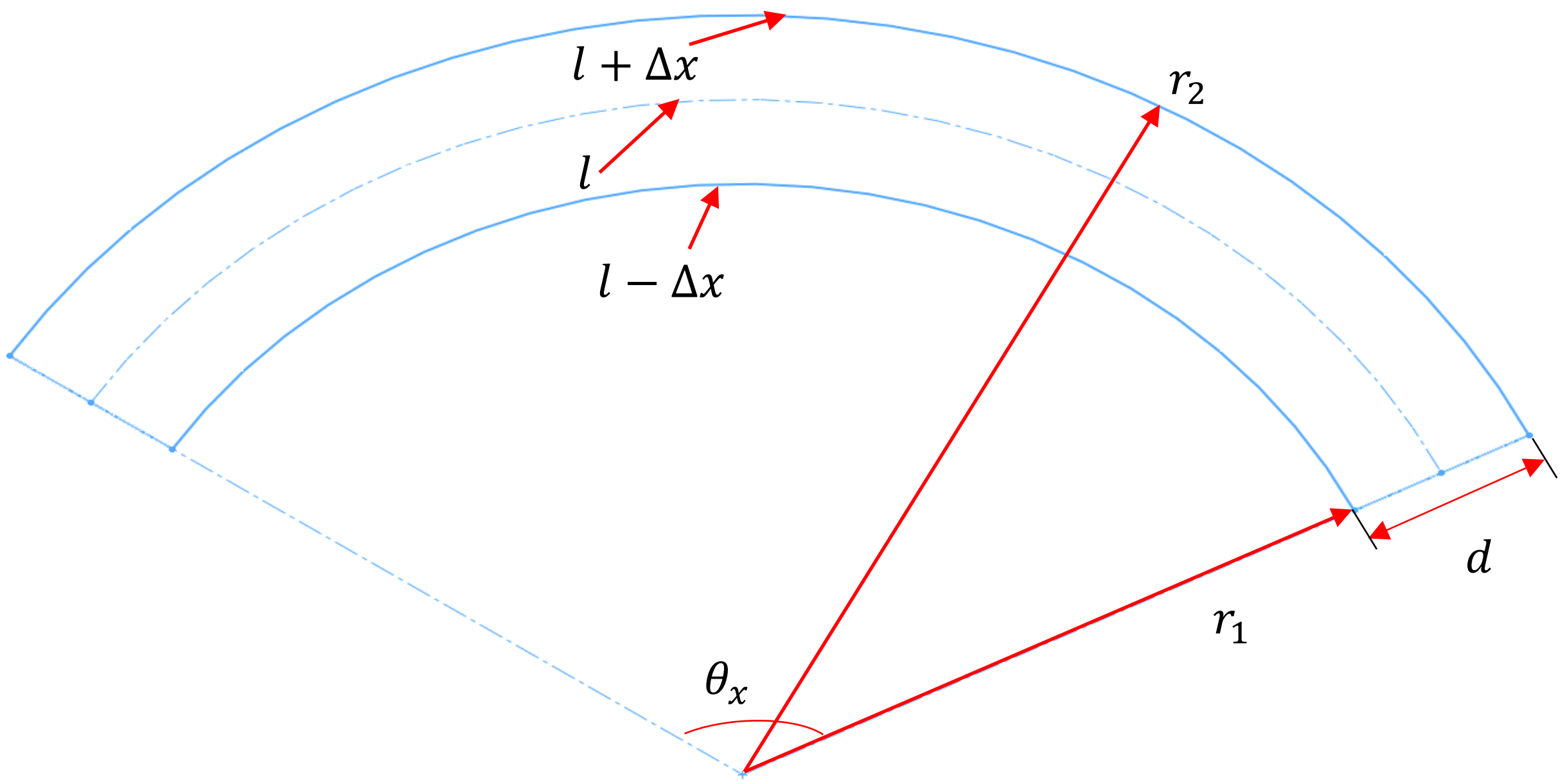

*Figure 3-1: Geometry of A Beam Under Planar Bending*

To simply the analysis of the planar bending, one basic assumption should be accepted that the curvature of the beam has the same value along the neutral axis of the robotic arm segment. Under this assumption, the bended arm segment can be approximated as a segment of a ring. As is shown in Figure 3-1, there are 2 tendons arranged opposite to each other with a distance of d. When the arm segment is bended with an angle of $\theta_x$, the length of the inner tendon is decreased by $\Delta x$. Assuming that the beam has the original length of $l$, the following geometric relations can be derived.

The length of tendon staying in the inner side of the arm segment

$$l_i = l - \Delta x$$

Since the distance between the neutral axis and the inner tendon is the same as the distance between the outer tendon and the natural axis, the length of tendon remaining in the outer side of the arm segment should be

$$l_o = l + \Delta x$$

According to the basic assumption, the beam has a constant curvature along the longitudinal direction, the following equations can be obtained

$$l - \Delta x = r_1 \theta_x \quad (1)$$

$$l + \Delta x = r_2 \theta_x \quad (2)$$

Subtract equation (2) by (1), the bending angle equation is obtained and available as follows

$$\theta_x = \frac{2\Delta x}{d} \qquad (3)$$

where $\theta_x$ = The planar bending angle of the arm segment, rad

$\Delta \boldsymbol{x}$= Displacement of the tendon pair

$d$ = Distance between 2 antagonistic tendons of a tendon pair

Equation (3) shows that the planar bending angle is proportional to the displacement of the tendon pair. The planar bending angle of a single arm segment can be calculated using equation (3) presented above once the tendon pair displacement and the distance between 2 antagonistic tendons of a tendon pair is known.

## 3.2 Spatial Bending of One Single Arm Segment

For the case of special bending of one single arm segment, the situation is more complicated. However, conclusions from the previous analyses of planar bending can still be applied to derive the kinematic relations for this spatial bending case using the geometry of the 2 antagonistic tendon pair as is presented in Figure 3-2 below.

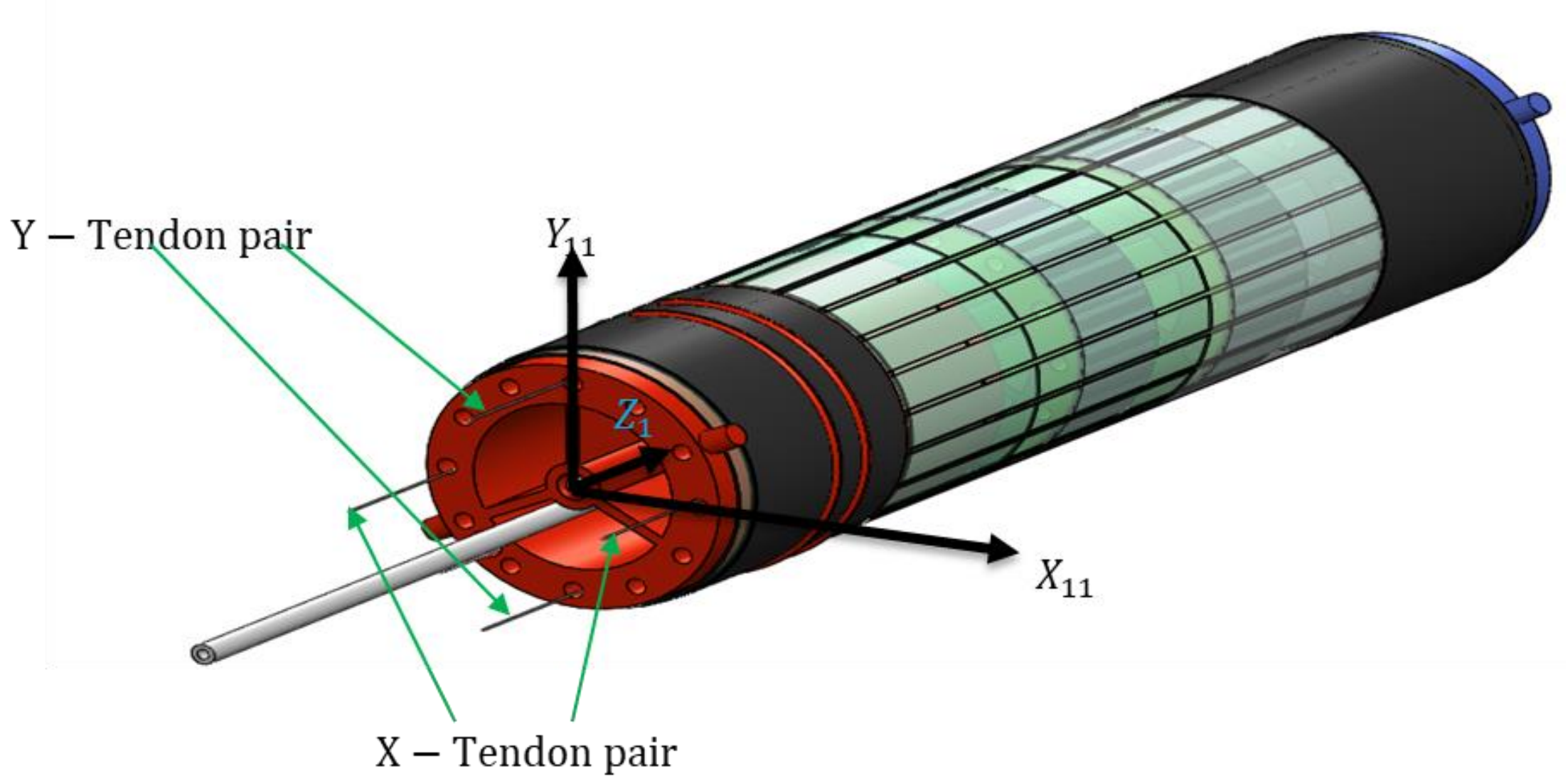

*Figure 3-2: Orientation of the 2 Antagonistic Tendon Pairs of a Single Arm Segment*

As is shown in Figure 3-2, $X_{11}Y_{11}Z_1$ is the local coordinate system of a single arm segment, and X-Tendon pair is on plane $Z_1X_{11}$ while Y-Tendon pair is located on plane $Z_1Y_{11}$ .In this case, plane $Z_1X_{11}$ plane $Z_1Y_{11}$ are orthogonal. Hence this can give the basic assumption for the corresponding kinematic analysis that the orientation of X-Tendon pair is 90 degrees away from the orientation of Y-Tendon pair. In this case, the 2 tendon pairs attached to the end of the arm segment are orthogonal, therefore X-Tendon pair and Y-Tendon pair can be actuated independently. This orthogonality means that the superposition rule is valid. In other words, the special motion of the single arm segment is the superposition of planar bending in 2 orthogonal planes of $Z_1X_{11}$ and $Z_1\mathrm{Y}_{11}$. From the planar bending equations derived previously, the special bending can be described by 2 bending angles of within orthogonal planes as follows

The bending angle within plane ZX: $$\theta_{ZX} = \frac{2\Delta x}{d} \tag{4}$$

The bending angle within plane ZY: $$\theta_{ZY} = \frac{2\Delta y}{d} \tag{5}$$

where $\Delta x$ is the displacement of the antagonistic X-Tendon pair

$\Delta y$ is the displacement of the antagonistic Y-Tendon pair

## 3.3 Spatial Bending of a 2-Segmented Robotic Arm

For the case of special bending of a 2-segmented robotic arm, the complicity is much higher than the previous 2 cases of bending of a single segment robotic arm. Figure 3-3 below shows the simplified geometrical model of a 2-segmented arm prototype developed for the variable stiffness continuum robot. As is shown Figure 3-3, the continuum robotic arm consists of 2 segments including segment 1 and segment 2. The 2 segments are connected in series by a connector which is not shown in the figure. For segment 1, it is driven by tendon pairs of $\mathrm{X}_1$ and $\mathrm{Y}_1$ which are orthogonal to each other, with the tendon ends attached to the distal side of segment 1. For segment 2, it is driven by tendon pairs of $\mathrm{X}_2$ and $\mathrm{Y}_2$ which are orthogonal to each other, with the tendon ends attached to the distal side of segment 2. Another noticeable thing is that the tendon attached to segment 2 end are also routed through segment 1, as is shown in Figure 3-3. For the continuum robot developed in this project, the orientation of tendon pairs attached to segment 2 is different from that of segment 1, since the orientation of segment 2 tendon pairs is 30 degrees away from tendon pairs attached to segment 1 distal end. In this case, it is inevitable to face kinematic coupling issues. In other words, the tendon pairs attached to each segment cannot be actuated independently. The reason behind this is that the tendon pairs attached to segment 2 also pass through segment 1, which means that the motion of tendons attached to segment 2 is partially constrained by the shape (geometry) and motion of segment 1.

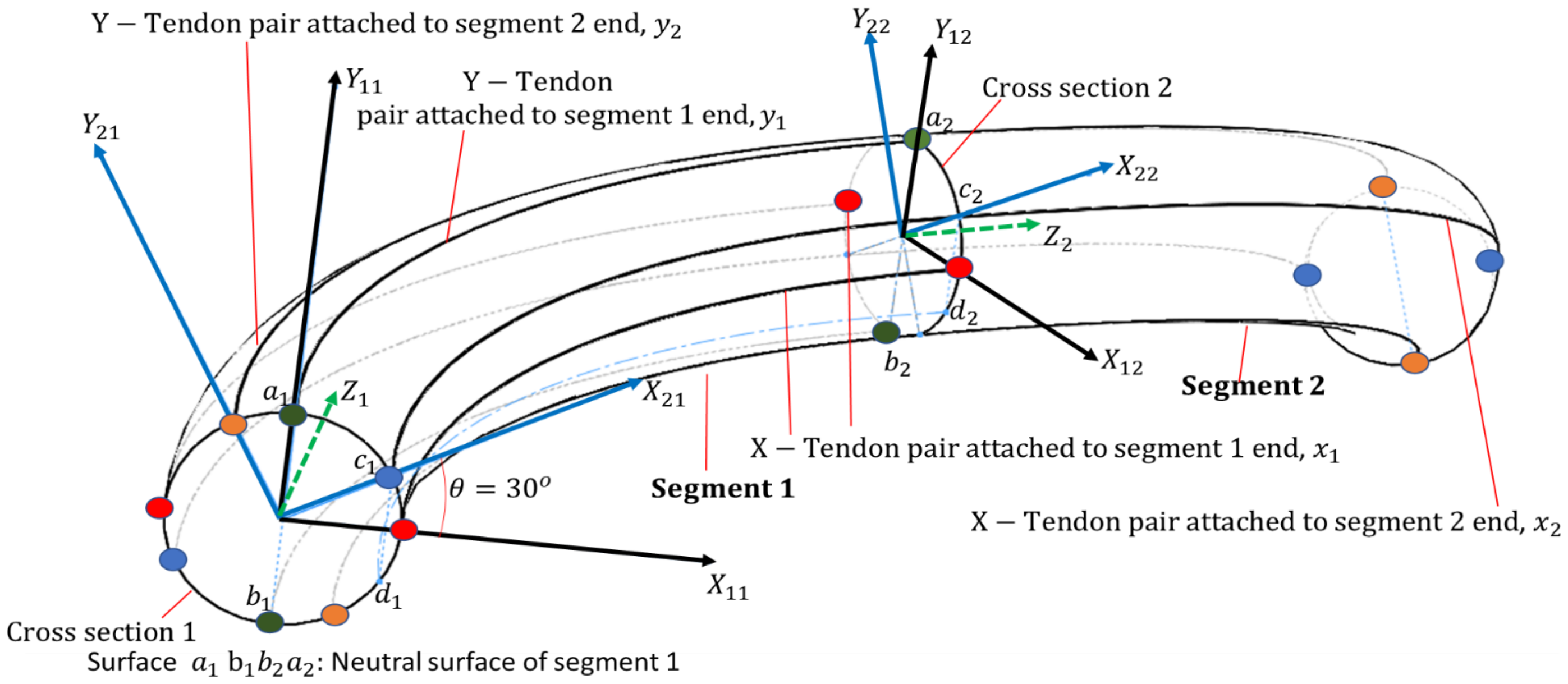

*Figure 3-3: Geometry of A 2-Segmented Robotic Arm*

To decouple the actuation of the 4 tendon pairs for the 2 segmented robotic arm, it is imperative to analyze the geometrical relationship between the 4 tendon pairs. For tendon pairs of $X_1$ and $Y_1$ attached to segment 1 end, they can be actuated independently since the tendon pairs pass through one arm segment only. However, translational displacement of tendon pairs $X_2$ and $Y_2$ attached to the end of segment 2 is coupled with segment 1 and cannot be actuated independently. Therefore the actuation of tendon pairs for segment 1 follows the following equations

X-Tendon Displacement, attached to segment 1 End: $x_{1o} = x_{1i}$

Y-Tendon Displacement, attached to segment 1 End: $y_{1o} = y_{1i}$

where $x_{1i}$ , $y_{1i}$ are desired displacements of tendons pairs attached to segment 1, corresponding to motion inputs from the user through knob 1 and 2 ; $x_{1o}$ , $y_{1o}$ are actuated tendon displacement of tendon pairs $\mathrm{X}_1$ and $\mathrm{Y}_1$ attached to the segment 1 end.

For tendons attached to segment 2, the bending of segment 2 contribute partially to the displacement of tendon pairs of $\mathrm{X}_2$ and $\mathrm{Y}_2$. However, the bending of segment 1 is also capable of causing the tendon pairs attached to segment 2 to have another component of displacement as the tendon pairs also pass through segment 1. For example, when the segment 1 bends toward the X axis, tendon pair $\mathrm{X}_1$ and $\mathrm{X}_2$ are required to contract simultaneously, otherwise the continuum structure would be broken or one of the tendon pairs would collapse. Therefore, the overall motion of tendon pairs attached to segment 2 is a combination of the motion component of the tendon pairs within segment 1 and another component of tendon displacement within segment 2. This means that the displacement of tendon pairs attached to segment 1 ($x_{1o}$and $y_{1o}$) need to be factored into the calculation of the displacement of tendon pairs attached to segment 2 ( $x_{2o}$ and $y_{2o}$ ). It is acceptable to take the calculation of X-tendon pair displacement of segment 2 ($x_{2o}$) as an example. Assume that $x_{2i}$ is the desired tendon displacement of tendon pair $\mathrm{X}_2$ inside segment 2, and the displacement of tendon pairs attached to segment 1 end are $x_{1o}$ and $y_{1o}$, respectively. Then the overall displacement of the $\mathrm{X}_2$ should be attributed to the desired $\mathrm{X}_2$ tendon pair displacement within segment 2 ($x_{2i}$) and displacement of tendon pairs attached to segment 1 end ( $x_{1o}$ and $y_{1o}$ ). Since the orientation of tendon pair $\mathrm{X}_2$ is 30 degrees away from $\mathrm{X}_1$ and 60 degrees away from $\mathrm{Y}_1$

As is shown on Figure 3-3, the overall displacement of tendon pair $\mathrm{X}_2$ can be expressed as

$$x_{2o} = x_{2i} + x_{1i}\cos(30^o) + y_{1i}\cos(60^o)$$

Similarly, the overall displacement of tendon pair $\mathrm{X}_2$ can be expressed as

$$y_{2o} = y_{2i} + y_{1i}\cos(30^o) + x_{1i}\cos(120^o)$$

where $x_{1i}$ , $y_{1i}$ are desired displacements of tendons attached to segment 1,corresponding to motion inputs from the user through knob 1 and 2; and $x_{2i}$ , $y_{2i}$ are desired displacements of tendons within the space of segment 2, corresponding to motion inputs from the user through the knob 3 and 4; $x_{2o}$ , $y_{2o}$ are actuated tendon displacement of tendon pairs $\mathrm{X}_2$and $\mathrm{Y}_2$ attached to the segment 2 end.

However, there exists another issue with robotic arm actuation, which is that the actuation of segment 2 can cause segment 1 to bend with a small angle, even when segment 1 is not actuated. The reason is that the main compliant structure of segment 1 does not have sufficient stiffness to withstand the force exerted by segment 2 when segment 2 is actuated by tendon pairs of $\mathrm{X}_2$and $\mathrm{Y}_2$. Therefore, extra actuation on tendon pairs $\mathrm{X}_1$ and $\mathrm{Y}_1$ which are attached to segment 1 is required to offset the force exerted by segment 2. The extra actuation can actually be achieved by the introduction of the actuation compensation factor $\alpha$ for the actuation of tendon pairs attached to segment 1 end. Then the corrected system of tendon pair displacement equations can be obtained as follows, by analyzing the geometrical relations of special bending as is shown in Figure 3-3.

Y-Tendon Displacement, attached to Segment 1 End:

$$y_{1o} = y_{1i} - \frac{1}{2}\alpha y_{2i}\cos(30^o) - \frac{1}{2}\alpha x_{2i}\cos(60^o) \qquad (6)$$

X-Tendon Displacement, attached to Segment 1 End:

$$x_{1o} = x_{1i} - \frac{1}{2}\alpha x_{2i}\cos(30^o) - \frac{1}{2}\alpha y_{2i}\cos(120^o) \qquad (7)$$

Y-Tendon Displacement, attached to Segment 2 End:

$$y_{2o} = y_{2i} + y_{1i}\cos(30^o) + x_{1i}\cos(120^o) \qquad (8)$$

X-Tendon Displacement, attached to Segment 2 End:

$$x_{2o} = x_{2i} + x_{1i}\cos(30^o) + y_{1i}\cos(60^o) \qquad (9)$$

where the actuation compensation factor $\alpha$ can be obtained by trial and error during multiple iterations of actuation tests of the fabricated variable stiffness robot prototype. For the continuum robot prototype fabricated in DISL, experiments showed that the actuation compensation factor should be $\alpha = 1.00765$ to offset the influence of segment 2 actuation on segment 1.

The coupling effect can be eliminated once the actuation of the tendon pairs follow the tendon displacement equations proposed above. Using these equations, the algorithm and corresponding code were created to control the 4 actuators simultaneously and allow users to manipulate the motion of the 2 segmented variable stiffness robotic arm. The Arduino codes are available in Appendix 1 at the end of this thesis.

## 3.4 Sensitivity of the Actuation System

In this section, the sensitivity of actuation for each segment of the continuum robot arm is derived to obtain the transfer function converting the user input to the robotic system output. In this case, the user input is the angular displacement of the knob, while the system output is the bending angle of each arm segment.

The Sensitivity of the actuation system can be defined as the system output divided by the input

$$K = \frac{\alpha}{\theta_{knob}}$$

where $\alpha$ is the bending angle of the arm segment, while $\theta_{knob}$ is the angular displacement of the knob which is the input from the user.

Since the knob can generate 20 pulses in one revolution, the knob angle increment can be expressed as

$$\theta_{i,knob} = 360/20 = 18\ deg \qquad (10)$$

In the Arduino code controlling the stepper motor displacement, each pulse generated by the knob corresponds to motor angular displacement of

$$\theta_{i,motor} = 6.1875\ \text{deg} \qquad (11)$$

Divide equation (11) by (10), we have

$$\frac{\theta_{i,motor}}{\theta_{i,knob}} = \frac{6.1875}{18} \qquad (12)$$

Since the spool has a radius of R=10 mm, the increment of tendon displacement should be

$$x_i = R\theta_{i,motor} \qquad (13)$$

From the actuation test, the bending angle increment of each arm segment can be expressed by the following equation

$$\alpha_i = 3361.1[deg/m]\ x_i \tag{14}$$

Combine equations (12)(13)(14) to obtain the system transfer function

$$\alpha\ =\ 0.20165\ \theta_{knob} \tag{15}$$

Therefore, the actuation sensitivity of the continuum robot is $0.20165\ deg/deg$ .

# Chapter 4 Testing, Demonstration of the Prototype

This chapter covers the methodology and results of a series of tests to verify the functionality of the variable stiffness continuum robot and evaluate the performance of the prototype. At the beginning, a qualitative motion control test was conducted to roughly examine the actuation algorithm. Then the continuum robotic arm was separated from the dock to measure the load capacity with respect to the applied vacuum pressure. Finally, a planar bending actuation test was implemented to verify the proportional relationship between the tendon pair displacement and bending angle.

## 4.1 Qualitative Actuation Test

### 4.1.1 Testing of Planar Motion of Segment 1

After the fabrication of the dock and the robotic arm, one actuator was mounted to the dock and attached to the Y-Tendon pairs of segment 1 of the 2-segmented robotic arm as is shown in Figure 4-1 below. Segment 2 is not actuated in this case. As is shown in Figure 4-1, segment 1 of the prototype can bend upward or downward by rotating knob 1, which means the algorithm utilized to control the planar bending of a single arm segment is valid.

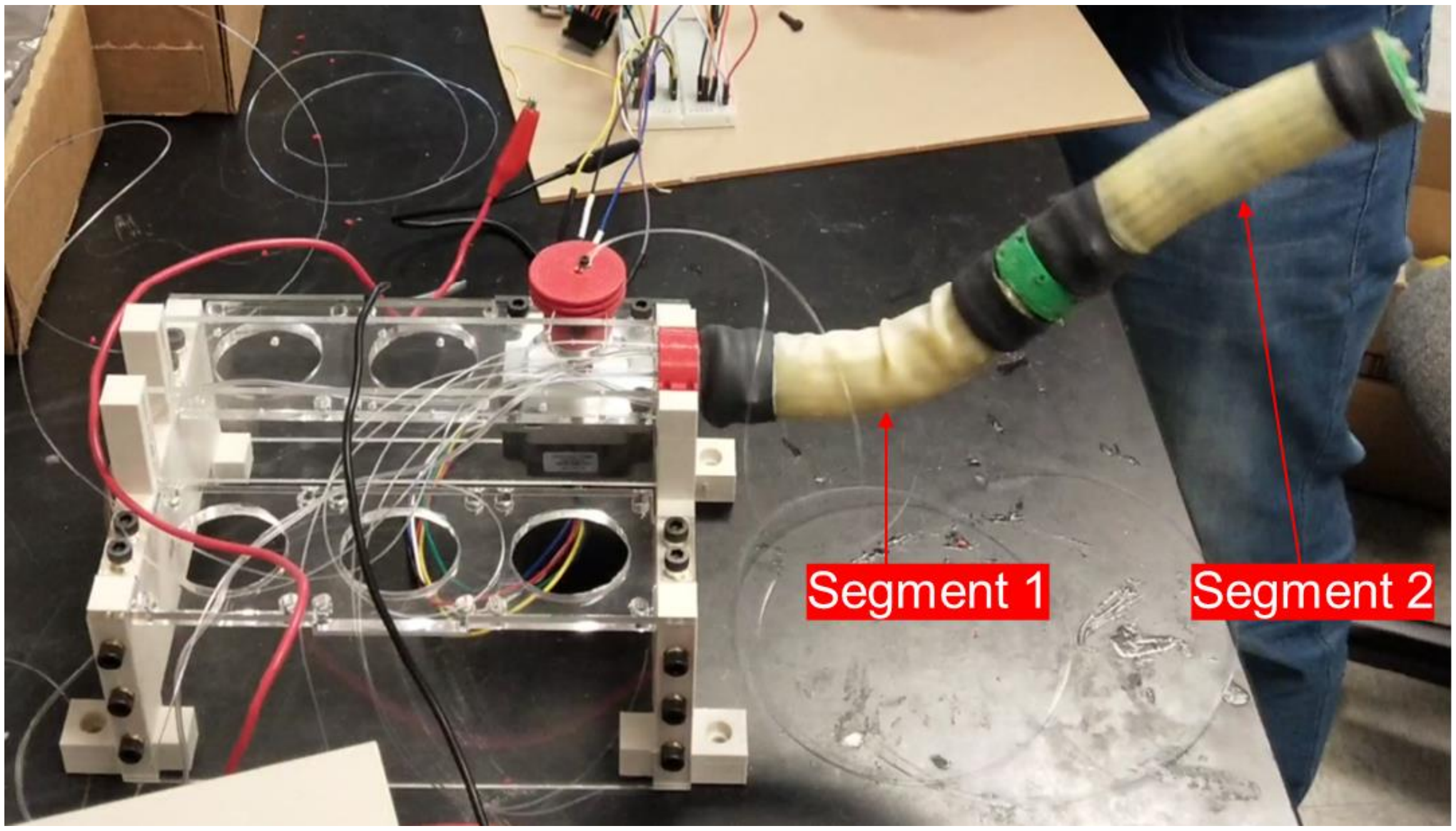

*Figure 4-1: Actuation Test for Planar Bending of Segment 1*

### 4.1.2 Testing of Spatial motion actuation of a segment 1

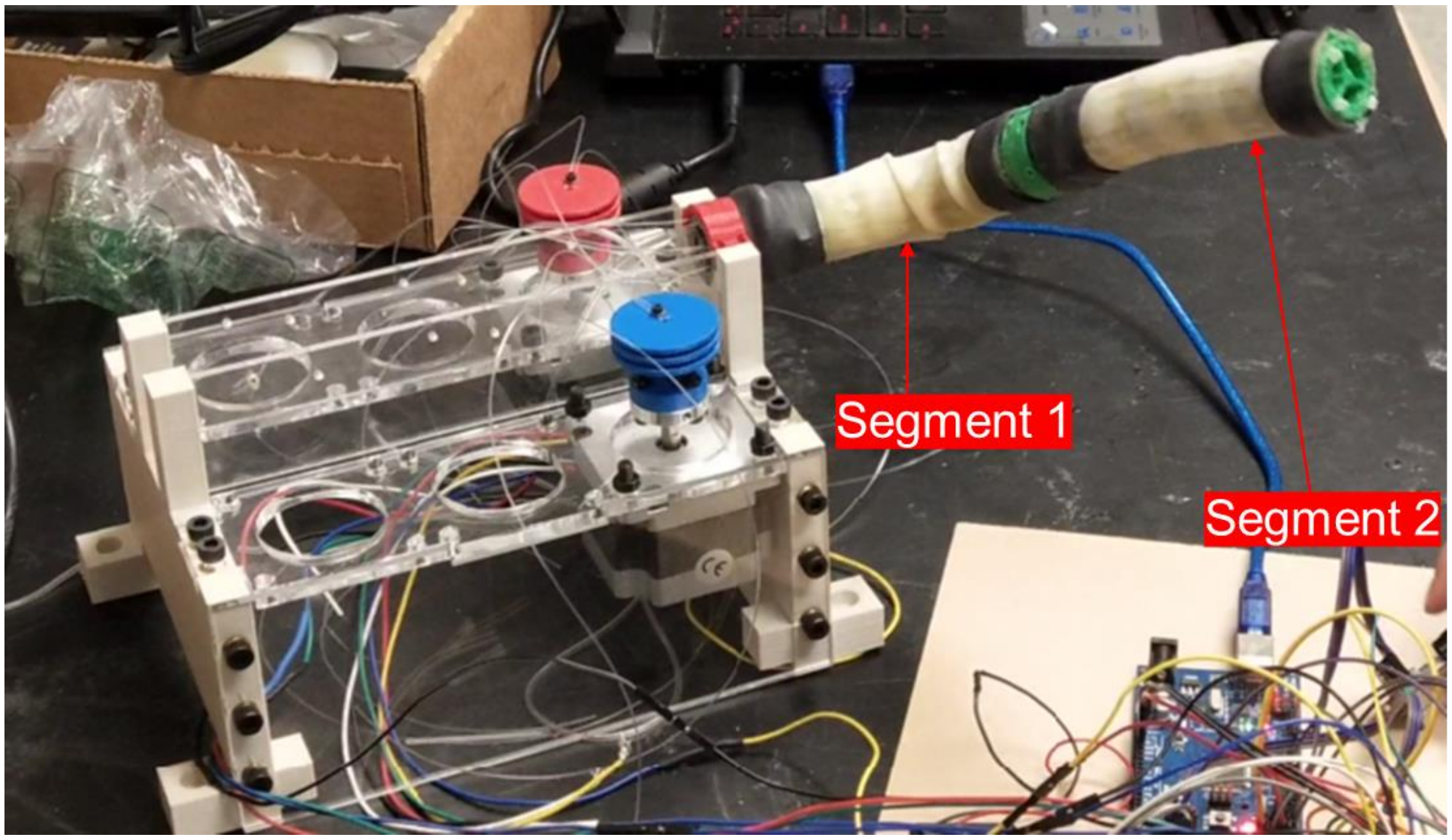

*Figure 4-2: Actuation Test for Spatial Bending of Segment 1*

After the success of the planar bending test, the second actuator was installed on the dock with the spool attached to the X-Tendon pair of segment 1. The spatial motion of the segment 1 can be actuated by 2 tendon pairs in this case. As is shown in Figure 4-2, when the 2 orthogonal tendon pairs are actuated simultaneously by rotating knob 1 and 2, segment 1 is bended within the 3D space, with components of bending in the horizontal plane ZX and the vertical plane ZY at the same time. Hence, the superposition rule for special bending actuation of a single segment should be valid.

### 4.1.3 Actuation Test for Segment 2

After the testing of segment 1, additional 2 actuators were mounted on the dock with spools connected with X-Tendon pair and Y-Tendon pair of segment 2 as is presented in Figure 4-3 below. In this section of test, segment 2 was actuated by rotating knob 3 and knob 4 while segment 1 was not actuated.

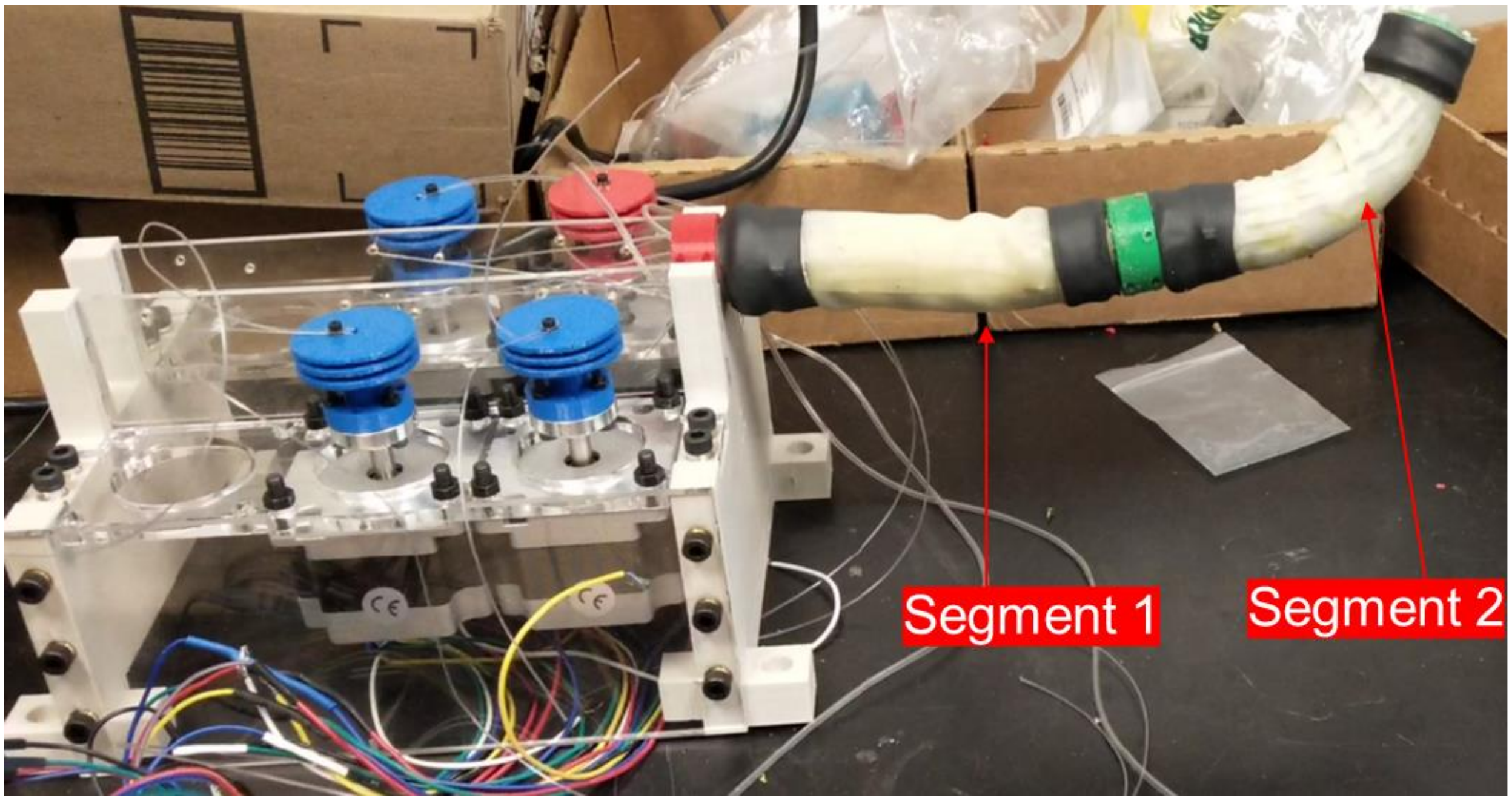

*Figure 4-3: Actuation Test for Segment 2 Of the Robotic Arm*

As is shown on Figure 4-3, segment 2 was actuated to have a bending angle while segment 1 remained horizontal and was not actuated. When manipulating knob 3 and 4 without rotating knob 1 and 2, segment 2 was actuated to have a spatial motion while segment 1 was stationary. This indicates that the system of displacement equations for actuation of the 4 tendon pairs is effective for decoupling the actuation of segment 1 and segment 2 of the continuum robot arm.

### 4.1.4 Testing of Spatial Motion for the 2-Segmented Robot Arm

In this section of test, the setup is the same as that of the test for segment 2. The test shows that segment 1 and segment 2 can be actuated simultaneously when rotating knobs 1,2,3,4 at the same time. Figure 4-4 below shows that the continuum robotic arm formed a S-shaped curved when manipulating the motion of each arm segment simultaneously through the 4 knobs.

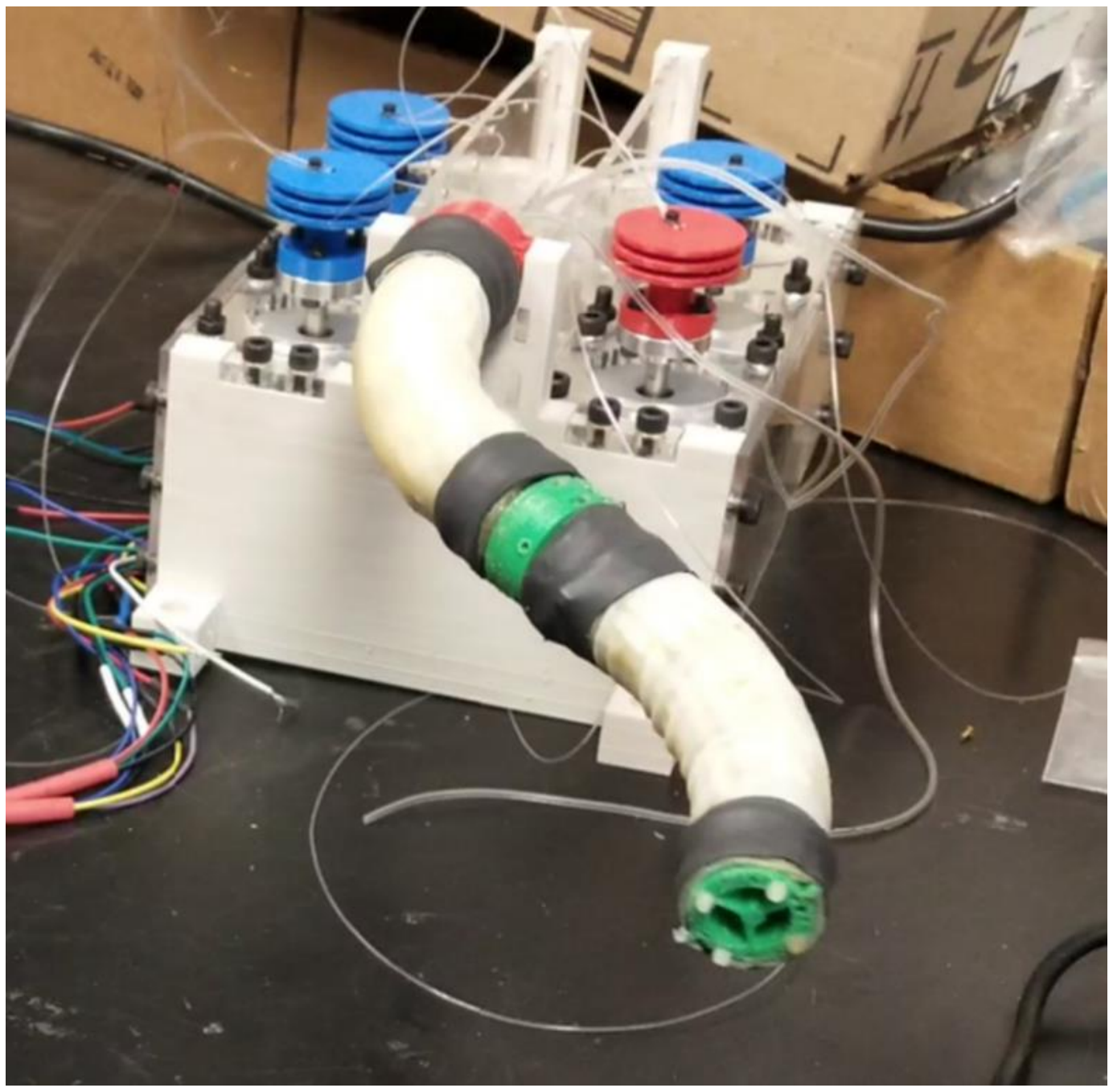

*Figure 4-4: Simultaneous Actuation of Segment 1 And 2 of the Continuum Robotic Arm*

## 4.2 Quantitative Actuation Test

In chapter 3, the proportional correlation between tendon pair displacement and bending angle was proposed to dictate the actuation of segments of the variable stiffness continuum robot. In this section of test, a quantitative actuation test was conducted to verify this proportional kinematic relationship between the tendon pair displacement and bending angle of the segment under the case of planar bending. The layout of the experimental setting is presented in Figure 4-5 as follows.

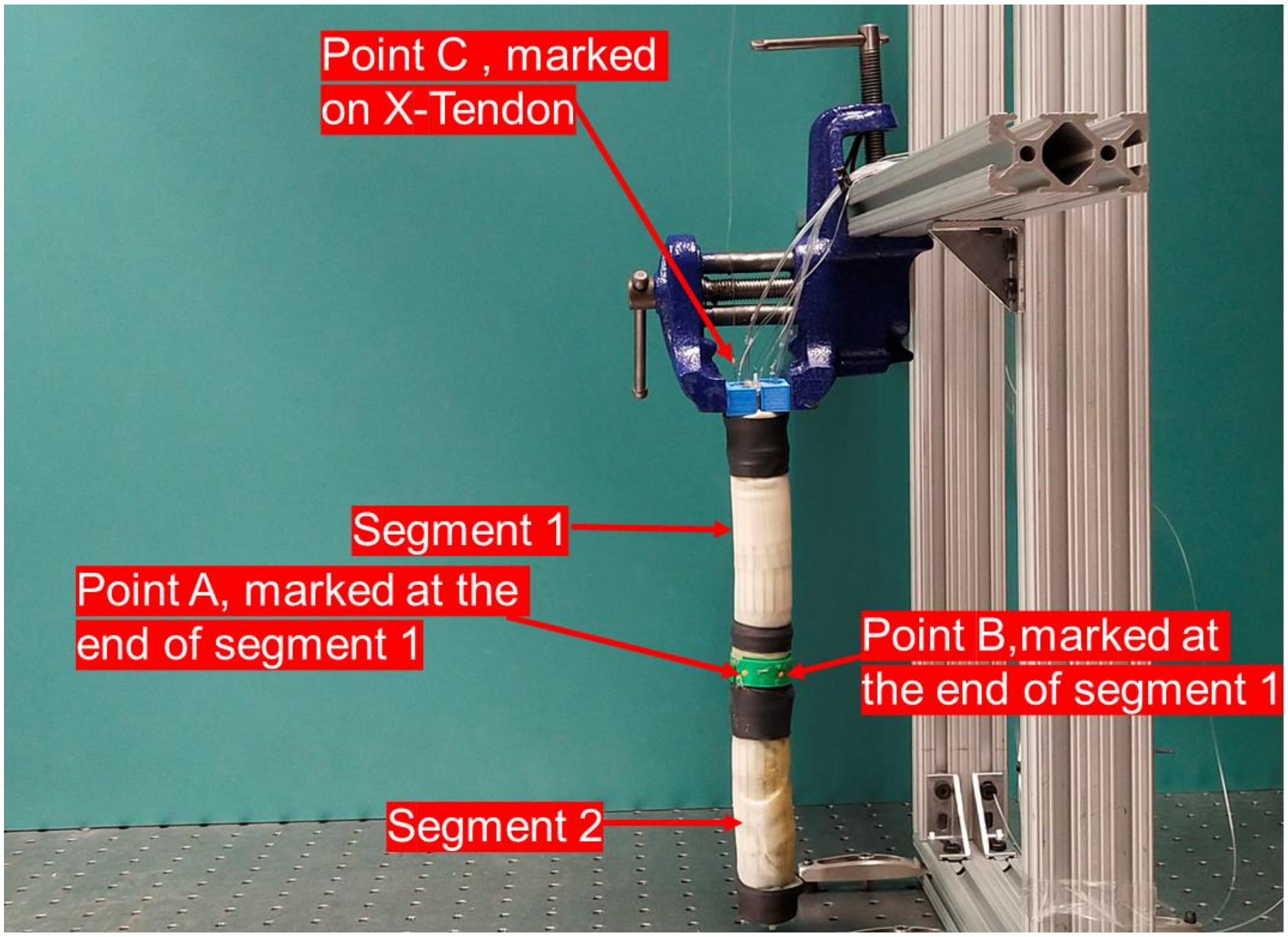

*Figure 4-5: Layout of the Quantitative Actuation Test for Segment 1 of the Robotic Arm*

As is shown on Figure 4-5, the robotic arm was fixed on the frame using a clamp. To observe the motion of segment 1 and tendon displacement, two points including point

A and B were marked at the distal end of segment 1, and point C was marked on X-Tendon. Then tendon was pulled by hand to actuate segment 1 to bend. As the same time, one camera was placed in front of the experimental setting to capture the motion of all marked points. The bending angle of segment 1 can be captured using the following equations including

Slope of line AB at each position during motion: $$k = \frac{y_A - y_B}{x_A - x_B} \quad (10)$$

Bending angle: $$\theta = \arctan(\mathrm{k}) \quad (11)$$

where $x_A$, $y_A$ , $x_B$, $y_B$ are coordinates of marked points A and B of segment 1 during the experiment, and $\theta$ represents the bending angle of segment 1 which varied during the test.

After the capturing the whole process of motion of segment 1, the video was imported to Tracker, which is a motion analysis software based on camera recordings, to obtain the trace and coordinates of the marked points on Segment 1 during the data acquisition process. Figure 4-6 in the next page shows the interface of Tracker when processing the experimental video to acquire coordinates of marked points at each time.

After obtaining coordinates of marked points, the bending angle of segment 1 at each time during the experiment was calculated using equations (10) and (11), and the plot of planar bending angle of segment 1 versus tendon displacement is presented in Figure 4-7 in the next page.

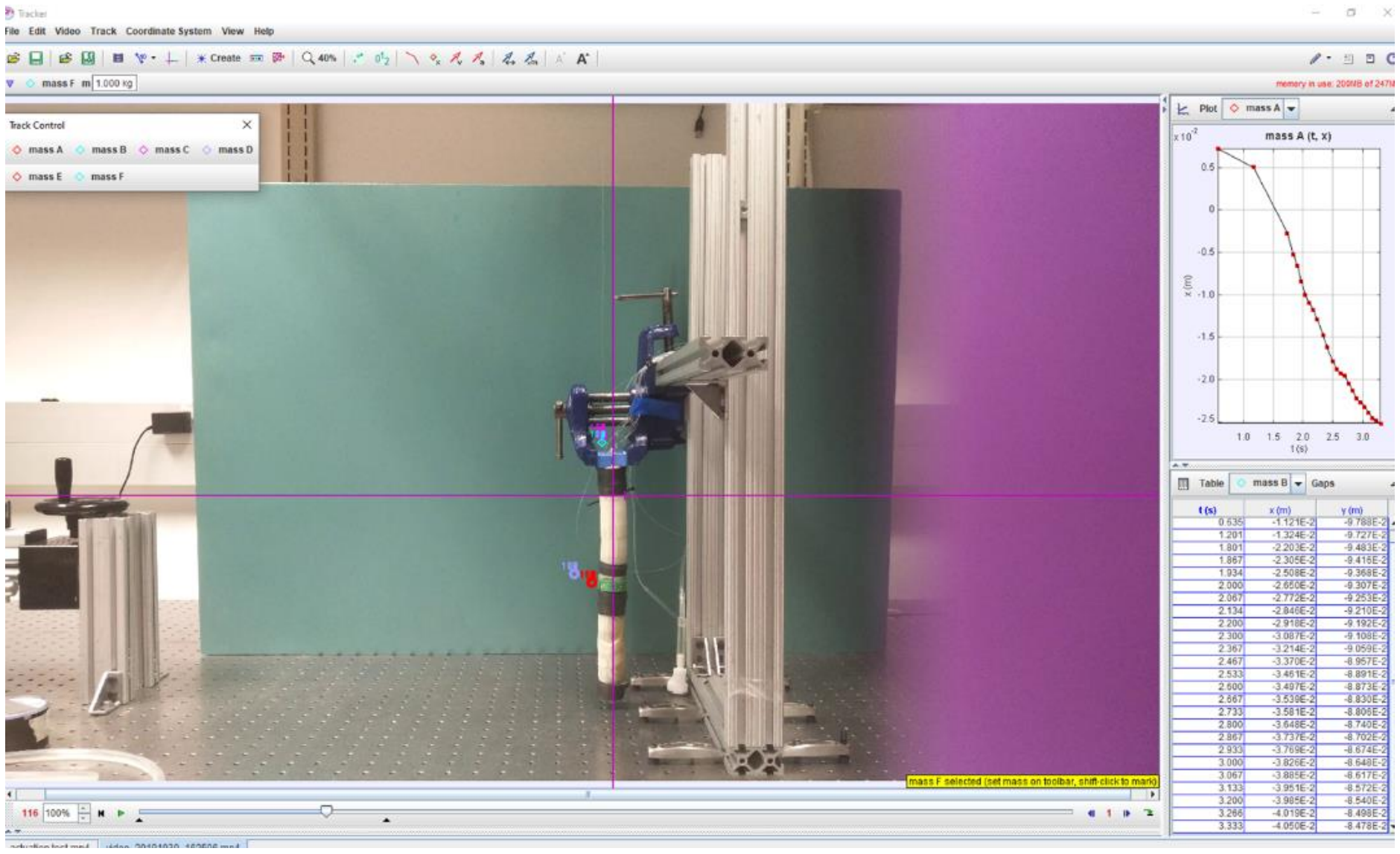

*Figure 4-6: Interface of Motion Analysis Software Tracker*

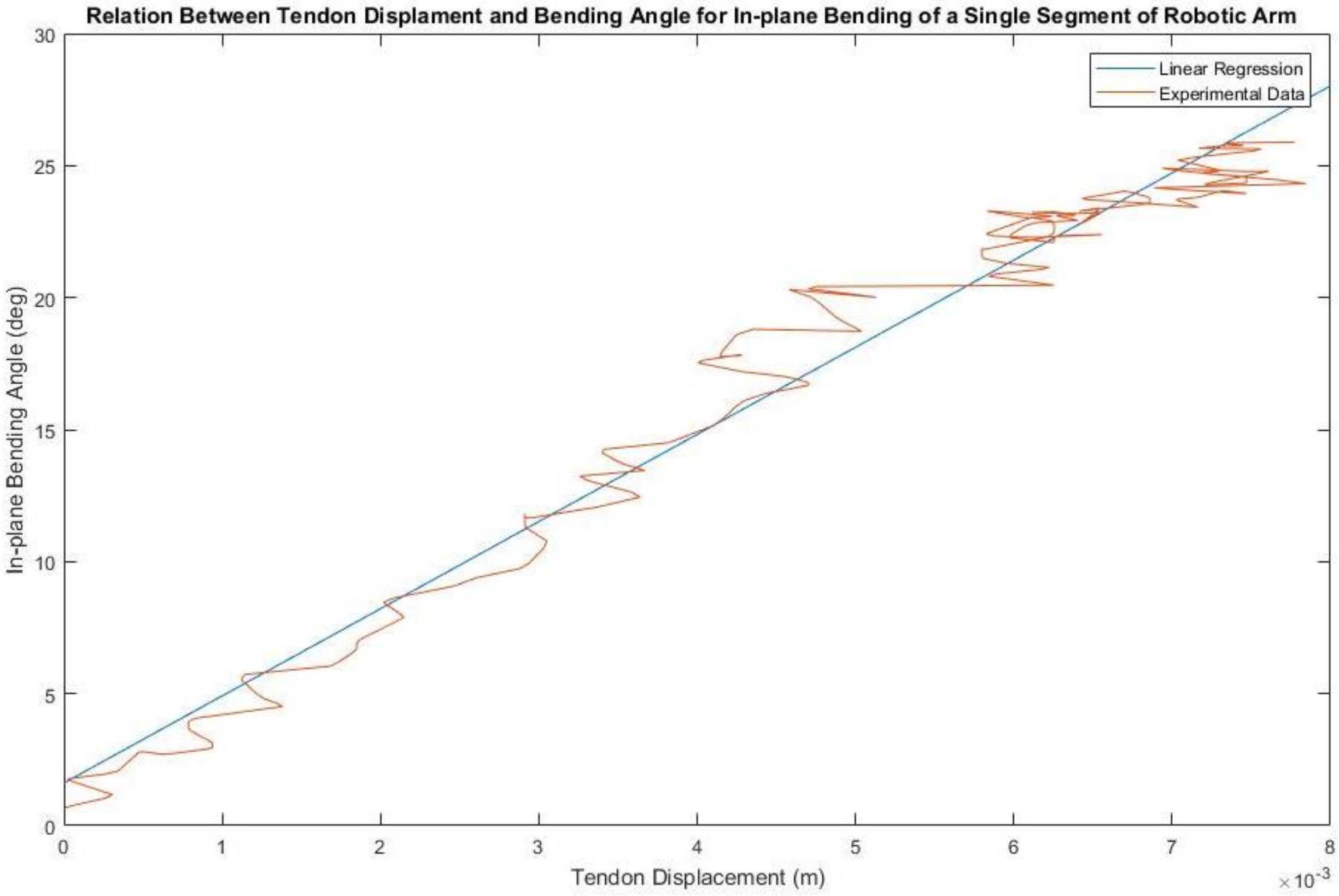

*Figure 4-7: Relation Between Tendon Displacement and Bending Angle for In-plane Bending of a Single Segment of Robotic Arm*

As is shown in Figure 4-7, the fitting line of the experimental plot of bending angle versus tendon displacement was plotted using linear regression. Apparently, the fitting line is quite consistent with the corresponding experimental plot. Therefore, it is verified that the bending angle is proportional to tendon displacement. The experimental relation between planar bending angle and tendon displacement was calculated to be

$$\theta = 3301.1\,x + 1.604 \qquad (12)$$

where x [m] is the displacement of the tendon pair, and $\theta$ [deg] is the bending angle.

In linear fitting equation (12) above, the slope of 3301.1 deg/m has a 95% confidence bound of (3246,3356), while the constant of 1.604 deg has a 95% confidence bound of (1.317, 1.911).

## 4.3 Load Capacity Test

The load capacity test was conducted to measure the load capacity of the continuum robotic arm with respect to the layer jamming pressure under 2 different cases. In case 1, the load was applied at the end of segment 2 to measure the load capacity of the overall structure of the robotic arm at different vacuum pressures of the layer jamming system. In case 2, the load was applied at the end of segment 1, which is the connector, to measure the load capacity of the segment 1 only at different vacuum pressures applied. In this experiment, the load capacity was measured under a series of vacuum pressures introduced for the layer jamming system ranging from 0 psi to 12 psi. The introduction of 0 psi inside the vacuum bag corresponds to the original unstiffened state of the robot arm. The experimental setup of the load capacity test is presented in Figure 4-5 below.

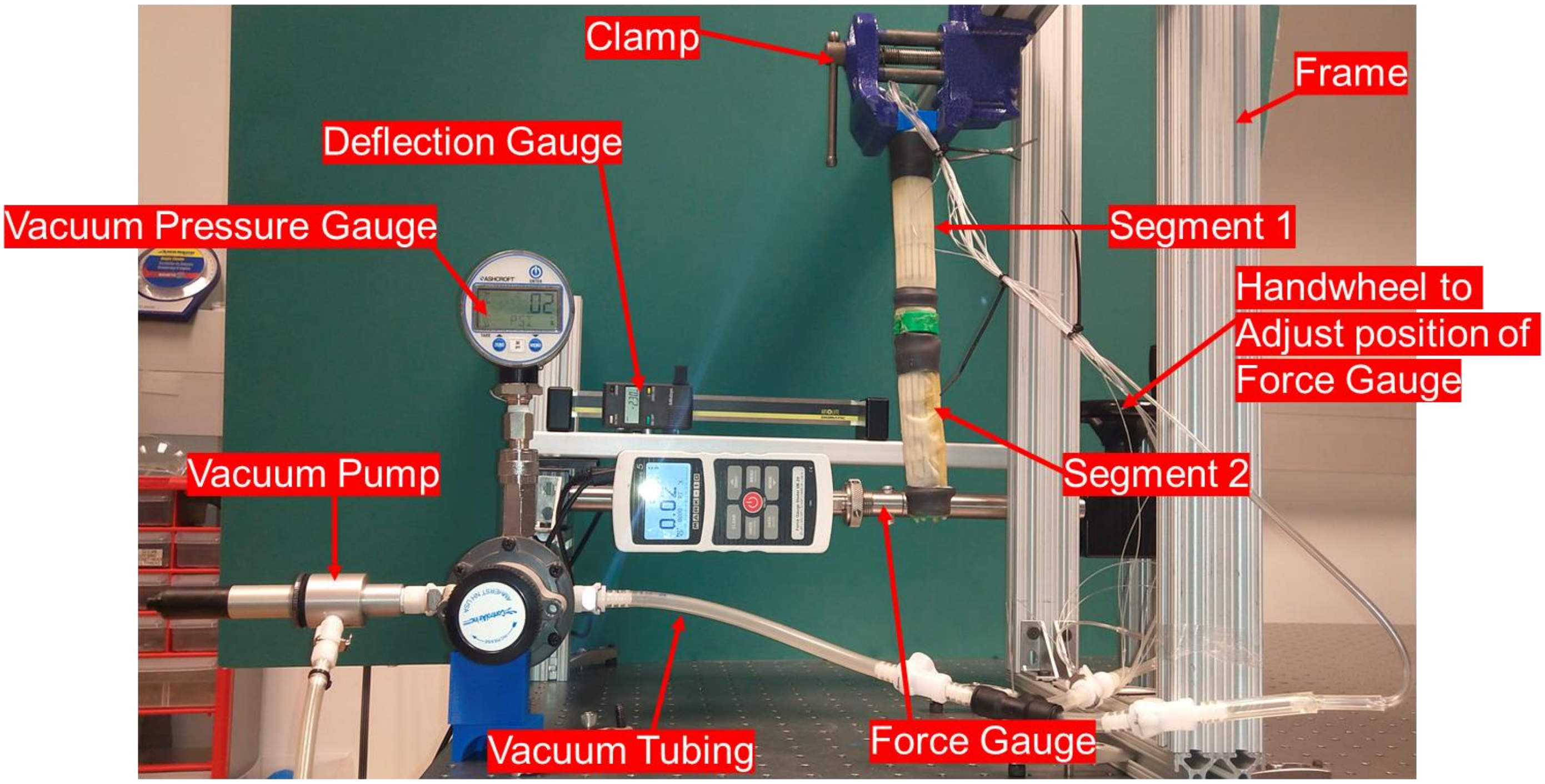

*Figure 4-8: Setup for The Load Capacity Test of The Overall Structure of the 2-Segmented Robot Arm*

As is shown in Figure 4-8, the deflection of loading point on the robot arm was measured by the deflection gauge while the load was measured by the force gauge. Load was applied by rotating the handwheel to cause the load point displacement. The 2 segments of robot arm were connected to the vacuum pump via the vacuum tubing as is shown in Figure 4-8.

During the experiment, the load and corresponding displacement was recorded from the deflection of 0 to the maximum deflection of 20 mm, with an increment of 1 mm deflection. Then plots of load versus loading point deflection under different vacuum pressure were created. From the plots of load versus deflection, plots of load capacity versus vacuum pressure were created. In this experiment, the load capacity of the continuum robotic arm corresponds to the load at the maximum deflection of 20 mm. In addition, the curve of stiffness ratio versus vacuum pressure was plotted to reveal the relation between negative pressure and stiffness enhancement of layer jamming.

### 4.3.1 Load Capacity Test of the 2-Segmented Robotic Arm

In this case, the load was applied at the end of segment 2 to measure the load capacity of the overall structure of the robotic arm at different vacuum pressures of the layer jamming system. Both segments were layer jammed under the same vacuum pressure during the experiment. The setup is already illustrated in Figure 4-8. The results are presented in Figure 4-9, 4-10, 4-11 respectively as follows.

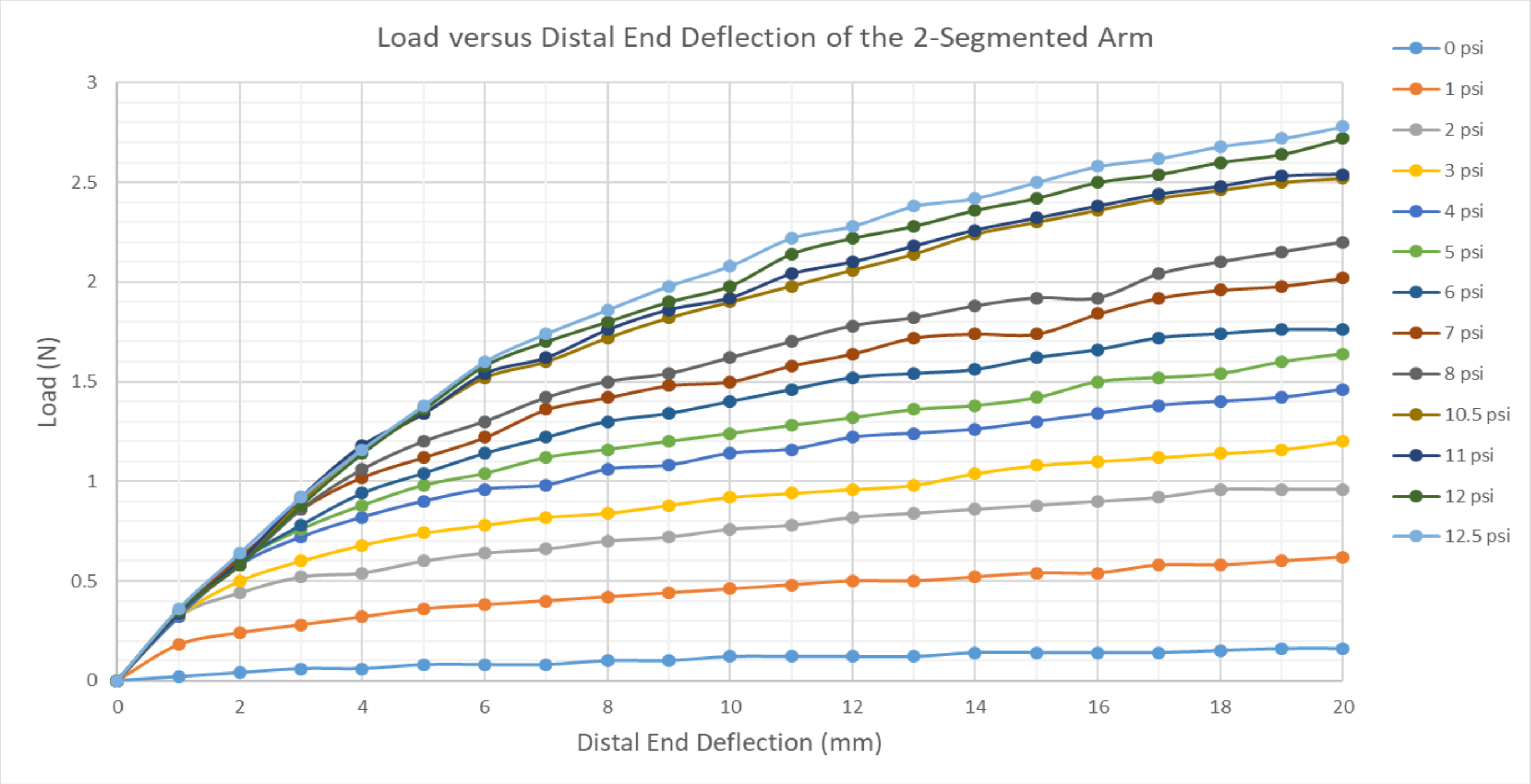

*Figure 4-9: Load Versus Distal End Deflection of the 2-Segmented Arm*

Figure 4-9 is the plot of load versus deflection at the distal end of segment under different vacuum pressure ranging from 0 to 12.5 psi. As is shown in this plot, the load increased quickly and linearly when the distal end deflection is sufficiently small, while the increase of load slowed down as the loading point deflection reached relatively larger values. There is another obvious trend on the plot that the load carried by the 2 segmented arm improved significantly as the vacuum pressure increased from 0 psi to 12.5 psi.

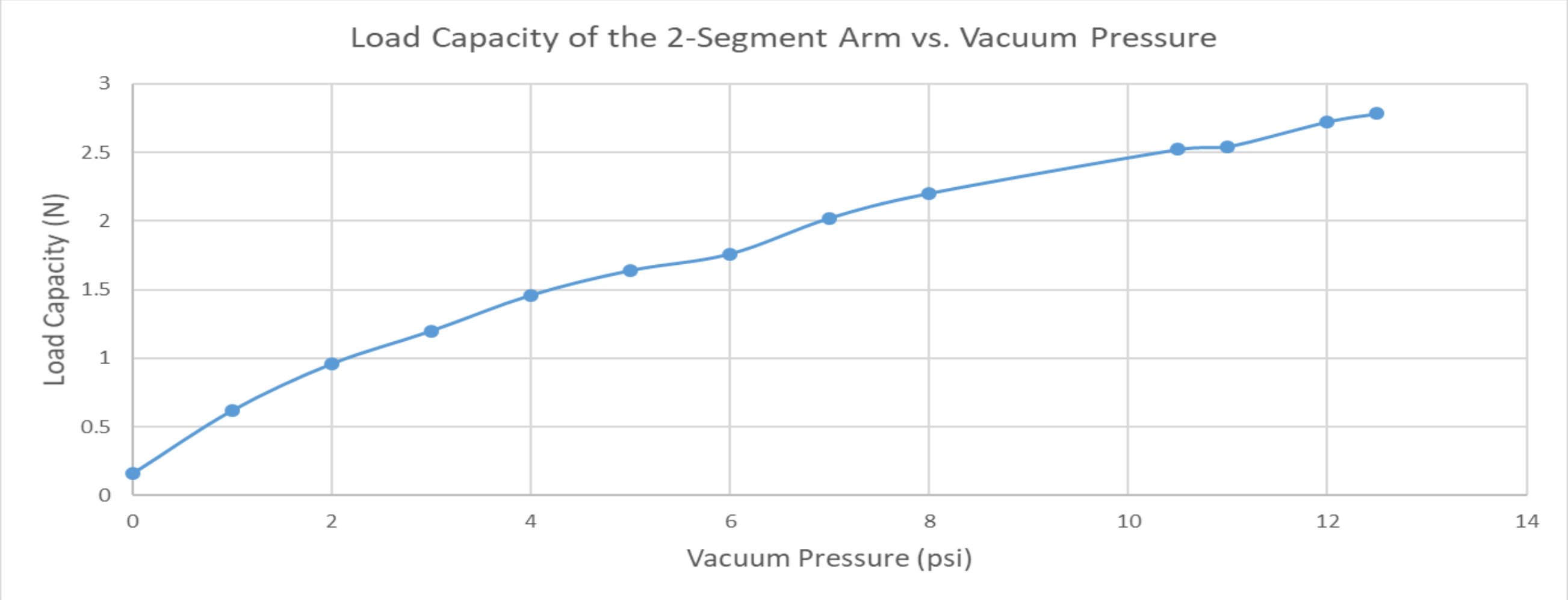

*Figure 4-10: Load Capacity of the 2-Segmented Arm Versus Vacuum Pressure*

Figure 4-10 above reveals the relation between the load capacity at the distal end of segment 2 and vacuum pressure of the layer jamming system. In this experiment, the load capacity is defined as the load carried by the arm under a 20 mm deflection. As is shown on Figure 4-10, the load capacity of the 2-segmented robot arm increased steadily from 0.2 N to 2.7 N as the vacuum pressure increased from 0 to 12.5 psi.

Next, the load capacity of the robotic arm was normalized to calculate the stiffness ratio by dividing the value by the original stiffness when zero pressure was applied. Figure 4-11 below reveals the relation between stiffness ratio of the 2-segmented robotic arm and vacuum pressure. As the vacuum pressure increase from 0 to 12.5 psi, the stiffness ratio also increased steadily from 1 to 17.5, which means that the stiffness of the variable stiffness continuum robotic arm can be 16.5 times larger than the counterpart at the original unstiffened state.

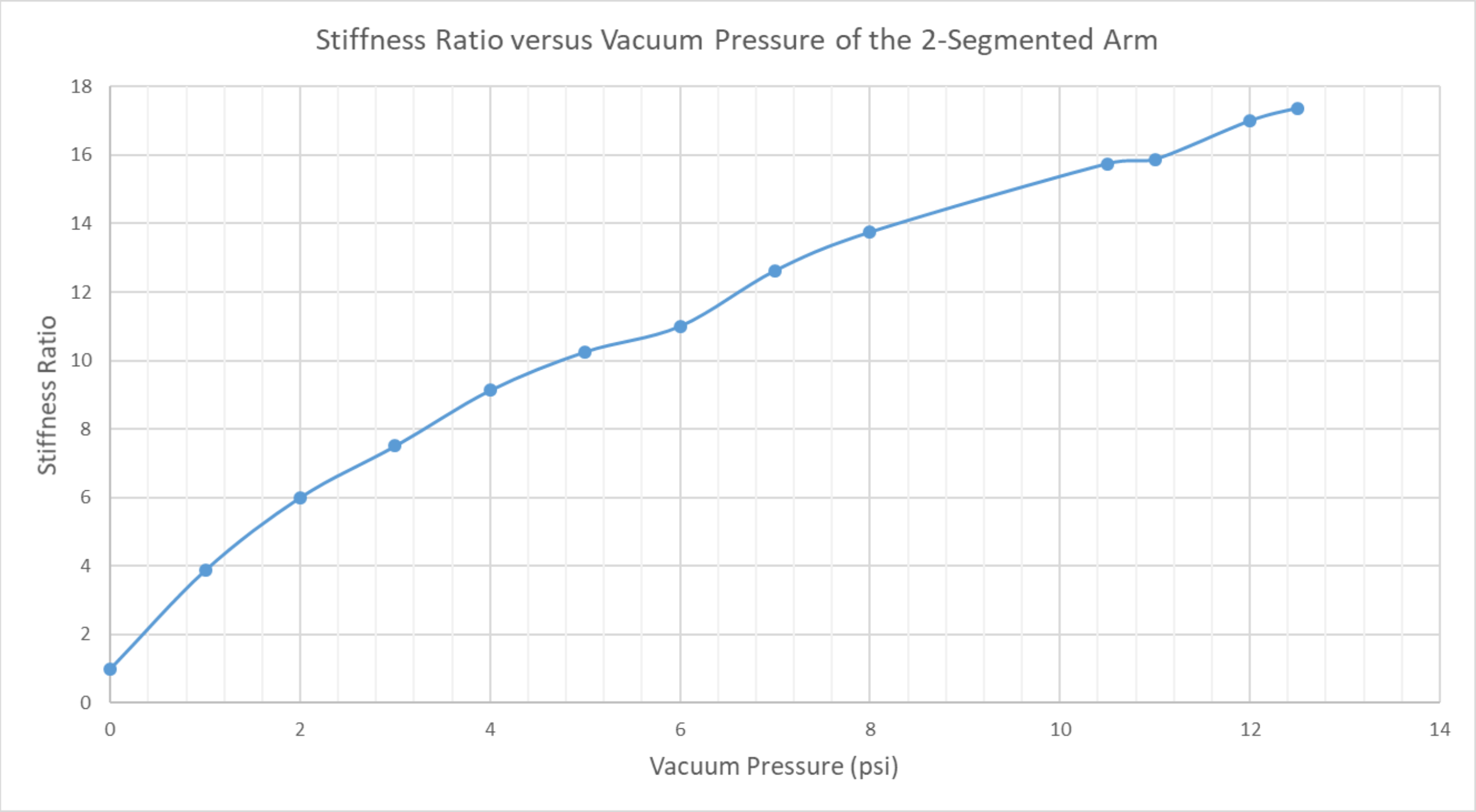

*Figure 4-11: Stiffness ratio of the 2-segmented robot arm versus vacuum pressure*

### 4.3.2 Load Capacity Test for Segment 1 Of the Robotic Arm

For load capacity test of segment 1 for the continuum robotic arm, the setup is almost the same as the test for the 2-segmented arm with both segments layer jammed, except for that the loading point is at the distal end of segment 1 rather than segment 2, as is shown on Figure 4-12 of the following page.

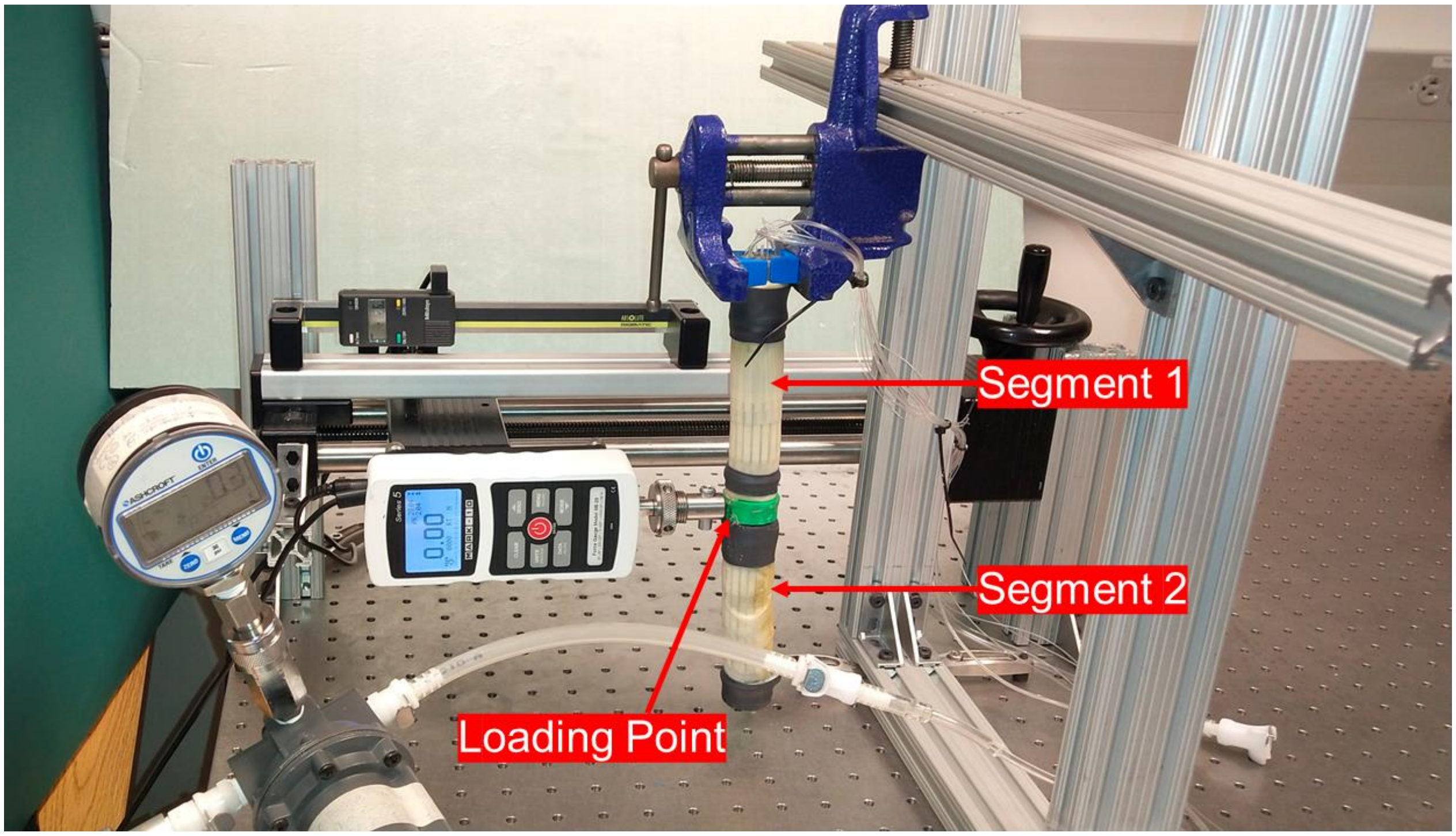

*Figure 4-12: Load Capacity Testing Setup for Segment 1 Of the Robotic Arm*

Results of the load capacity test are available in Figure 4-13, 4-14, 4-15 on the following pages. Figure 4-13 is the plot of load versus deflection at the distal end of segment under different values of vacuum pressure ranging from 0 to 12.5 psi. The trend of the plot is almost the same as the counterpart from the load capacity test of the 2-segmented robotic arm. However, the load withstood by the segment 1 is much higher at each vacuum pressure compared with the case where the load was applied on the tip of segment 2. The maximum load can reach 10 N at 12.5 psi in this case.

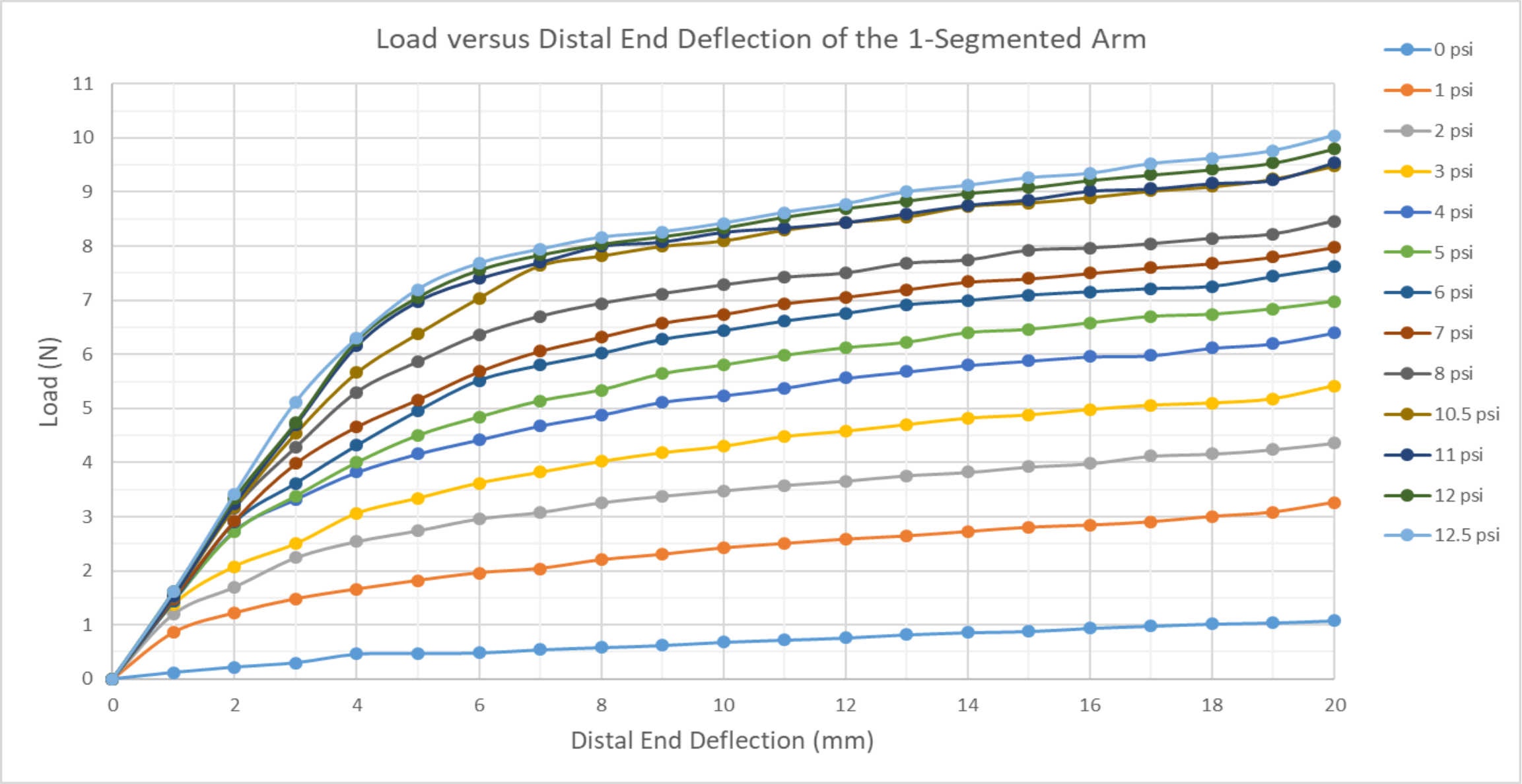

*Figure 4-13: Load Versus Distal End Deflection of The Segment 1 Of the Robotic Arm*

Figure 4-14 below shows the relation between the load capacity at the distal end of segment 2 and vacuum pressure of the layer jamming system. As is indicated on Figure 4-14, the load capacity of segment 1 increased steadily from 1 N to 10 N as the vacuum pressure increased from 0 to 12.5 psi.

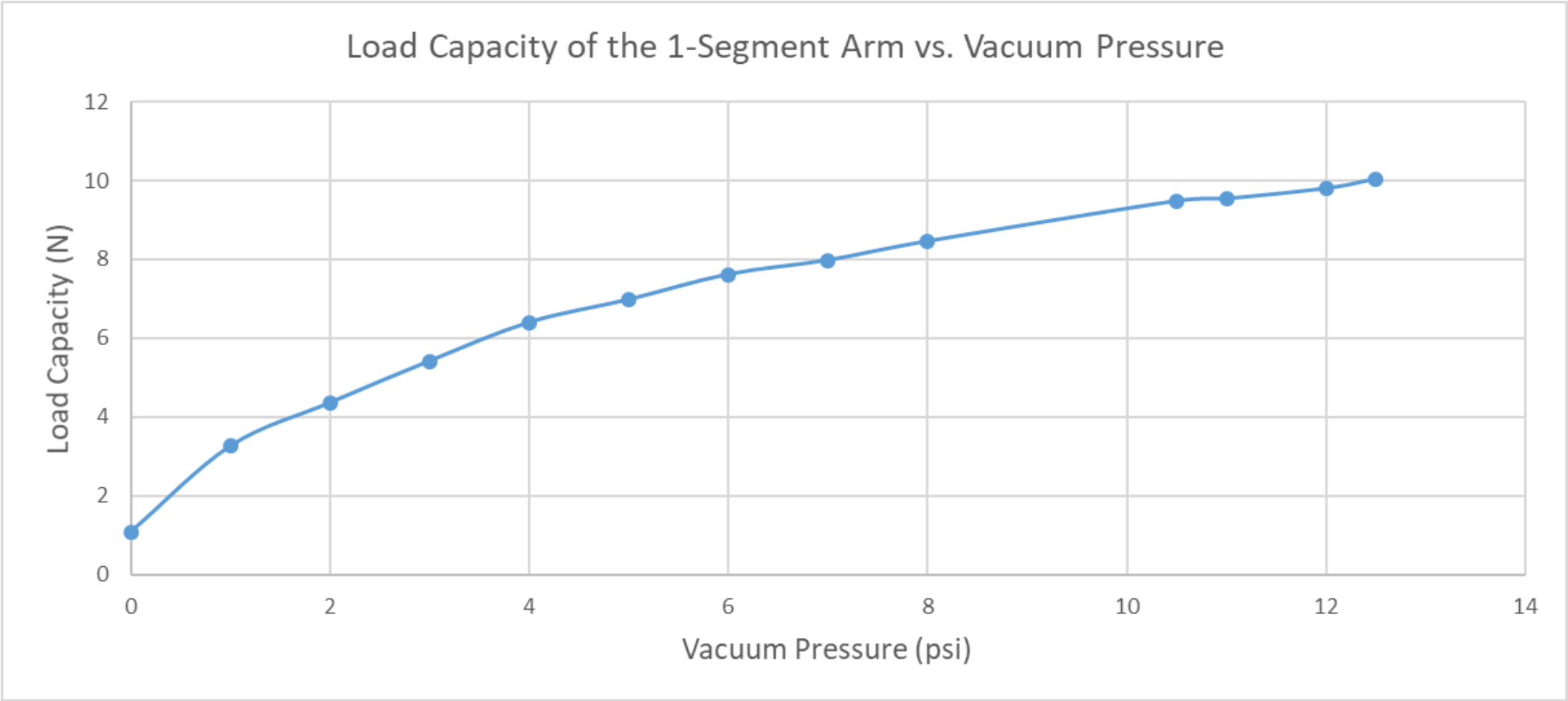

*Figure 4-14: Load Capacity Versus Vacuum Pressure of Segment 1 Of the Continuum Robotic Arm*

Figure 4-15 below reveals the relation between stiffness ratio and vacuum pressure of segment 1. As the vacuum pressure increase from 0 to 12.5 psi, the stiffness ratio also increased steadily from 1 to 9.2, which means that the stiffness of the segment 1 can be improved by 8.2 times via layer jamming.

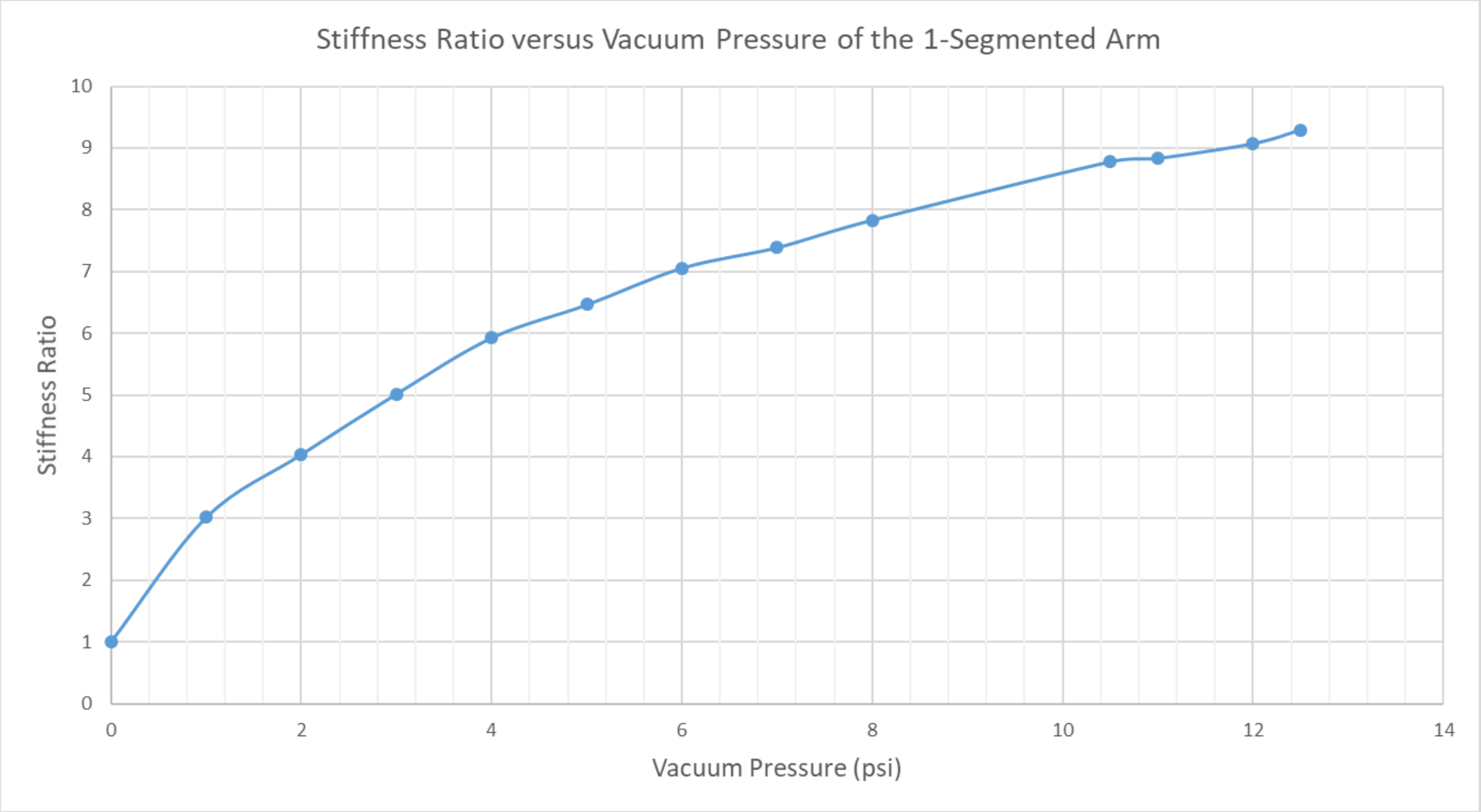

*Figure 4-15: Stiffness Ratio of Segment 1 Versus Vacuum Pressure*

### 4.3.3 Load Capacity Test with Standard Weights

In this section, standard weights were utilized to verify the load capacity of the variable stiffness continuum robotic arm with tendons unconnected to the actuators. As is shown in Figure 4-16 (a), the stiffness of the robotic arm at the unjammed state is extremely small, and the structure is quite compliant and cannot even resist deformation under gravity. After the introduction of 12.5 psi negative pressure inside both segments, the robot arm had sufficient stiffness to maintain its shape and resist gravity, as is shown in Figure 4-16 (b).

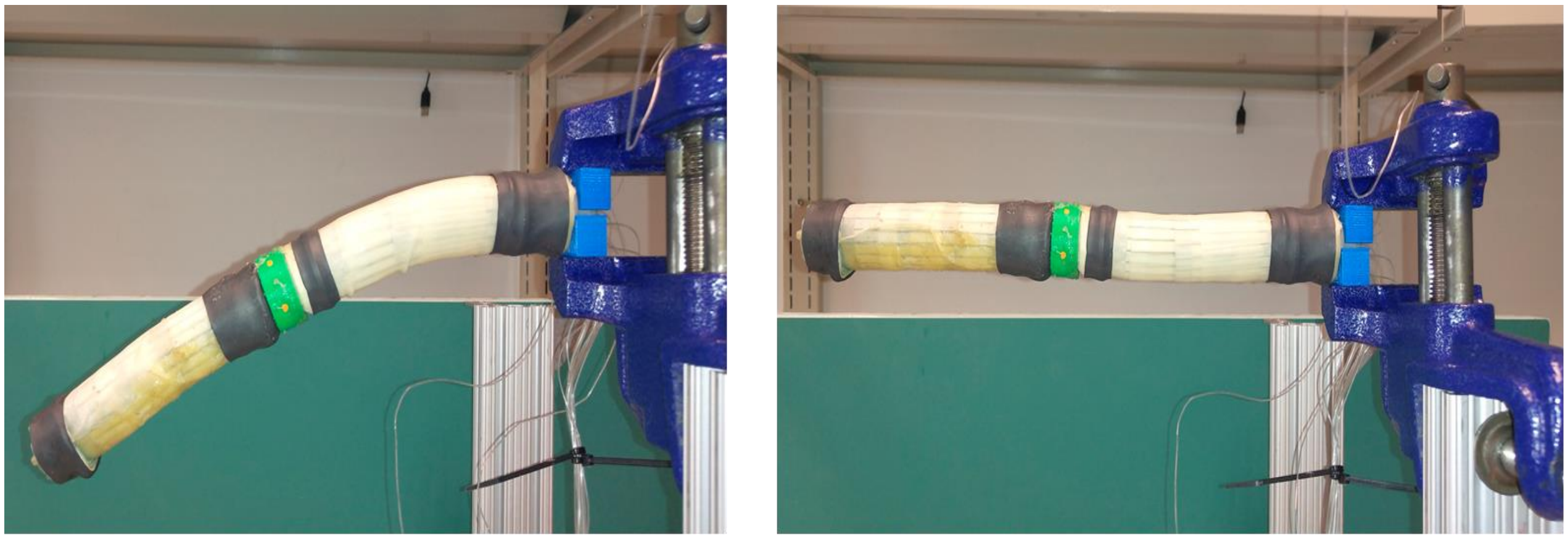

(a) Unjammed (b) Jammed with 12.5 psi Negative Pressure

*Figure 4-16: The Robot Arm Under Unjammed State and Jammed State Respectively*

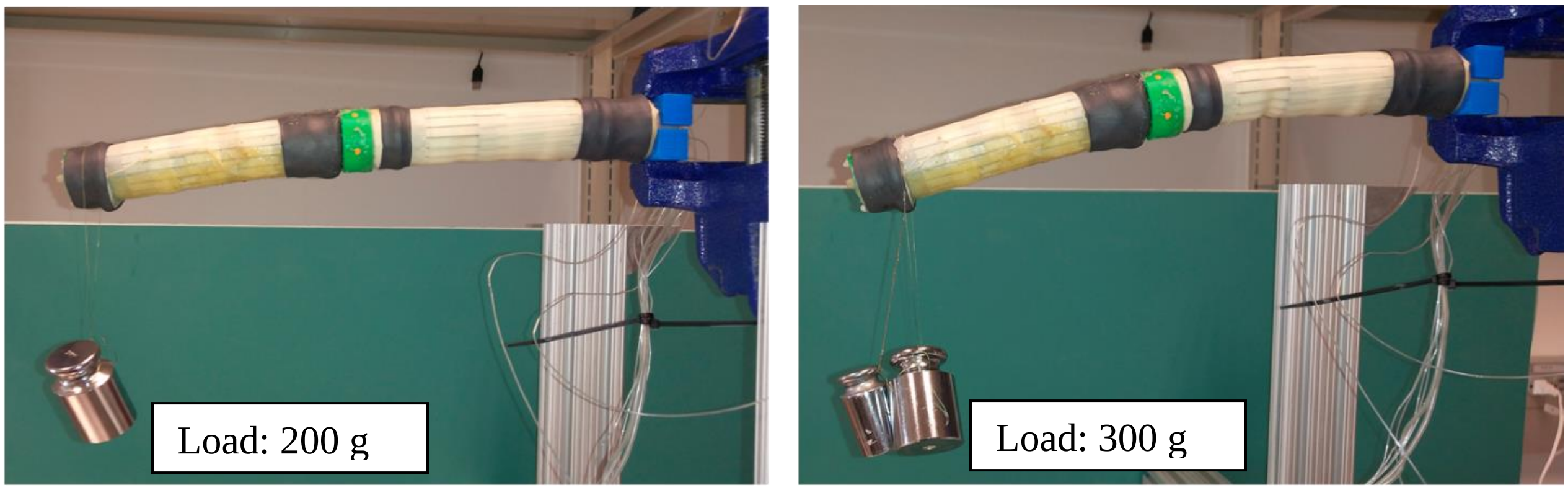

(a) Layer Jammed Structure Loaded with 200 g of Weight at the Distal End of Segment 2

(b) Layer Jammed Structure Loaded with 300 g of Weight at the Distal End of Segment 2

*Figure 4-17: Qualitative Load Capacity Test When Load Applied at The End of Segment 2*

To verify the stiffness of the robot arm at the end of segment 2, a weight of 200 grams was hung at the distal side of the robotic arm, as is shown on Figure 4-17 (a). The experiment showed the structure can withstand the 200 g weight. The structure can also withstand a load of 300 grams but with a larger tip deflection as is presented on Figure 4-17 (b).

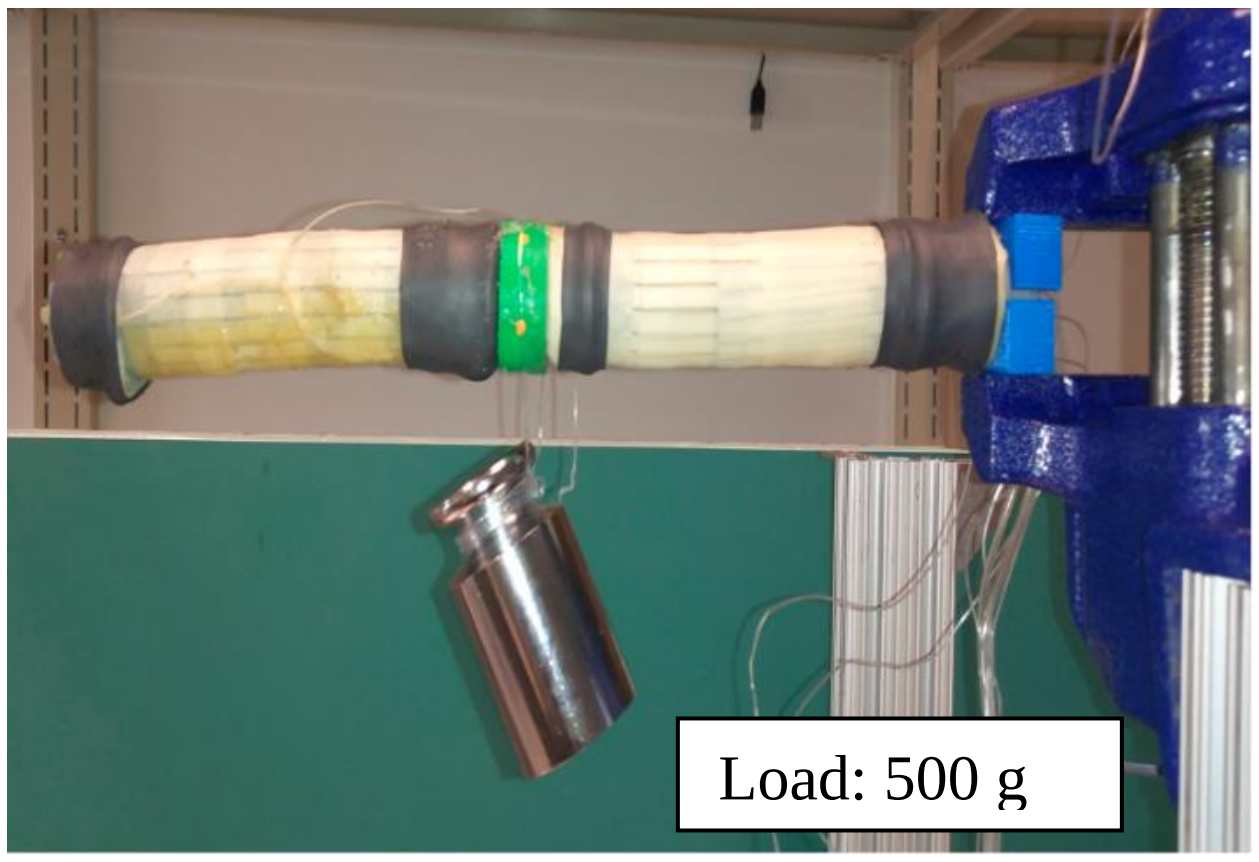

(a) Layer Jammed Structure Loaded with 500 g of Weight at the Distal End of Segment 1

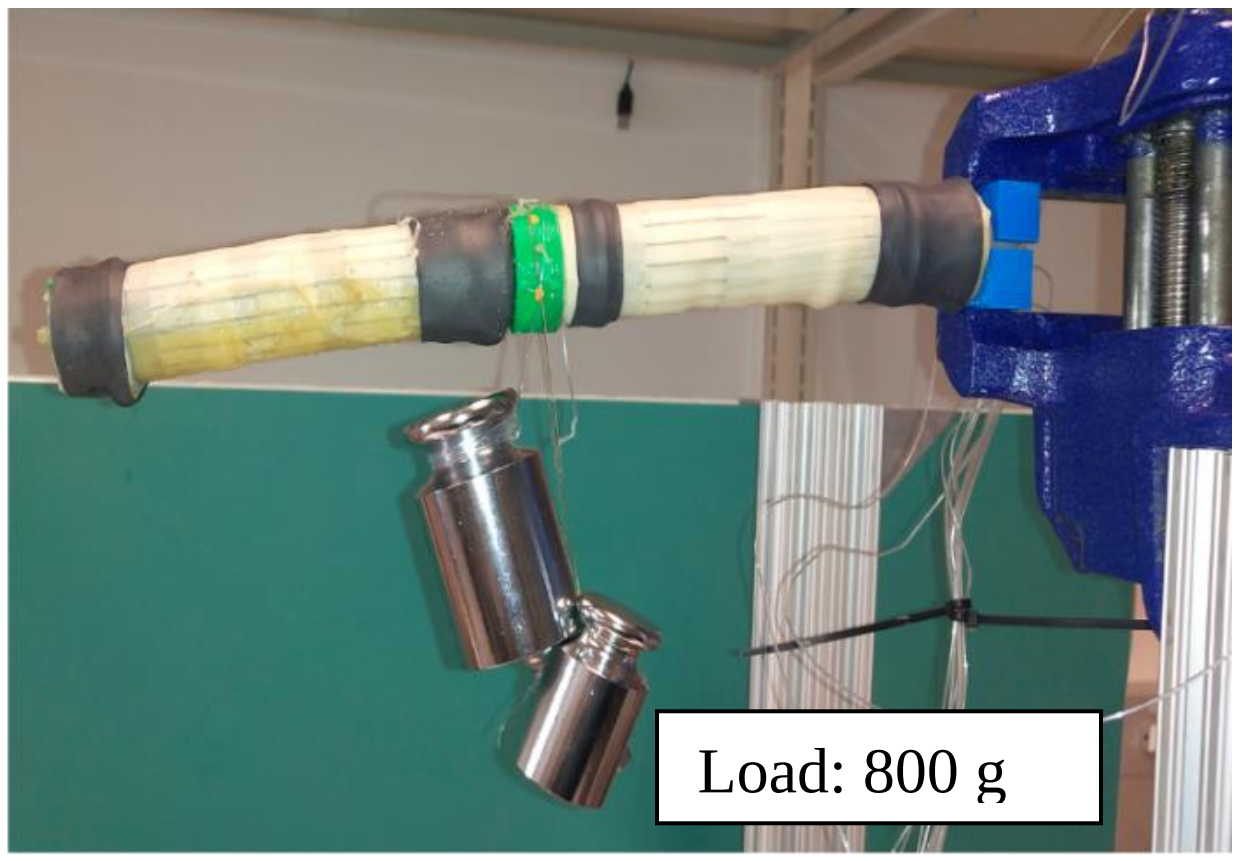

(b) Layer Jammed Structure Loaded with 800 g of Weight at the Distal End of Segment 1

*Figure 4-18: Qualitative Load Capacity Test for Segment 1*

Next, weights were attached to the connector to examine the load capacity of segment 1. As is depicted in Figure 4-18 (a), segment 1 remained quite rigid when a 500 g weight was attached to the green connector. The structure can also resist the 800 grams of weight exerted on the end of segment 1, while this caused a noticeable tip deflection as is presented in Figure 4-18 (b).

The result of this experiment indicates that the robot arm cannot withstand any load even if it is its own weight, while still able to have a relatively high load capacity of up to 800 grams after negative pressure was introduced inside the vacuum bags. Hence the layer jamming system is quite effective for stiffness enhancement of the compliant robot arm.

# Chapter 5 Conclusion

## 5.1 Contributions

In this soft robotics research, several technical approaches for variable stiffness and actuation of continuum robots were discussed, and the main purpose of this research is to design a continuum robot with modular structure, high dexterity while still possessing sufficient load carrying capacity. Based on this target, the design of a 3 segmented tendon driven variable stiffness continuum robot with layer jamming was proposed. The fabrication process was quite smooth, and the second arm segment was able to be integrated into an existing 1-segmented robotic arm. Therefore, the concept of modular design of arm segments was verified by the 2-segmented robotic arm prototype. The robotic arm of the final prototype has 2 segments driven by 2 orthogonal tendon pairs respectively.

Detailed fabrication process was also explained in this thesis. Then the author analyzed the kinematics of the multiple segmented compliant arm structure, to propose several equations to govern the coordinated translation of the tendon pairs to decouple the actuation of multiple arm segments which are connected in series. Finally, the author conducted load capacity tests and actuation tests both qualitatively and quantitatively, to evaluate the performance of load carrying and stiffness variation capacity of the variable stiffness continuum robotic arm.

## 5.2 Future Works

One issue with the current prototype is that the strength of the spacer disks of the compliant structure is not enough to withstand a load of more than 10 N, of which the reason is that the infill densify of the 3D printed part has a small value of 10%. In the future, the infill density of 3D printed parts installed on the prototype can be improved to 50% to ensure a higher load capacity of the variable stiffness robotic arm. Another method of improving the load capacity is to reduce the spacing between adjacent spacer plates of each arm segment, as the friction layers of the current arm segment buckled overtly when a heavy load is applied at the tip of the arm. The third approach for load capacity improvement is to design the compliant structure as a single body part which integrate the center backbone into the series of spacer disks. The current design of compliant main structure of each segment has less strength since the spacer disks are secured on the center backbone using glue.

The latex rubber vacuum bag of the current design does not have sufficient durability neither, since it is vulnerable to fatigue and rupture after the bending of each arm segment for over 100 times. It is imperative to find a vacuum bagging material which has enough strength, toughness and flexibility for arm segments to be fabricated in the future.

Since the actuation test conducted in this research so far has only studied the simplest scenario of planar bending for a single arm segment using one camera, future actuation tests will be focused on spatial continuum robot arm motion, by exploring the kinematics between the displacement of multiple tendon pairs and the corresponding shape and spatial motion of the robotic arm with 2 segments. To achieve this goal, more than 3 cameras will be deployed to capture and track the motion of key points on the antagonistic tendon pairs and robot arm segments.

## 5.3 Summary

In this research, a layer jammed tendon driven variable stiffness continuum robot with 3 segments was designed. Because of time constraint, a 2-segmented variable stiffness robotic system prototype was fabricated and tested. The robotic system consists of 2 arm segments connected in series and one actuation system formed by the dock and 4 stepper motors with spools. Testing results of the variable stiffness continuum robotic arm prototype showed that the 2 segments of the arm can actuated independently and serpentine within the 3D space. It was observed in experiment that the load capacity of the 2-segmented robot arm can be improved by 17 times at 12.5 psi through layer jamming with 2 friction layers in vacuum bag. The load capacity of the robotic arm with 2 segments was measured to be 2.7 N, which increased to around 10 N for a single segmented robotic arm. In addition, the qualitative actuation test verified that the in-plane bending angle for each segment is approximately proportional to displacement of the tendon pair.

# Appendix A.  Arduino Codes for Tendon Actuation

```
#include <Arduino.h>
#include "A4988.h"


// Stepper Motor 1 configuration    [21]
// using a 200-step motor (most common)
#define MOTOR_STEPS 200
// configure the pins connected
#define DIR 23
#define STEP 22
#define MS1 46
#define MS2 47
#define MS3 48
A4988 stepper1(MOTOR_STEPS, DIR, STEP, MS1, MS2, MS3);
float i=27.5;  // angle increment index
float alpha=1.00765 ;  // 1.00765,1.1085 Segment 1 Actuation Compensation Factor1
float beta=1;  //Compensation Factor2



float Ay2;
 float Ax2;
 float Ay1;
 float Ax1;

// Rotary endoder 1 setup
volatile boolean TurnDetected;  // need volatile for Interrupts
volatile boolean rotationdirection;  // CW or CCW rotation
const int PinCLK=3;   // Generating interrupts using CLK signal
const int PinDT=24;    // Reading DT signal
const int PinSW=25;    // Reading Push Button switch

const int interruptpin1=3 ; // Interrupt pin1 number

int RotaryPosition=0;    // To store Stepper Motor Position
```

```
int PrevPosition;     // Previous Rotary position Value to check accuracy
float AnglesToRotate=0;      // How much angle to move Stepper

int encodata;
 float Rx2;  // Segment 2 x rotation angle
 float Ry2;  // Segment 2 y rotation angle
 float Rx1;  // Segment 1 x rotation angle
float Ry1;  // Segment 1 y rotation angle

// Interrupt routine runs if CLK goes from HIGH to LOW
void isr ()  {
  delay(4);  // delay for Debouncing
  encodata=digitalRead(PinCLK);
  if (digitalRead(PinCLK))
    rotationdirection= digitalRead(PinDT);
  else
    rotationdirection= !digitalRead(PinDT);
  TurnDetected = true;
}



// Stepper Motor 2 configuration
// using a 200-step motor (most common)
#define MOTOR_STEPS 200
// configure the pins connected
#define DIR2 28
#define STEP2 29
#define MS1_2 49
#define MS2_2 50
#define MS3_2 51
A4988 stepper2(MOTOR_STEPS, DIR2, STEP2, MS1_2, MS2_2, MS3_2);
int i2=20;  // angle increment index




// Rotary endoder 2 setup
volatile boolean TurnDetected2;  // need volatile for Interrupts
volatile boolean rotationdirection2;  // CW or CCW rotation
const int PinCLK2=2;   // Generating interrupts using CLK signal
const int PinDT2=26;    // Reading DT signal
const int PinSW2=27;    // Reading Push Button switch
```

```
const int interruptpin2=PinCLK2 ; // Interrupt pin1 number

int RotaryPosition2=0;    // To store Stepper Motor Position

int PrevPosition2;     // Previous Rotary position Value to check accuracy
float AnglesToRotate2=0;      // How much angle to move Stepper

int encodata2;


// Interrupt routine runs if CLK goes from HIGH to LOW
void isr2 ()  {
  delay(4);  // delay for Debouncing
  encodata2=digitalRead(PinCLK2);
  if (digitalRead(PinCLK2))
    rotationdirection2= digitalRead(PinDT2);
  else
    rotationdirection2= !digitalRead(PinDT2);
  TurnDetected2 = true;
}


// Stepper Motor 3 configuration
// using a 200-step motor (most common)
//#define MOTOR_STEPS 200
// configure the pins connected
#define DIR3 31
#define STEP3 30
#define MS1_3 40
#define MS2_3 41
#define MS3_3 42
A4988 stepper3(MOTOR_STEPS, DIR3, STEP3, MS1_3, MS2_3, MS3_3);
int i3=20;  // angle increment index

// Rotary endoder 3 setup
volatile boolean TurnDetected3;  // need volatile for Interrupts
volatile boolean rotationdirection3;  // CW or CCW rotation
const int PinCLK3=18;   // Generating interrupts using CLK signal
const int PinDT3=32;    // Reading DT signal
const int PinSW3=33;    // Reading Push Button switch

const int interruptpin3=PinCLK3 ; // Interrupt pin1 number
```

```
int RotaryPosition3=0;    // To store Stepper Motor Position

int PrevPosition3;     // Previous Rotary position Value to check accuracy
float AnglesToRotate3=0;      // How much angle to move Stepper

int encodata3;


// Interrupt routine runs if CLK goes from HIGH to LOW
void isr3 ()  {
  delay(4);  // delay for Debouncing
  encodata3=digitalRead(PinCLK3);
  if (digitalRead(PinCLK3))
    rotationdirection3= digitalRead(PinDT3);
  else
    rotationdirection3= !digitalRead(PinDT3);
  TurnDetected3 = true;
}


// Stepper Motor 4 configuration
// using a 200-step motor (most common)
//#define MOTOR_STEPS 200
// configure the pins connected
#define DIR4 37
#define STEP4 36
#define MS1_4 43
#define MS2_4 44
#define MS3_4 45
A4988 stepper4(MOTOR_STEPS, DIR4, STEP4, MS1_4, MS2_4, MS3_4);
int i4=20;  // angle increment index

// Rotary endoder 4 setup
volatile boolean TurnDetected4;  // need volatile for Interrupts
volatile boolean rotationdirection4;  // CW or CCW rotation
const int PinCLK4=19;   // Generating interrupts using CLK signal
const int PinDT4=34;    // Reading DT signal
const int PinSW4=35;    // Reading Push Button switch

const int interruptpin4=PinCLK4 ; // Interrupt pin1 number

int RotaryPosition4=0;    // To store Stepper Motor Position

int PrevPosition4;     // Previous Rotary position Value to check accuracy
```

```
float AnglesToRotate4=0;      // How much angle to move Stepper

int encodata4;


// Interrupt routine runs if CLK goes from HIGH to LOW
void isr4 ()  {
  delay(4);  // delay for Debouncing
  encodata4=digitalRead(PinCLK4);
  if (digitalRead(PinCLK4))
    rotationdirection4= digitalRead(PinDT4);
  else
    rotationdirection4= !digitalRead(PinDT4);
  TurnDetected4 = true;
}


void setup() {
    // Set target motor RPM to 1RPM and microstepping to 1 (full step mode)
    stepper1.begin(180, 16);


   //Rotary encoder 1 setup
   pinMode(PinCLK,INPUT);
pinMode(PinDT,INPUT);
pinMode(PinSW,INPUT);
digitalWrite(PinSW, HIGH); // Pull-Up resistor for switch


// Rotary encoder 2 setup
stepper2.begin(180, 16);
pinMode(PinCLK2,INPUT);
pinMode(PinDT2,INPUT);
pinMode(PinSW2,INPUT);
digitalWrite(PinSW2, HIGH); // Pull-Up resistor for switch
```

```
//attachInterrupt (digitalPinToInterrupt(PinCLK2),isr2,FALLING); // interrupt 0
always connected to pin 2 on Arduino UNO

// Rotary encoder 3 setup
stepper3.begin(180, 16);
pinMode(PinCLK3,INPUT);
pinMode(PinDT3,INPUT);
pinMode(PinSW3,INPUT);
digitalWrite(PinSW3, HIGH); // Pull-Up resistor for switch

// Rotary encoder 4 setup
stepper4.begin(180, 16);
pinMode(PinCLK4,INPUT);
pinMode(PinDT4,INPUT);
pinMode(PinSW4,INPUT);
digitalWrite(PinSW4, HIGH); // Pull-Up resistor for switch

attachInterrupt (digitalPinToInterrupt(PinCLK2),isr2,CHANGE); // interrupt 0 always
connected to pin 2 on Arduino UNO
attachInterrupt (digitalPinToInterrupt(PinCLK),isr,CHANGE); // interrupt 0 always
connected to pin 2 on Arduino UNO
attachInterrupt (digitalPinToInterrupt(PinCLK3),isr3,CHANGE); // interrupt 3 setup
attachInterrupt (digitalPinToInterrupt(PinCLK4),isr4,CHANGE); // interrupt 3 setup
Serial.begin(9600); // opens serial port, sets data rate to 9600 bps
}

void loop() {
   if (!(digitalRead(PinSW))) {   // check if button is pressed
   if (RotaryPosition == 0) {  // check if button was already pressed
   } else {
     stepper1.move(-(RotaryPosition*i));
       RotaryPosition=0; // Reset position to ZERO
     }
   }



if (TurnDetected3 ||TurnDetected || TurnDetected2|| TurnDetected4)  {
```

```
if (TurnDetected3)  {

PrevPosition3 = RotaryPosition3; // Save previous position in variable
   if (rotationdirection3) {
     RotaryPosition3=RotaryPosition3-1;} // decrase Position by 1
   else {
     RotaryPosition3=RotaryPosition3+1;} // increase Position by 1

   TurnDetected3 = false;  // do NOT repeat IF loop until new rotation detected

   // Which direction to move Stepper motor
   if ((RotaryPosition3 + 1) == PrevPosition3) { // Move motor CCW

     AnglesToRotate3=-i*1.8/16;
     //stepper3.rotate(AnglesToRotate3);
   }
  else if ((PrevPosition3 + 1) == RotaryPosition3) { // Move motor CW

          AnglesToRotate3=i*1.8/16;
        //   stepper3.rotate(AnglesToRotate3);
   }
   else {AnglesToRotate3=0 ;}


// Serial print encoder
Serial.print("encodata3:");
Serial.println(encodata3,DEC);

}


if (TurnDetected)  {
   PrevPosition = RotaryPosition; // Save previous position in variable
   if (rotationdirection) {
     RotaryPosition=RotaryPosition-1;} // decrase Position by 1
   else {
     RotaryPosition=RotaryPosition+1;} // increase Position by 1

   TurnDetected = false;  // do NOT repeat IF loop until new rotation detected

   // Which direction to move Stepper motor
   if ((RotaryPosition + 1) == PrevPosition) { // Move motor CCW

     AnglesToRotate=-i*1.8/16;
```

```
 // stepper1.rotate(AnglesToRotate);
 }
else if ((PrevPosition + 1) == RotaryPosition) { // Move motor CW

      AnglesToRotate=i*1.8/16;
    //   stepper1.rotate(AnglesToRotate);
 }
else {AnglesToRotate=0 ;}

// Serial print encoder
   Serial.print("encodata1:");
Serial.println(encodata,DEC);

     }



if (TurnDetected2)  {
   PrevPosition2 = RotaryPosition2; // Save previous position in variable
   if (rotationdirection2) {
     RotaryPosition2=RotaryPosition2-1;} // decrase Position by 1
   else {
     RotaryPosition2=RotaryPosition2+1;} // increase Position by 1

   TurnDetected2 = false;  // do NOT repeat IF loop until new rotation detected

   // Which direction to move Stepper motor
   if ((RotaryPosition2 + 1) == PrevPosition2) { // Move motor CCW

     AnglesToRotate2=-i*1.8/16;
     //stepper2.rotate(AnglesToRotate2);
   }
  else if ((PrevPosition2 + 1) == RotaryPosition2) { // Move motor CW

      AnglesToRotate2=i*1.8/16;
    //    stepper2.rotate(AnglesToRotate2);
   }
   else {AnglesToRotate2=0 ;}

// Serial print encoder

Serial.print("encodata2:");
Serial.println(encodata2,DEC);

     }
```

```
if (TurnDetected4)  {

PrevPosition4 = RotaryPosition4; // Save previous position in variable
   if (rotationdirection4) {
     RotaryPosition4=RotaryPosition4-1;} // decrase Position by 1
   else {
     RotaryPosition4=RotaryPosition4+1;} // increase Position by 1

   TurnDetected4 = false;  // do NOT repeat IF loop until new rotation detected

   // Which direction to move Stepper motor
   if ((RotaryPosition4 + 1) == PrevPosition4) { // Move motor CCW

     AnglesToRotate4=-i*1.8/16;
     //stepper4.rotate(AnglesToRotate4);
   }
  else  if ((PrevPosition4 + 1) == RotaryPosition4) { // Move motor CW

           AnglesToRotate4=i*1.8/16;
         //  stepper4.rotate(AnglesToRotate4);
   }
   else {AnglesToRotate4=0 ;}



// Serial print encoder
Serial.print("encodata4:");
Serial.println(encodata4,DEC);
}

//Rx2=(AnglesToRotate3+AnglesToRotate*0.866025+AnglesToRotate2*0.5)*16/1.8;
//Ry2=(AnglesToRotate4+AnglesToRotate2*0.866025-0.5*AnglesToRotate)*16/1.8;
//Ry1=AnglesToRotate2*16/1.8;
//Rx1=AnglesToRotate*16/1.8;

Rx2=(AnglesToRotate3+AnglesToRotate*0.866025+AnglesToRotate2*0.5);
Ry2=(AnglesToRotate4+AnglesToRotate2*0.866025-0.5*AnglesToRotate);
Ry1=AnglesToRotate2-alpha*0.5*AnglesToRotate4*0.866025-
(alpha*beta)*0.5*AnglesToRotate3*0.5;
Rx1=AnglesToRotate-
alpha*0.5*AnglesToRotate3*0.866025+(alpha*beta)*0.5*AnglesToRotate4*0.5;
```

```
int j=6;

//for(int s=0; s<i; s++)
//{stepper1.rotate(Ry2/i);
// stepper2.rotate(Rx2/i);
// stepper3.rotate(Ry1/i);
// stepper4.rotate(Rx1/i);
//}

Ay2=Ry2/j;
Ax2=Rx2/j;
Ay1=Ry1/j;
Ax1=Rx1/j;


for(int s=0; s<j; s++)
{stepper1.rotate(Ay2);
 stepper2.rotate(Ax2);
 stepper3.rotate(Ay1);
 stepper4.rotate(Ax1);
}


AnglesToRotate=0;       // reset angles to rotate to zero
AnglesToRotate2=0;
AnglesToRotate3=0;
AnglesToRotate4=0;


}

}
```

# Appendix B. Drawings of Robotic Arm Components

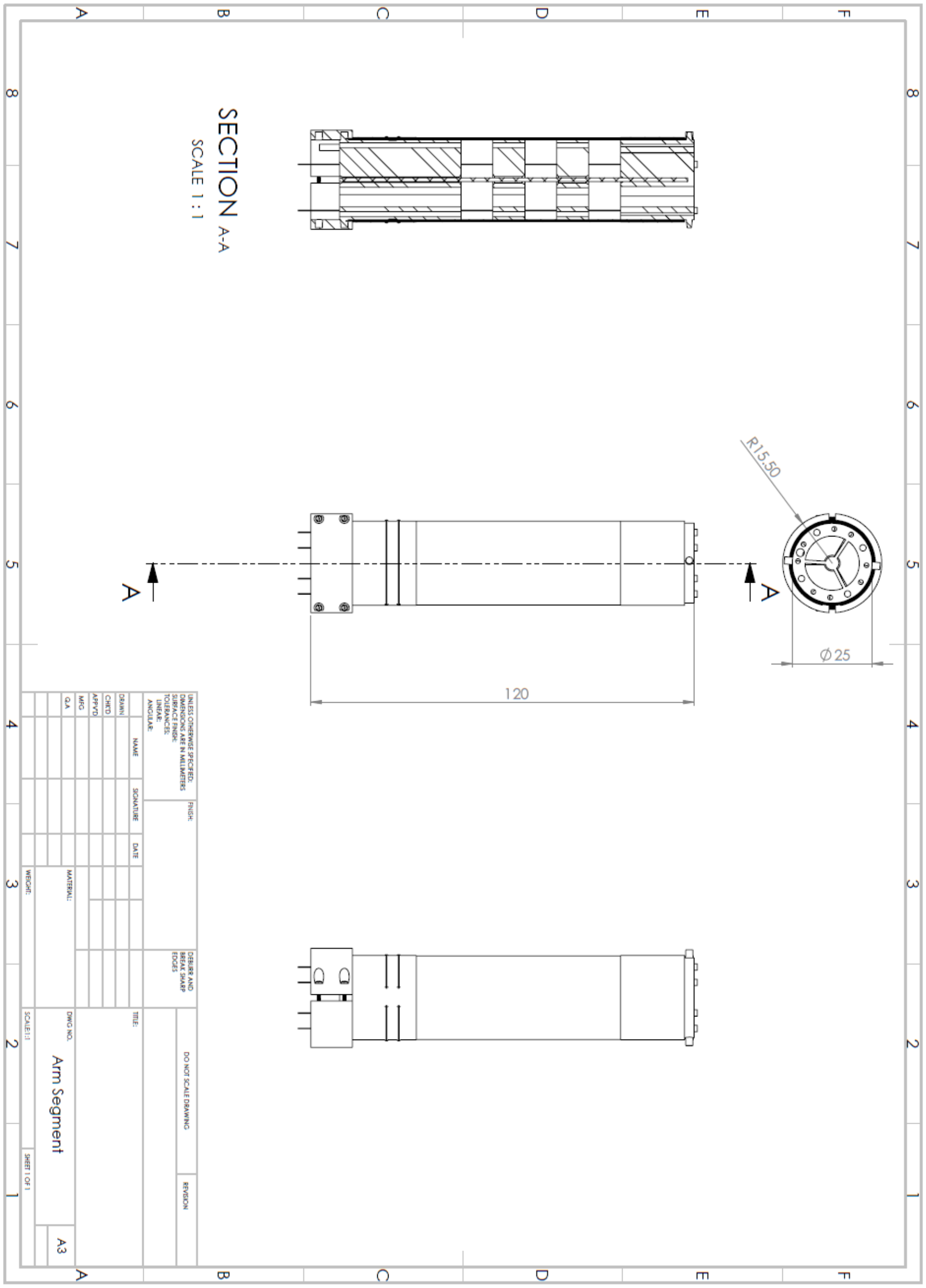

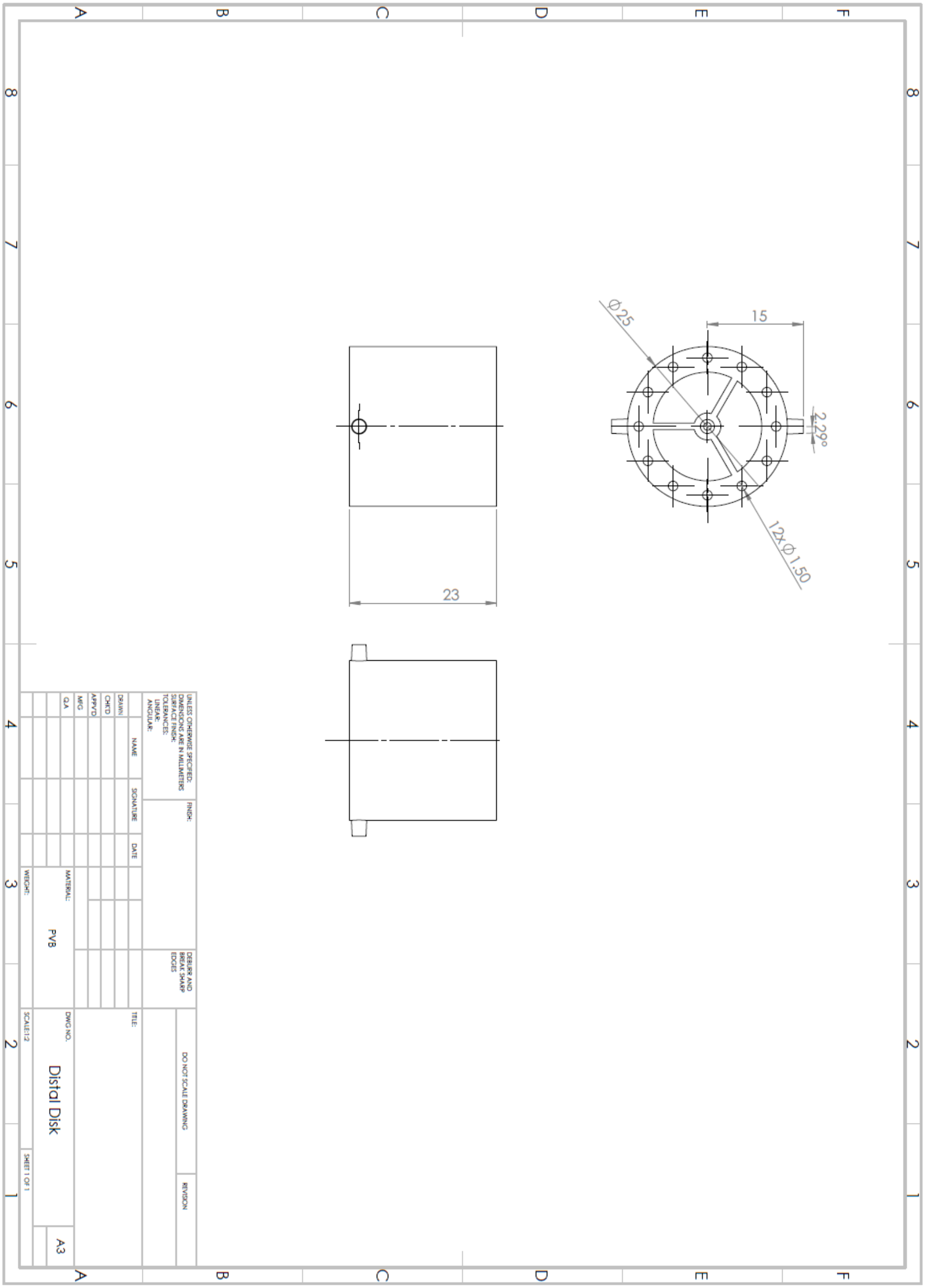

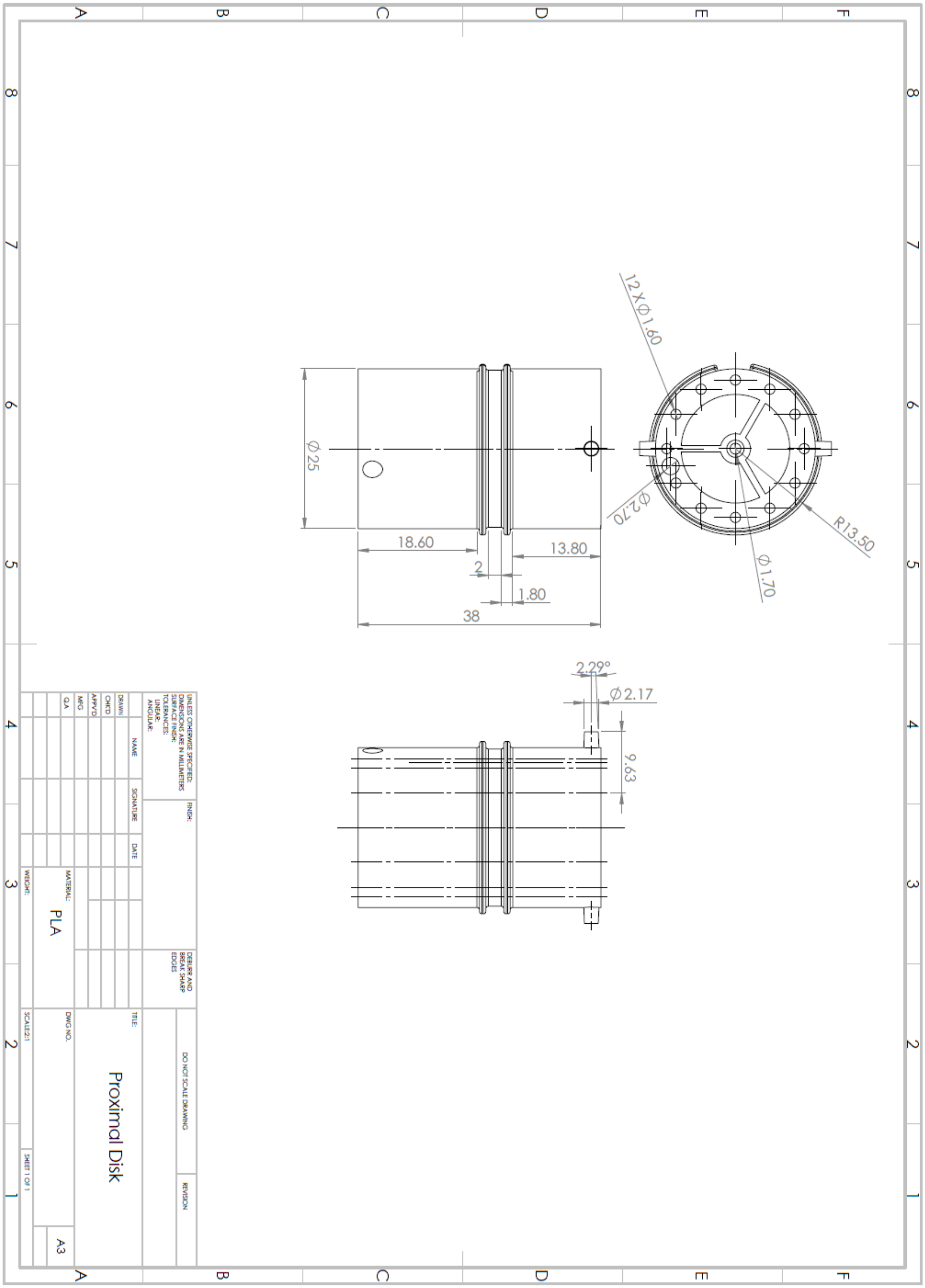

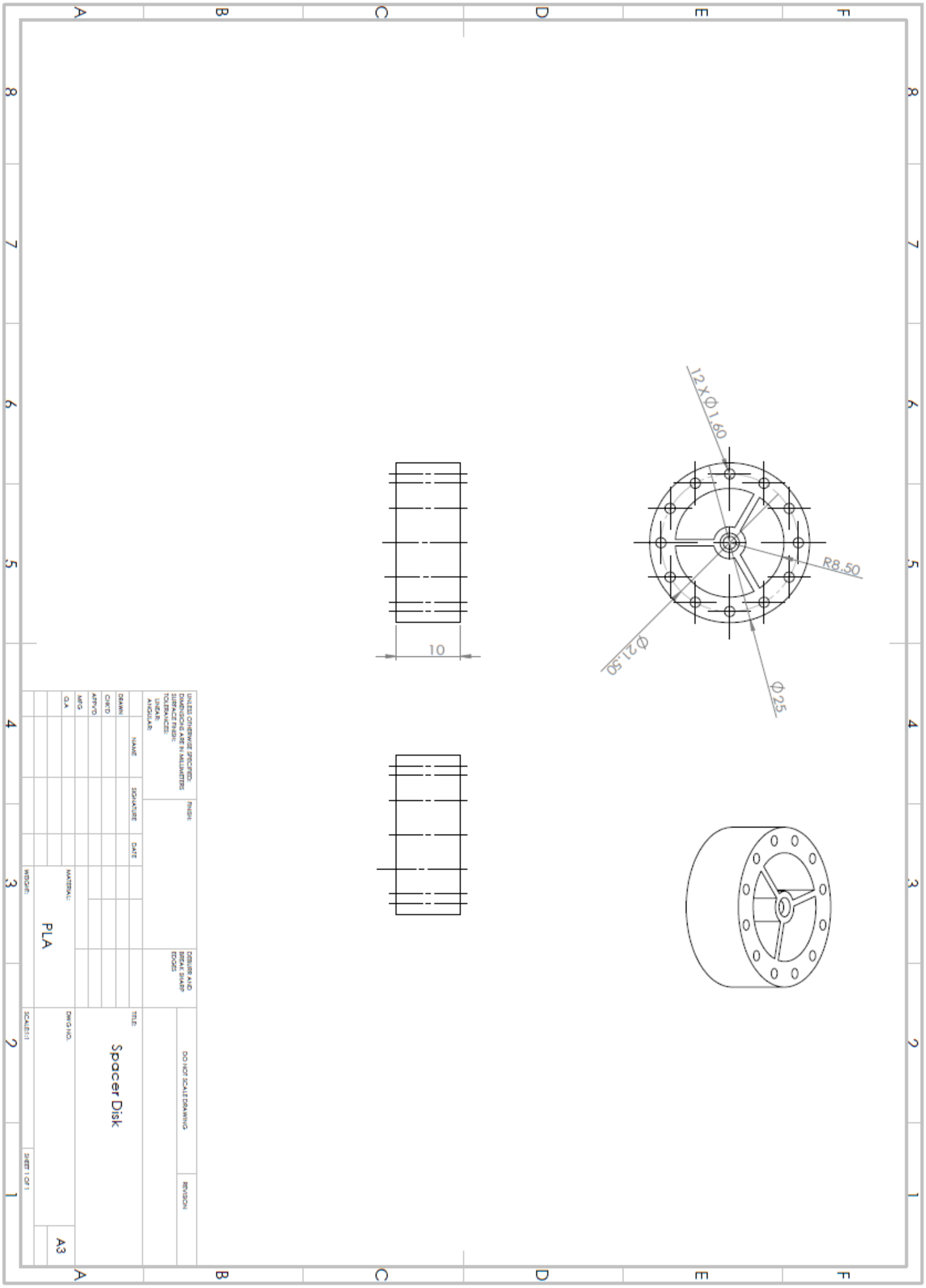

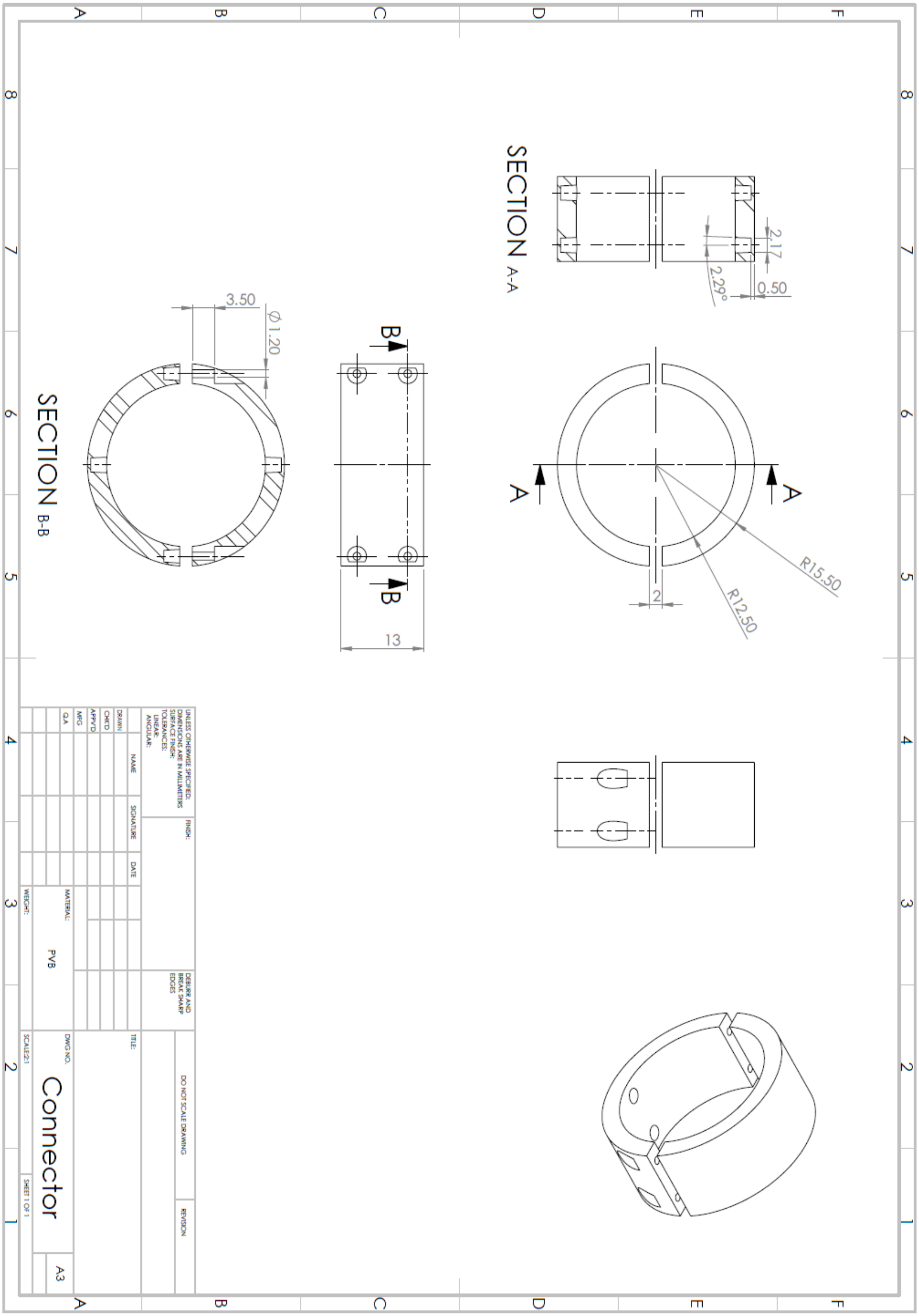